\documentclass[10pt,twocolumn,letterpaper]{article}

\usepackage{cvpr}              % To produce the CAMERA-READY version
\usepackage{xspace}
\newcommand{\paper}{\textbf{OneHOI}\xspace} % or OmniHOI

\usepackage[accsupp]{axessibility}  % Improves PDF readability for those with disabilities.

\usepackage{soul}
\setuldepth{foobar}

\usepackage{lipsum}
\usepackage{duckuments} % placeholder duck
\usepackage{multirow} % for table
\usepackage{siunitx} % for align neg values
\usepackage{booktabs}
\usepackage{threeparttable}
\usepackage[normalem]{ulem}
\useunder{\uline}{\ul}{}
\makeatletter
\newcommand\thefontsize{The current font size is: \f@size pt}
\makeatother
\usepackage{seqsplit} % for breakable text at any point 
\usepackage[misc,geometry]{ifsym} % envelope
\definecolor{cvprblue}{rgb}{0.21,0.49,0.74}
\usepackage[pagebackref,breaklinks,colorlinks,allcolors=cvprblue]{hyperref}

\def\paperID{} % *** Enter the Paper ID here 164
\def\confName{CVPR}
\def\confYear{2026}

\title{\paper: Unifying Human-Object Interaction Generation and Editing}

\author{
Jiun Tian Hoe$^{1}$ 
\quad Weipeng Hu$^{1,2}$\footnotemark[1]
\quad Xudong Jiang$^{1}$%\footnotemark[1]
\quad Yap-Peng Tan$^{1,4}$ 
\quad Chee Seng Chan$^{3}$\footnotemark[1] \\ % \vspace{0.3em}
{\normalsize $^1$Nanyang Technological University} \quad
{\normalsize $^2$Sun Yat-sen University} \quad
{\normalsize $^3$Universiti Malaya} \quad
{\normalsize $^4$VinUniversity} \\
{\tt\small Code and dataset: \url{https://jiuntian.github.io/OneHOI/}}
\vspace{-12pt}
}

\begin{document}
%\maketitle
\twocolumn[{
\maketitle
\renewcommand\twocolumn[1][]{#1}
\begin{center}
    \vspace{-12pt} %15, 16 to fit 1 line
    \captionsetup{type=figure}
    \includegraphics[width=0.95\linewidth]{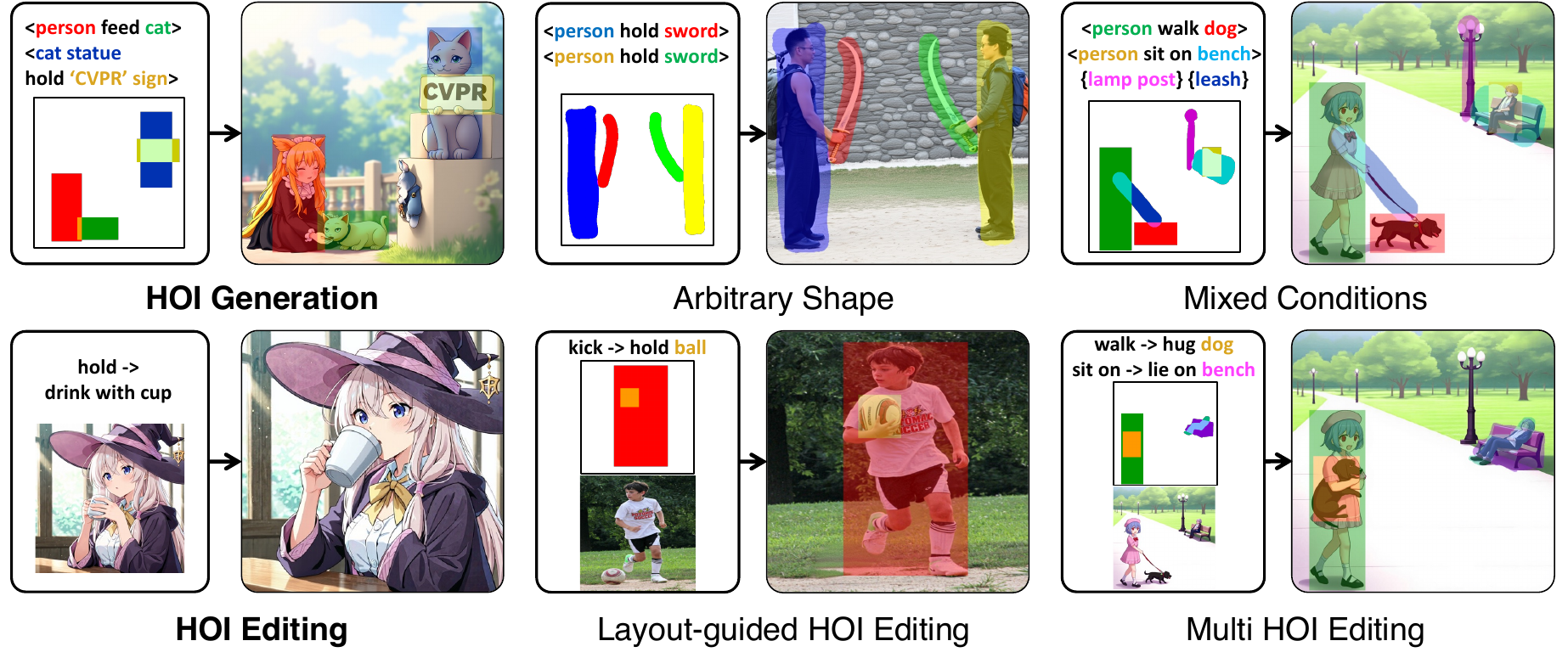}
    \vspace{-10pt} % 10
    \captionof{figure}{
    \paper unifies Human-Object Interaction (HOI) generation and editing in a single, versatile model. It excels at challenging HOI editing, from text-guided changes to novel layout-guided control and novel multi-HOI edits. For generation, \paper synthesises scenes from text, layouts, arbitrary shapes, or mixed conditions, offering unprecedented control over relational understanding in images.}
    \label{fig:teaser}
    %\vspace{-2pt}
\end{center}
}]
\footnotetext[1]{Corresponding authors: Weipeng Hu (huwp7@mail.sysu.edu.cn)  and \\ Chee Seng Chan (cs.chan@um.edu.my)}

% \blfootnote{{\textsuperscript \Letter} corresponding authors} % must insert next line below, so that "Abstract" title wouldn't skew to right

\begin{abstract}
Human-Object Interaction (HOI) modelling captures how humans act upon and relate to objects, typically expressed as \(\langle \text{person}, \text{action}, \text{object}\rangle\) triplets. Existing approaches split into two disjoint families: HOI generation synthesises scenes from structured triplets and layout, but fails to integrate mixed conditions like HOI and object-only entities; and HOI editing modifies interactions via text, yet struggles to decouple pose from physical contact and scale to multiple interactions. We introduce \paper, a unified diffusion transformer framework that consolidates HOI generation and editing into a single conditional denoising process driven by shared structured interaction representations. At its core, the Relational Diffusion Transformer (R-DiT) models verb-mediated relations through role- and instance-aware HOI tokens, layout-based spatial Action Grounding, a Structured HOI Attention to enforce interaction topology, and HOI RoPE to disentangle multi-HOI scenes. Trained jointly with modality dropout on our HOI-Edit-44K, along with HOI and object-centric datasets, \paper supports layout-guided, layout-free, arbitrary-mask, and mixed-condition control, achieving state-of-the-art results across both HOI generation and editing.
\end{abstract}

\section{Introduction}
\label{sec:intro}

Human-Object Interaction (HOI) lies at the forefront of visual understanding, focusing not just on what appears in an image but also on how entities relate. It represents the world through structured triplets \(\langle \text{person}, \text{action}, \text{object}\rangle\), capturing the grammar of interaction. Mastering HOI is crucial for next-generation AI, from building dynamic AR/VR worlds to enabling content creation that understands why and how things connect, not merely what they are.

\begin{figure}[ht]
    \centering
    \includegraphics[width=\linewidth]{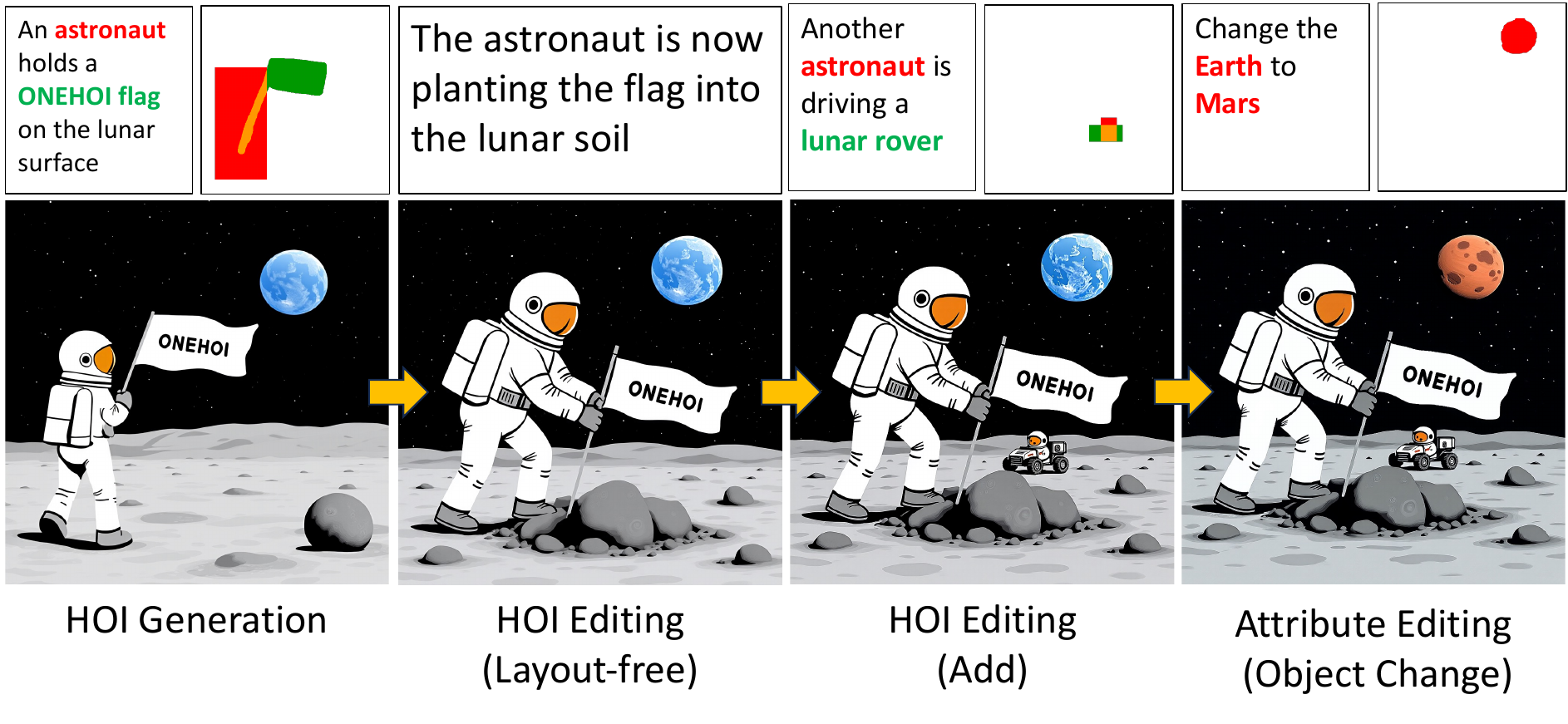}
    \vspace{-20pt} % 20
    \caption{\textbf{Unified HOI generation and editing.} \paper enables a single-model multi-step workflow. It begins with (i) \textbf{Mixed-Condition Generation}, synthesising a complex scene from layout-guided HOIs with arbitrary shape. Then, it performs (ii) \textbf{Layout-free HOI Editing}, (\eg,~change him to plant the flag), followed by (iii) \textbf{Layout-guided HOI Editing} (\eg,~add another astronaut and driving a rover) and (iv) \textbf{Attribute Editing} (\eg,~change to Mars). More examples in \cref{fig:workflow_girl} of the Appendix.
    }
    \label{fig:workflow_lunar}
    \vspace{-14pt} % 10
\end{figure}

Existing studies follow two main directions. \textit{Recognition and detection} approaches \cite{qahoi2021,cao2023unihoi,luo2024sichoi} aim to identify and localize HOI, improving perceptual understanding but offering no generative capability. \textit{Generative methods} \cite{hoe2024interactdiffusion,cha2025verbdiff,xu2025hoiedit,hoe2025interactedit} in contrast, have evolved into two disjoint families: \textbf{HOI generation}, which synthesises scenes from triplets conditioned on spatial layouts for controllability, but struggles with flexible control, such as \textit{mixing HOI triplets} with object-only entities or accepting \textit{arbitrary shape} layouts; and \textbf{HOI editing}, which modifies images via text, cannot reliably decouple and recompose pose and physical contact. Besides, it fails to scale beyond a single interaction, lacks fine spatial control, and relies on implicit priors rather than explicit structural modelling. % Both ultimately aim to teach models how humans interact with the world, from opposite ends of the spectrum.

This paper asks a simple but fundamental question: \textit{\textbf{Can HOI generation and editing be unified within a single framework?}} 
We posit that joint training creates a substantial synergy, as the broad interaction semantics (\eg, poses, contact points) learned during generation can provide the deep structural HOI knowledge that editing-only models lack, enabling more plausible and physically-aware edits.
% If both reconstruct or modify interactions under different forms of conditioning, they could share one generative formulation combining the spatial precision of generation with the semantic flexibility of editing.

% A unified model must meet two requirements: (i) fine-grained spatial controllability, and (ii) semantic coherence to preserve interaction intent. 
Achieving this unification requires a high-fidelity backbone with a flexible architecture for multi-modal conditioning. Diffusion Transformers (DiTs) \cite{Peebles2023DiT} are a promising candidate. They combine diffusion’s fidelity with transformers’ global reasoning to produce high-quality images \cite{esser2024sd3mmdit,labs2025flux1kontext,wu2025qwenimagetechnicalreport} and enable fine-grained spatial control \cite{zhang2025eligen}. Yet, they have a critical flaw: \textit{DiTs treat scenes as collections of independent objects and lack explicit interaction modelling, yielding visually detailed but relationally shallow results}.

To address this, we introduce \paper, a unified framework for HOI generation and editing. Our key insight is that both tasks are two views of a single conditional denoising process. Besides layouts and captions, our model also conditions on structured interaction representations, reframing diffusion from arranging pixels to realising relationships.

At the core of \paper\ lies a new Relational DiT (R-DiT) with three tightly coupled modules:
(i) \textit{HOI Encoder} to inject role- and instance-aware cues into HOI token;
(ii) \textit{Structured HOI Attention} to enforce a verb-mediated topology among HOI tokens and
(iii) \textit{HOI RoPE} to assign distinct positional identities to disentangle interactions in multi-HOI scenes. Together, these form a unified grammar that enables reasoning over interactions, not just regions. % (ii) and HOI role-specific layout constraints;

Trained jointly for generation and editing on our new HOI-Edit-44K dataset with modality-dropout, supplemented by established HOI and object-level datasets, \paper\ is a unified pipeline that supports layout-guided, layout-free, arbitrary-mask, and mixed-condition controls, handling single and multiple interactions, see \cref{fig:teaser,fig:workflow_lunar}.

Our main contributions are:
\begin{itemize}
\item \paper, a unified DiT-based framework for HOI generation and editing, scaling to multi-HOI scenes and, for the \textit{first time}, enabling multi-HOI editing.
\item A novel R-DiT that embeds explicit interaction representations via three modules (\ie~HOI Encoder, Structured HOI Attention, and HOI RoPE), enabling precise yet flexible control under diverse conditions, including layout-guided, layout-free, arbitrary masks, and mixed inputs. 
\item A new large-scale paired dataset, HOI-Edit-44K, addressing the scarcity of paired data, with 44K identity-preserving examples, for training of robust HOI editing.
\item State-of-the-art performance across benchmarks for controllable HOI generation, layout-free editing, and novel layout-guided single- and multi-HOI editing tasks.
\end{itemize}
\section{Related Works}
\label{sec:related}
\begin{figure*}[t]
    \centering
    \begin{minipage}{.70\linewidth}
            \begin{subfigure}[t]{1\linewidth}
                \includegraphics[width=\textwidth]{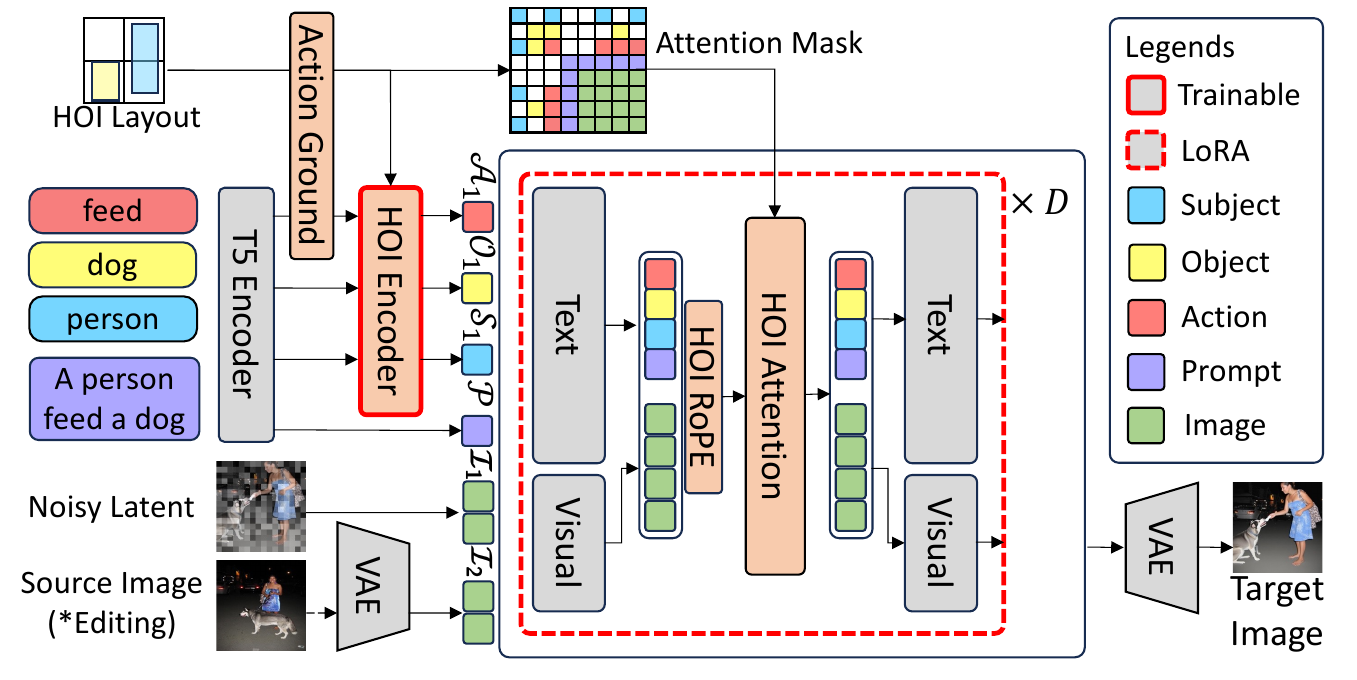}
                \caption{An overview of \paper pipeline.}
                \label{fig:arch}
            \end{subfigure}
        \end{minipage}
    \hfill\vline\hfill
    \begin{minipage}{.27\linewidth}
        % \centering
        \begin{subfigure}[t]{1\linewidth}
            \centering
            \includegraphics[width=.57\linewidth]{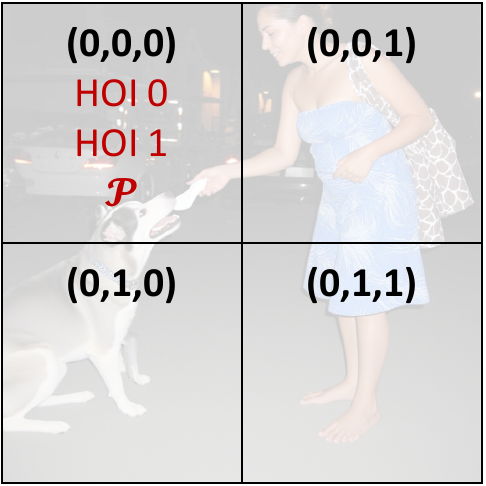}
            \caption{Original RoPE}
            \label{fig:ori-rope}
        \end{subfigure} \\
        \begin{subfigure}[b]{1\linewidth}
            \centering
            \includegraphics[width=.57\linewidth]{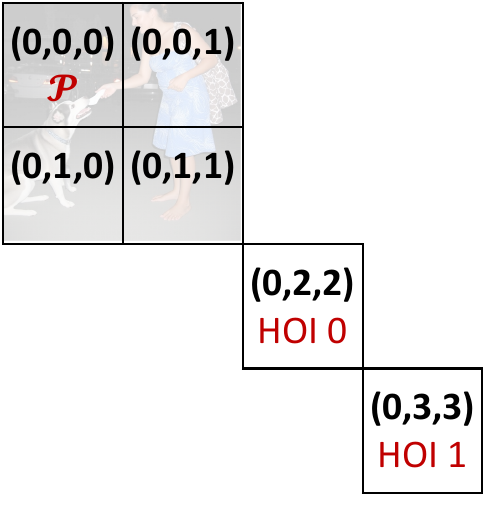}
            \caption{HOI RoPE}
            \label{fig:rope}
        \end{subfigure} 
    \end{minipage}
    %\vspace{-6pt} % 5
    \caption{
    % \paper unifies both HOI editing and generation tasks with any combinations of conditions on a DiT backbone. HOI Encoder injects role, instance, and layout cues into HOI tokens. HOI RoPE provides unique positional indices for each interaction instance to ensure separation. Structured HOI Attention (within the backbone) uses masking to enforce verb-mediated topology and spatial grounding.
    \textbf{(a)} \paper unifies HOI editing and generation tasks on a DiT backbone. The pipeline features an HOI Encoder to inject role and instance cues, and Structured HOI Attention to enforce verb-mediated topology and spatial grounding. \textbf{(b, c)} To separate instances, in contrast to the Original RoPE (b), HOI RoPE (c) provides unique positional indices for each interaction.
    }
    \vspace{-10pt} % 8
\end{figure*}

\noindent\textbf{Controllable Generation and Human-Object Interaction.}
Research on fine-grained control \cite{lian2023llmgrounded} and spatial conditioning (e.g., GLIGEN \cite{gligen2023}, MIGC \cite{zhou2024migc} and EliGen \cite{zhang2025eligen}) has enabled object placement via layouts or attention manipulation. However, they focus on individual entities, specifying \textit{where} objects are, but not \textit{how they relate}.
Generative HOI research addresses this gap, diverging into:
\begin{itemize}
    \item \textbf{Layout-Conditioned Generation.} Methods like InteractDiffusion \cite{hoe2024interactdiffusion} synthesise images from triplets conditioned on spatial layouts for controllability, but struggle with flexible control (\eg, \textit{mixing HOI triplets} with object-only entities or accepting \textit{arbitrary shape} layouts) and fail when layout guidance is partial or absent.
    \item \textbf{Text-Guided Editing.} Methods like HOIEdit \cite{xu2025hoiedit} and InteractEdit \cite{hoe2025interactedit} modify interactions in existing images. They cannot reliably decouple and recompose the pose and physical contact, fail to scale beyond a single interaction, lack precise spatial control, and rely on implicit model priors rather than explicit interaction modelling.
\end{itemize}
This fragmented development leaves a clear gap: no unified framework bridges these modalities and the multi-HOI editing is largely unaddressed.%, and the HOI's relational power has not been integrated into the superior DiT architecture. 
\textbf{ \paper addresses these limitations directly}, introducing the first framework to unify generation and editing, enabling precise yet flexible control under diverse conditions (\eg, layout-guided, layout-free, arbitrary masks, and mixed inputs) within one model. % bring structured HOI modelling to DiTs,

\noindent\textbf{Diffusion Transformers (DiTs) for Image Synthesis.} The landscape of image synthesis has been reshaped by diffusion models \cite{ddpm2020,ddim2021}, which have rapidly surpassed GANs in generating high-fidelity images. Latent Diffusion Models \cite{stablediffusion2021} democratised this by operating in a compressed latent space, significantly reducing computational costs. While early models used U-Nets \cite{unet2015}, DiTs \cite{Peebles2023DiT} marked a pivotal shift. Replacing convolutions with a pure transformer architecture yielded superior scaling properties, establishing DiTs as the new standard. State-of-the-art systems like Flux.1 \cite{labs2025flux1kontext} and Qwen-Image \cite{wu2025qwenimagetechnicalreport} leverage Multi-Modal DiT (MM-DiT) \cite{esser2024sd3mmdit} variants with flow-matching objectives \cite{lipman2023flowmatching}, achieving unprecedented quality and controllability, yet they lack explicit interaction modelling.
% confirming DiTs as the premier architecture for generation.

% \noindent\textbf{Human-Object Interaction (HOI)} represents structured relations between people and objects as a $\left\langle \text{human, action, object} \right\rangle$ triplet. Early work primarily addressed detection: localizing human and object instances and classifying their interaction category from a fixed vocabulary, with extensions to Visual-Language models-based reasoning \cite{geng2025horp}, long-tail handling \cite{cao2023unihoi}, and open-vocabulary detection \cite{thid2022,lei2024CMD-SE}. More recently, generative formulations have appeared. InteractDiffusion \cite{hoe2024interactdiffusion} synthesizes HOI images by conditioning on triplets and \emph{precise} bounding boxes, offering strong spatial control and scaling to multi-HOI scenes. Complementarily, HOI editing modifies the interaction in an existing image while preserving subject/object identity; recent methods operate with text-only guidance \cite{xu2025hoiedit,hoe2025interactedit}, thus providing box-free control but typically remaining single-HOI and lacking fine spatial specificity. Despite this progress, the landscape remains bifurcated: generation methods generally assume precise boxes (thereby supporting multi-HOI), whereas editing methods are box-free but neither multi-HOI nor precisely box-controllable. We pursue a unified formulation that supports both generation and editing, \emph{with or without} boxes and for single or multiple interactions, within one framework.

\section{Methodology}\label{sec:method}

\Cref{fig:arch} overviews our unified pipeline for HOI generation and editing. Given a global text prompt $\mathcal{P}$ and either a set of structured interaction $\{\langle s,o,a\rangle_n\}_{n=1}^N$ or independent objects $\{\langle o\rangle_n\}_{n=1}^N$ with optional layout $\mathcal{B}=\{b^s_n,b^o_n\}$ or $\mathcal{B}=\{b^o_n\}$, our pipeline produces an image that realises all specified targets. We denote the sets of T5 \cite{2020t5}-encoded tokens corresponding to these triplets as $\mathcal{H}=\bigcup_{n=1}^N\{\mathcal{S}_n,\mathcal{A}_n,\mathcal{O}_n\}$, where $\mathcal{S}_n, \mathcal{O}_n, \mathcal{A}_n$ represent subject, object, and action tokens, respectively for instance $n$. For generation, we sample noise $\mathcal{I}_1$ in the latent space and run the conditional denoiser. For editing, we encode the source image into latents $\mathcal{I}_2$, concatenate them with the noise $\mathcal{I}_1$, and run the \emph{same} denoiser conditioned on the new interaction targets. 

% Our core idea is to build a representation within the DiT that \emph{explicitly models interaction structure}, moving beyond simple object placement. We achieve this 
Our core idea is the introduction of Relational DiT (R-DiT), a modified backbone that \textit{explicitly models interaction structure}. We build the R-DiT by introducing four key components to a standard layout-conditioned DiT baseline, Eligen \cite{zhang2025eligen}, as validated in our ablation (\cref{subsec:ablation}). These components inject increasingly sophisticated relational understanding: (i) \textbf{Action Grounding}, which introduces action-specific semantic and spatial cues; (ii) \textbf{HOI Encoder}, adding fine-grained role and instance identity; (iii) \textbf{Structured HOI Attention}, enforcing a verb-mediated attention topology and layout constraints; and (iv) \textbf{HOI RoPE}, ensuring interaction instances separation in complex scenes. More details in \cref{supp-sec:implementation-details}.

\begin{figure}[t]
    \centering
    \includegraphics[width=\linewidth]{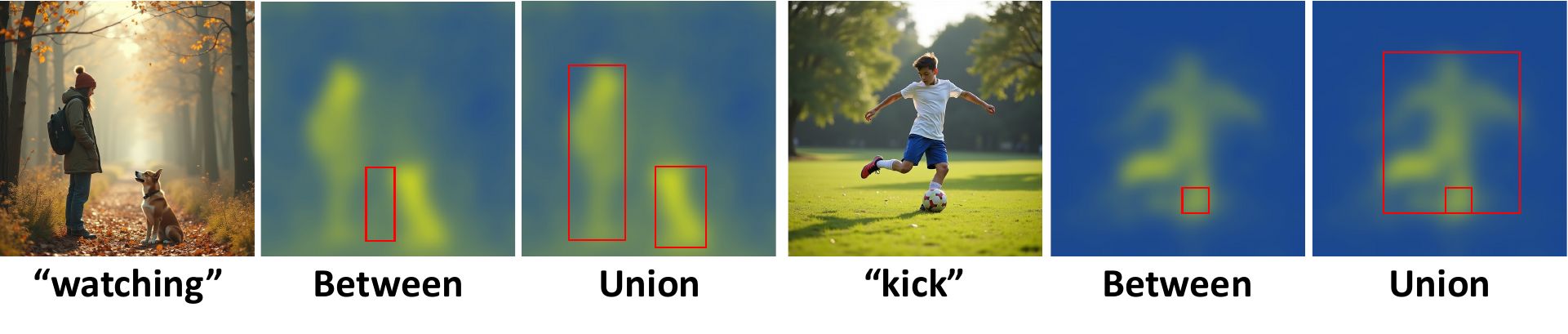}
    \vspace{-24pt} % 20
    \caption{Action-token\textrightarrow image attention heatmaps from the baseline. The ``Between" region proposed in InteractDiffusion \cite{hoe2024interactdiffusion} misses where the action actually attends, while our ``Union" region (subject $\cup$ object) better matches the attention footprint.}
    \label{fig:union}
    \vspace{-14pt} % 10
\end{figure}

% We cast both tasks as the \emph{same} conditional denoising problem driven by two inputs: (i) \textbf{HOI tokens} with role/instance cues (from the HOI Encoder), and (ii) optional \textbf{layout} (subject/object boxes and an action region). Our HOI Attention captures intra-interaction structure and, when layout is provided, restricts HOI-image attention to each role's designated region while remaining compatible with layout-free inputs. A single DiT-based denoiser is trained under both regimes, yielding precise spatial control when layout is provided and graceful behaviour under text-only conditioning. Together with HOI RoPE, this formulation scales to multi-interaction scenes with clear instance separation.

\subsection{Action Grounding}

Standard layout-conditioned models only ground objects. To model interactions, however, the model must also have basic awareness of the \emph{action} itself, both semantically and spatially.
We introduce \textit{Action Grounding} (AG) to provide this foundational capability. It builds upon a baseline that grounds subject $\mathcal{S}_n$ and object $\mathcal{O}_n$ tokens to regions $R_n^s$ and $R_n^o$ by introducing two action-specific cues: \textbf{(i) Semantic Action Token} $\mathcal{A}_n$ (T5~\cite{2020t5}-encoded) for each action label (\eg~"feed") in the HOI triplet and \textbf{(ii) Spatial Action Region} $R_n^a$ associated with this action.

Previous work \cite{hoe2024interactdiffusion} defines the action region with a ``between'' operator, which uses the intersection of the subject and object boxes when they overlap, else a rectangle spanning them when they disjoint. While adequate as a conditioning cue, this band often fails to match \textit{where} the action token actually attends (too narrow or misplaced; see \cref{fig:union}).

We define it instead as the \textbf{union} of the subject and object regions. By rasterising the subject and object shapes/boxes $b_{n}^{s},b_{n}^{o}$, we form regions $R_{n}^{s}$ and $R_{n}^{o}$, and set $R_n^a=R_n^s \cup R_n^o$. This choice (i) aligns better with the natural attention patterns of DiT, (ii) is robust for both overlapping and disjoint pairs, and (iii) provides a stable target for grounding the action (via \cref{subsec:attention}). This establishes the foundational understanding of interaction that is missing in object-only models, upon which our subsequent modules are built.

\subsection{HOI Encoder}
Models risk \emph{role confusion} or \emph{blending wrong interactions} in multi-HOI scenes. For example, given \textlangle person1, chase, dog\textrangle~and \textlangle person2, hold, cat\textrangle, a model might incorrectly render `person1' \textit{holding} the `cat' (\textbf{blending wrong interactions}) or a \textit{dog chasing} `person1' (\textbf{role confusion}). Hence, simply providing $\mathcal{S}_n, \mathcal{O}_n, \mathcal{A}_n$ tokens is insufficient. The model must explicitly know \emph{which token plays which role} (subject/object/action) and \emph{which interaction instance} it belongs to. HOI Encoder tackles this by injecting compact, explicit identity cues into the HOI token streams $\mathcal{H}$.
% HOI Encoder addresses this by augmenting the HOI token streams $\mathcal{H}$ with compact, explicit identity cues.
{
\setlength{\abovedisplayskip}{1pt}
\setlength{\belowdisplayskip}{3pt}
\noindent\textbf{Formulation.}
Let \(d\) as T5 output dimension (\(d{=}4096\)). For an interaction instance \(n\) and role \(r\in\{s,o,a\}\), let \(x^{r}_{n}\in\mathbb{R}^{d}\) be the T5-embedding. We build three side signals:
\[
e_{\text{role}}(r)\in\mathbb{R}^{64},\quad
e_{\text{inst}}(n)\in\mathbb{R}^{64},\quad
e_{\text{box}}(b_n^r)\in\mathbb{R}^{256},
\]
where \(e_{\text{role}}(r)\) is a learnable role embeddings, \(e_{\text{inst}}(n)\) is a fixed sinusoidal embedding of the instance index, and $e_{\text{box}}(b_n^r)$ is Fourier embedding \cite{nerf2022} of the role's box.}

We then normalize the HOI token $h^r_n$ with Layer Normalization, concatenate it with the side signals and project the result with a small MLP, and apply a gated residual:
\begin{align}
    \tilde{h}^r_n &=\mathrm{MLP}([\mathrm{LN}(h^r_n); e_\text{box}(b^r_n); e_\text{role}(r); e_\text{inst}(n)]),\\ % \in\mathbb{R}^{d}
    \tilde{h}^r_n &=h^r_n + \tanh(\lambda)~\cdot~\tilde{h}^r_n,
\end{align}
where \(\lambda\in\mathbb{R}\) is a learnable gate that smoothly ramps in the conditioning to stabilise training. The augmented tokens \(\tilde{h}^{r}_{n}\) are then fed into the DiT backbone. This provides the fine-grained identity information necessary for multi-HOIs relational modelling.

% relational modelling in multi-interaction contexts.

\begin{figure}[t]
    \centering
    \begin{subfigure}[t]{0.58\linewidth} % 0.63
        \centering
    \includegraphics[width=1\linewidth]{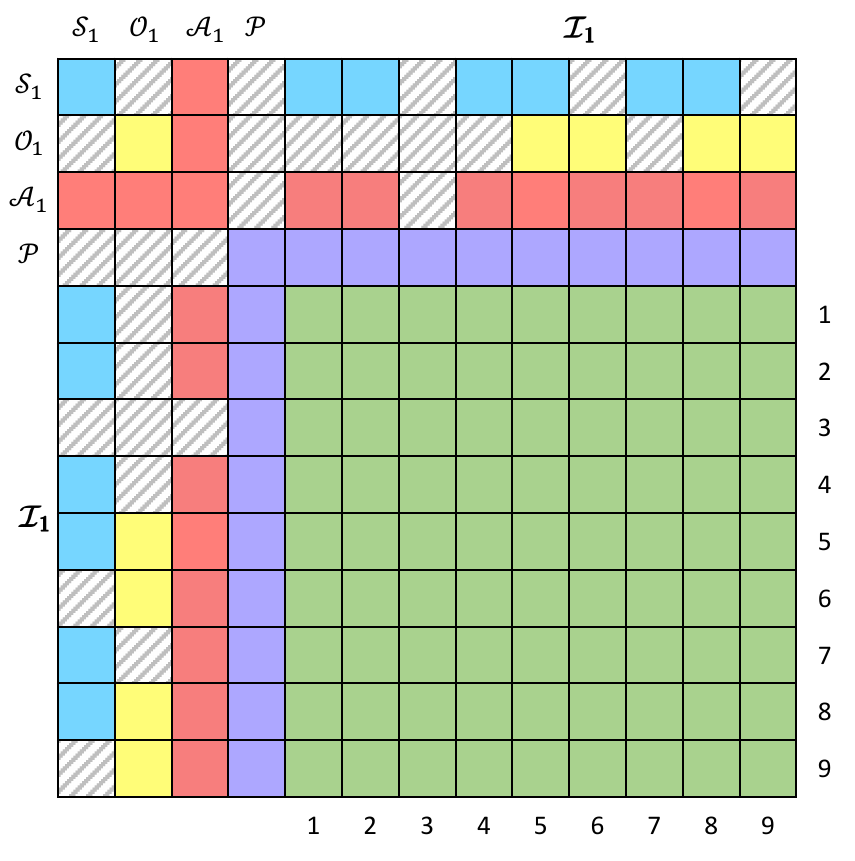}
        \vspace{-20pt}
        \caption{}
    \end{subfigure}
    \hfill
    \begin{subfigure}[t]{0.33\linewidth}
        \centering
        \raisebox{30pt}{
        \includegraphics[width=0.93\linewidth]{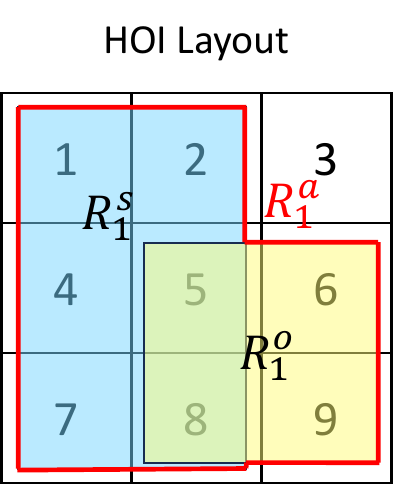}
        }
        \vspace{-20pt}
        \caption{}
    \end{subfigure}
    \vspace{-10pt}
    \caption{(a) HOI attention mask. Colours match \cref{fig:arch} legend, grey hatched indicates blocked attention. Direct $\mathcal{S}_n{\leftrightarrow}\mathcal{O}_n$ is blocked to enforce verb-mediated topology. $\mathcal{S}_n,\mathcal{O}_n,\mathcal{A}_n$ attend to image $\mathcal{I}_1$ only within $R^s_n,R^o_n,R^a_n$, respectively, as shown in (b).}
    \label{fig:attn_mask}
    \vspace{-12pt} % 10
\end{figure}

\subsection{Structured HOI Attention}
\label{subsec:attention}
Standard layout conditioning often treats subjects and objects as \emph{independent entities}. This means it can place them correctly but fails to capture the interaction structure, as it ignores the specific semantic and geometric relationship dictated by the \emph{action}. This independence leads to plausible but incorrect outputs, such as failing to render the 'holding' interaction in \cref{fig:ablation}-(2) or generating other awkward poses. We introduce Structured HOI Attention to explicitly embed this relational structure via a \emph{verb-mediated} attention topology. It governs attention patterns via masking, controlling both how HOI tokens $\mathcal{H}$ interact amongst themselves and how they ground to the image $\mathcal{I}$.

\noindent\textbf{HOI$\leftrightarrow$HOI Topology.} 
Our key insight is that action is central to defining the interaction structure. For each instance $n$, we prevent the direct links between subject$\leftrightarrow$object and enforce a verb-mediated pathway (cf. top-left of \cref{fig:attn_mask}):
\begin{equation}\nonumber
\mathcal{S}_n \leftrightarrow \mathcal{A}_n,\qquad
\mathcal{O}_n \leftrightarrow \mathcal{A}_n,\qquad
\text{\emph{block}}\ \ \mathcal{S}_n \leftrightarrow \mathcal{O}_n .
\end{equation}
All cross-instance HOI links $(n{\neq}m)$ are also disabled. This forces relational information to flow through the action tokens $\mathcal{A}_n$, directly reflecting the interaction's structure.

\noindent\textbf{HOI$\leftrightarrow$Image Grounding.}
When layout is provided, we constrain HOI$\rightarrow$image attention between HOI query $q\in\{\mathcal{S}_n,\mathcal{A}_n,\mathcal{O}_n\}$ and image key $k\in\mathcal{I}$ as:
\begin{equation}
M{_\mathcal{HI}}(q,k)=
\begin{cases}\label{eq:attn}
0, & q\in\mathcal{S}_n \ \text{and}\ k\in R^s_n,\\
0, & q\in\mathcal{O}_n \ \text{and}\ k\in R^o_n,\\
0, & q\in\mathcal{A}_n \ \text{and}\ k\in R^a_n,\\
-\infty, & \text{otherwise.}
\end{cases}
\end{equation}
This rule applies symmetrically for image$\rightarrow$HOI attention. When layout is absent, these constraints are removed (all connections allowed). This component compels the model to learn the semantic and spatial structure of the interaction.
{\setlength{\abovedisplayskip}{3pt}
\setlength{\belowdisplayskip}{3pt}
\noindent\textbf{Final Attention.}
The attention mask $\mathcal{M}$ (\cref{fig:attn_mask}) aggregates (i) the \textbf{HOI$\leftrightarrow$HOI topology}, (ii) the \textbf{HOI$\leftrightarrow$image grounding} constraints $M{_\mathcal{HI}}$, and (iii) the standard connections for prompt$\leftrightarrow$image and image$\leftrightarrow$image. The prompt$\leftrightarrow$HOI tokens are blocked. The final attention is:
\begin{equation}\label{eq:final-attn}
\mathrm{Attn}(Q,K,V,\mathcal{M})
=\mathrm{softmax}\!\left(\frac{QK^\top}{\sqrt{d}}+\mathcal{M}\right)V,
\end{equation}
with $\mathcal{M}_{qk}=0$ for allowed pairs and a large negative value (implementing $-\infty$) otherwise.}

% \noindent\textbf{Token blocks and binary regions.}
% We have image tokens $\mathcal{I}$, prompt tokens $\mathcal{P}$, and HOI tokens $\mathcal{H}=\bigcup_{n=1}^N\{\mathcal{S}_n,\mathcal{A}_n,\mathcal{O}_n\}$. For each instance $n$, we rasterize the subject/object boxes to binary regions over $\mathcal{I}$: $R^s_n=\mathrm{rast}(b^s_n)$ and $R^o_n=\mathrm{rast}(b^o_n)$, and define the action region as their union $R^a_n=R^s_n\cup R^o_n$.
% Each $R^\cdot_n\in\{0,1\}^{|\mathcal{I}|}$ indexes image tokens; when layout is absent we use all-ones vectors. Prompt–image connectivity follows the base Flux pipeline and is unchanged; HOI$\leftrightarrow\mathcal{P}$ is disabled.

\subsection{HOI RoPE (HRoPE)} 
Processing multi-HOIs simultaneously risks ``cross-talk'', where feature from one instance leaks and influences another, causing blended interactions or attributes swap. For instance, given \textlangle person1, chase, dog\textrangle~and \textlangle person2, hold, cat\textrangle, cross-talk might cause the model to generate ``person1 holding the cat'', incorrectly blending the two instances. HOI RoPE is a specialized positional indexing scheme to separate interaction instances. 
It is applied to the query $Q$ and key $K$ for all HOI tokens $\mathcal{H}$ in the attention (\cref{eq:final-attn}).
The image stream uses 3D RoPE \cite{su2024rope} over a spatial grid of size $H\times W$ following \cite{labs2025flux1kontext}. We assign all HOI tokens $\mathcal{H}$ belonging to the same instance $n$ a single, distinct positional index from the image grid and other instances:
\begin{equation}\nonumber
z_{\text{HOI}}(n)=(0,~ T{+}n, T{+}n), \quad\text{where}\quad T=\max(H,W).  
\end{equation}
This assigns each interaction a unique ``slot'' in the RoPE space (cf. \cref{fig:rope}). Applied across all layers, HRoPE reduces inter-instance interference in multi-HOI scenes.

% \noindent\textbf{Multiple interactions.}
% The HOI$\leftrightarrow$HOI mask is block-diagonal over instances and the HOI$\leftrightarrow$image masks use per-instance regions $(R^s_n,R^o_n,R^a_n)$, the construction scales to arbitrary $N$.

\section{Experiments}
\label{sec:experiment}

\begin{table*}[t]
\caption{Quantitative comparison for \textbf{layout-free HOI editing} on IEBench benchmark. Our method significantly outperforms others across all metrics for editing and image quality. Best results are in \textbf{bold}, second best are \uline{underlined}. Final row shows the closed-source baseline.}
\label{tab:hoiedit}
\vspace{-8pt} % 5
\centering
\resizebox{0.9\linewidth}{!}{%
\begin{tabular}{@{}lccccS[table-format=-1.4]@{}}
\toprule
\multirow{2}{*}{Method} & \multicolumn{2}{c}{HOI Editing}        & \multicolumn{3}{c}{Image Quality}                  \\ \cmidrule(l){2-3} \cmidrule(l){4-6} 
                        & Editability-Identity & HOI Editability & PickScore      & HPS             & {ImageReward}   \\ \midrule
Null-Text Inversion \cite{hertz2022p2p,mokady2023nti}     & 0.443                & 0.390           & 20.81          & 0.2483          & -0.3329         \\
% PnP Diffusion           & 0.365                & 0.240           & 20.59          & 0.2520          & -0.6624         \\
MasaCtrl \cite{cao2023masactrl}                & 0.371                & 0.260           & 20.14          & 0.2212          & -0.7136         \\
HOIEdit \cite{xu2025hoiedit}                & 0.349                & 0.240           & 19.51          & 0.2129          & -1.0289         \\
InstructPix2Pix \cite{brooks2022instructpix2pix}        & 0.380                & 0.269           & 20.28          & 0.2178          & -0.7717         \\
TurboEdit \cite{deutch2024turboedit}              & 0.434                & 0.326           & 20.36          & 0.2437          & -0.3821         \\
EditFriendlyDDPM \cite{huberman2024editfriendly}       & 0.438                & 0.320           & 20.48          & 0.2470          & -0.3875         \\
OmniGen \cite{xiao2025omnigen}                & 0.354                & 0.231           & 19.74          & 0.2120          & -1.0055         \\
FireFlow \cite{deng2024fireflow}               & 0.451                & 0.350           & 20.76          & 0.2530          & -0.4385         \\
Flux.1 Kontext \cite{labs2025flux1kontext}         & 0.471                & 0.328           & 20.45          & 0.2427          & -0.5137         \\
OmniGen2 \cite{wu2025omnigen2}                & 0.496                & 0.437           & 20.90          & 0.2595          & -0.0869         \\
Qwen Image Edit \cite{wu2025qwenimagetechnicalreport}        & {\ul 0.580}          & 0.460           & 20.81          & 0.2585          & 0.0748          \\
InteractEdit \cite{hoe2025interactedit}           & 0.573                & {\ul 0.514}     & {\ul 21.08}    & {\ul 0.2640}    & \uline{0.1630}    \\ 
\midrule
Ours                    & \textbf{0.638}       & \textbf{0.596}  & \textbf{21.26} & \textbf{0.2805} & \textbf{0.4713} \\
Improvements            & 10.0\%               & 16.0\%          & 0.85\%         & 6.25\%          & {189\%}         \\ 
\midrule
Nano Banana & 0.623 & 0.530 & 20.97 & 0.2544 & 0.1810 \\	
\bottomrule
\end{tabular}%
}
\vspace{-5pt}
\end{table*}
\begin{table*}[t]
\centering
\caption{Quantitative results for our novel \textbf{layout-guided HOI editing} tasks. We report strong performance for both single- and multi-HOI editing, establishing the first baseline for these new capabilities.}
\label{tab:layout-hoiedit}
\vspace{-8pt} % 5
 \resizebox{0.9\linewidth}{!}{%  0.8
\centering
\begin{threeparttable}[b] % three-part-table

\begin{tabular}{@{}lccccccS[table-format=-1.4]@{}}
\toprule
\multirow{2}{*}{~~Task} & \multirow{2}{*}{Method}                                                      & \multicolumn{3}{c}{Layout-guided HOI Editing}    & \multicolumn{3}{c}{Image Quality}  \\ \cmidrule(l){3-5} \cmidrule(l){6-8} 
                      & & Editability-Identity & HOI Editability & Spatial & PickScore & HPS    & {ImageReward} \\ \midrule
\multicolumn{1}{r}{\multirow{2}{*}{Single HOI Editing}} 
& InteractEdit + InteractDiffusion & 0.559 & 0.520 & 0.749 & 20.53 & 0.2418 & -0.3072 \\
& Ours~~ & \textbf{0.638}                & \textbf{0.570}           & \textbf{0.822}   & \textbf{21.04}     & \textbf{0.2678} & \textbf{0.2897}        \\ \midrule
\multicolumn{1}{r}{Multi HOI Editing} & Ours$^*$ & 0.435                & 0.329           & 0.675   & 21.22     & 0.2742 & 0.1954        \\ \bottomrule
\end{tabular}

\begin{tablenotes}[flushleft] %\small %\footnotesize
    \item[*] There is no other baseline that performs layout-guided multi-HOI editing task, thus we report only ours.
\end{tablenotes}
\end{threeparttable} % three-part-table
 }
\vspace{-15pt} % 10
\end{table*}
\begin{table}[t]
\caption{Quantitative comparison for \textbf{HOI generation} task. Our method outperforms leading layout-conditioned and HOI-aware models on both controllability and image quality metrics.}
\label{tab:hoigen}
\vspace{-10pt} % 5
\centering
\resizebox{\columnwidth}{!}{%
\begin{tabular}{@{}lccccS[table-format=-1.4]@{}}
\toprule
\multirow{2}{*}{Method} & \multicolumn{2}{c}{Controllability} & \multicolumn{3}{c}{Image Quality} \\ \cmidrule(l){2-3} \cmidrule(l){4-6} 
                        & Spatial          & HOI              & PickScore      & HPS             & {ImageReward}   \\ \midrule
GLIGEN \cite{gligen2023}                 & 0.5150           & 0.3344           & 20.46          & 0.2322          & -0.4103         \\
InstanceDiffusion \cite{wang2024instancediffusion}       & 0.5228           & 0.3476           & 20.06          & 0.2312          & -0.2532         \\
MIGC++\cite{zhou2024migc,zhou2024migc++} & 0.5331           & 0.3616           & 20.16          & 0.2208          & -0.6492         \\
Eligen \cite{zhang2025eligen}     & 0.4371           & 0.3061           & {\ul 21.28}    & {\ul 0.2496}    & \uline{0.3921}    \\
InteractDiffusion \cite{hoe2024interactdiffusion}   & {\ul 0.5768}     & {\ul 0.4505}     & 20.37          & 0.2283          & -0.3194         \\ \midrule
Ours                    & \textbf{0.6104}  & \textbf{0.4528}  & \textbf{21.41} & \textbf{0.2617} & \textbf{0.5224} \\
Improvements            & 5.8\%            & 0.5\%            & 0.6\%          & 4.8\%           & {33.2\%}        \\ \bottomrule
\end{tabular}%
}
\vspace{-16pt} % 16 (12 in review version)
\end{table}

We implement \paper by adapting the MM-DiT backbone from Flux.1 Kontext \cite{labs2025flux1kontext}. We train using LoRA \cite{hu2022lora} for 10K steps with a batch size of 16 using the AdamW \cite{adam2014} optimizer (8-bit). More details are provided in \cref{supp-sec:implementation-details}, see \cref{supp-sec:human-study} for human preference study.
% 361M / 343M, 12B, 2.5% of Flux

\subsection{Unified Training Strategy}
To enable a single model for both generation and editing under diverse conditions, we employ a joint training strategy with modality dropout. Batches alternate between  \textbf{generation} and \textbf{editing} and we optimize with the standard diffusion flow-matching objective \cite{lipman2023flowmatching}. During training, we randomly drop input modalities: layout (bounding boxes $b^r_n$) with probability $p_{\text{layout}}=0.25$, HOI labels ($\langle s,o,a\rangle_n$ replaced by object-only) with $p_{\text{hoi}}=0.25$, and the global text prompt $\mathcal{P}$ with $p_{\text{txt}}=0.30$, ensuring at least one modality remains. The attention masking (\cref{subsec:attention}) is applied consistently, defaulting to unconstrained attention for dropped layouts. This ensures the model operates robustly across various tasks and input combinations.

% \noindent\textbf{Masked attention during training.}
% Regardless of batch type, we apply the same attention mask from \cref{subsec:attention}. When boxes are present, subject/object/action tokens may attend only to their designated regions. When boxes are dropped, the corresponding HOI$\leftrightarrow$image connections default to all-pass to image tokens (no $\infty$ entries).

\subsection{Datasets}
\noindent\textbf{HOI-Edit-44K (ours).}
To address the lack of paired data for HOI editing, we constructed a large-scale dataset, HOI-Edit-44K, which we will release publicly. We collect source images with verified HOIs from two streams: (i) \textbf{Flux.1 generations} that realise a verified source interaction, and (ii) \textbf{HICO-DET images}. For each source image, we synthesise potential single-HOI edits using Flux.1 Kontext~\cite{labs2025flux1kontext} and InteractEdit~\cite{hoe2025interactedit}. A (source, edited) image pair is retained only upon passing two rigorous automated checks:
\begin{itemize}
\item \textbf{HOI correctness.} We run PViC~\cite{zhang2023pvic} HOI detector on the edited image and require the predicted HOI to match the target HOI. The detected layout are recorded for the pair.
\item \textbf{Identity preservation.} We extract DINOv2 features~\cite{oquab2023dinov2} from subject and object crops in both source and edited images and keep the pair only if both cosine similarities exceed a threshold of 0.75.
\end{itemize}
This stringent filtering process discarded approximately 90\% of initial candidates, primarily due to incorrect interactions or identity drift. The final dataset comprises 44K high-quality HOI editing pairs, each including the source images, target interaction triplet, edited image, and corresponding layout. This provides diverse, identity-preserving interaction edits at scale, crucial for training our unified model. See \cref{supp-subsec:hoiedit44k} for more details and generalization. 

\noindent\textbf{SA-1B~\cite{kirillov2023segmentanything}.} We sample 35K images and derive layouts from object masks \cite{lian2023llmgrounded}, providing \emph{object-only layout} supervision (no HOI) that strengthens spatial layout control.

\noindent\textbf{HICO-DET~\cite{hicodet2018}.} We use 37K training images to learn HOI generation priors. The test set is used only for evaluation.

% \noindent\textbf{Notes.}
% For Flux.1–sourced inputs, we keep a source image only if an HOI detector confirms that its interaction matches the intended prompt; otherwise it is discarded. All boxes stored in our dataset come from detector outputs to ensure consistency across sources. \todo{Dataset statistics} (verb/object distributions, box areas, multi-interaction frequency).

\begin{figure*}
\centering
\setlength{\tabcolsep}{1pt} % General space between cols (6pt standard)
\renewcommand{\arraystretch}{1} % General space between rows (1 standard) 3.6 / 3.3 
\resizebox{0.9\linewidth}{!}{%
\begin{tabular}{cccccc}
Source & HOIEdit & Qwen Image Edit & Flux.1 Kontext & InteractEdit & Ours\\
\includegraphics[width=.160\linewidth]{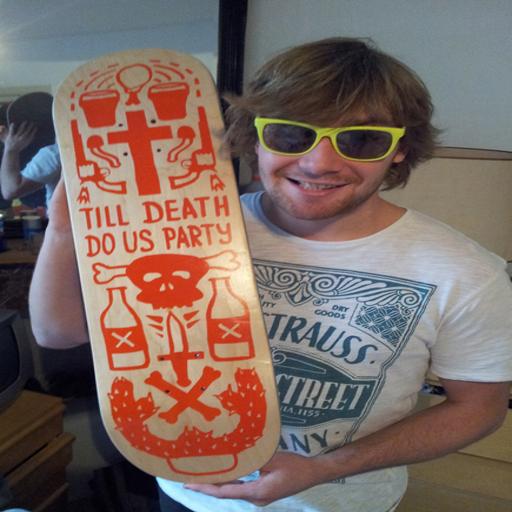}
& \includegraphics[width=.160\linewidth]{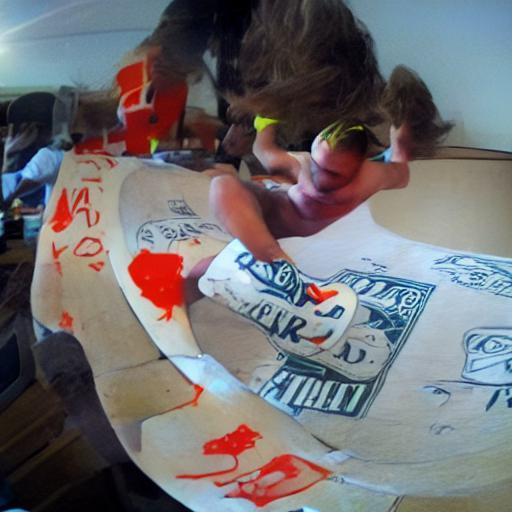}
& \includegraphics[width=.160\linewidth]{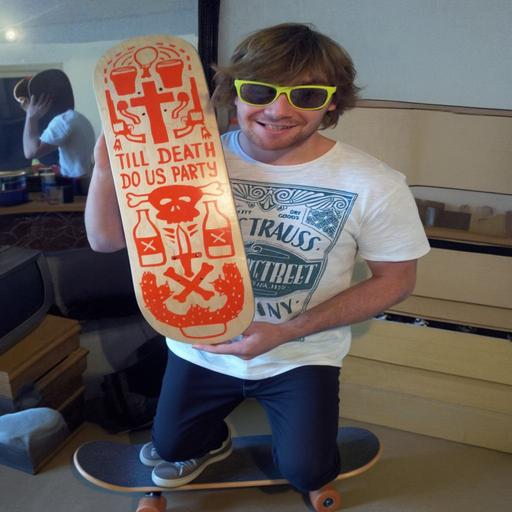}
& \includegraphics[width=.160\linewidth]{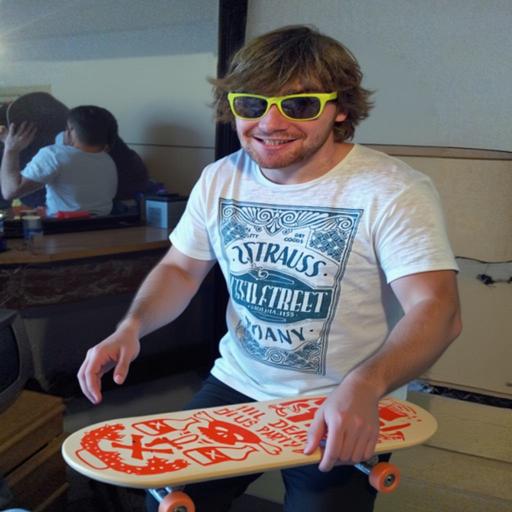}
& \includegraphics[width=.160\linewidth]{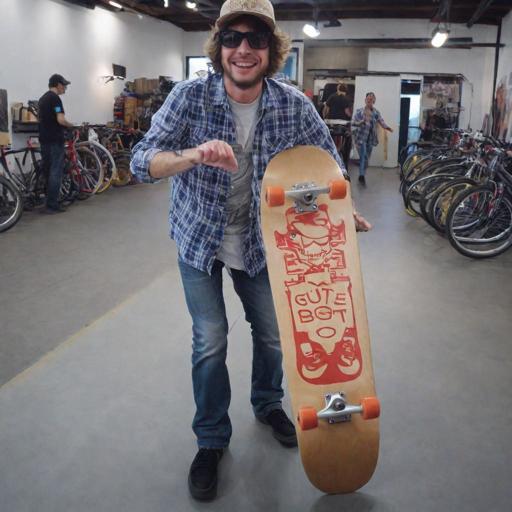}
& \includegraphics[width=.160\linewidth]{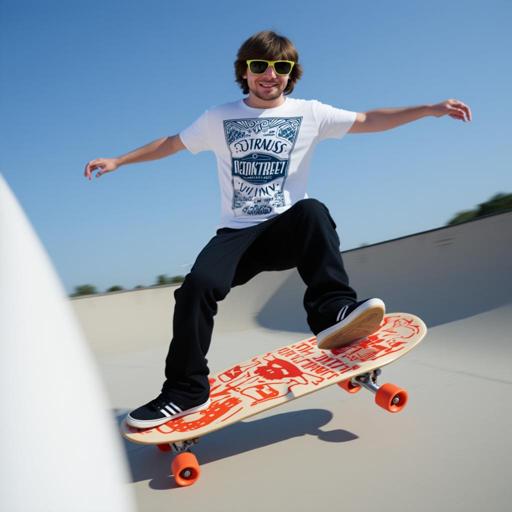}\\
\multicolumn{6}{c}{hold \textrightarrow ride skateboard}\\
\includegraphics[width=.160\linewidth]{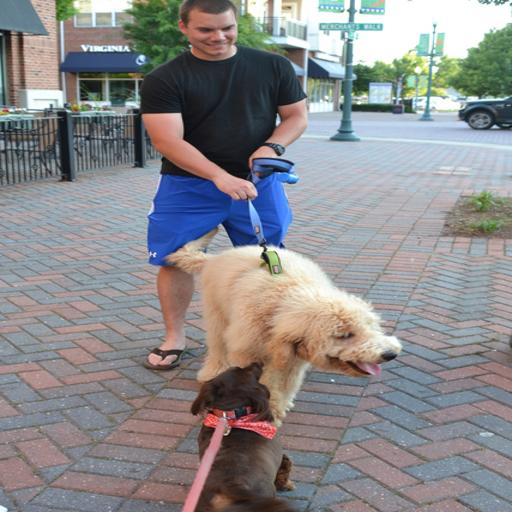}
& \includegraphics[width=.160\linewidth]{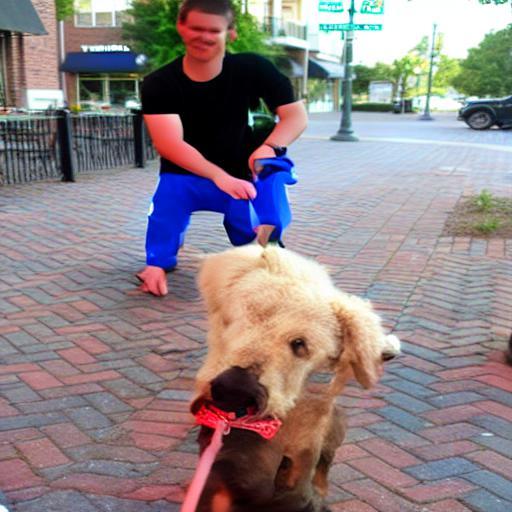}
& \includegraphics[width=.160\linewidth]{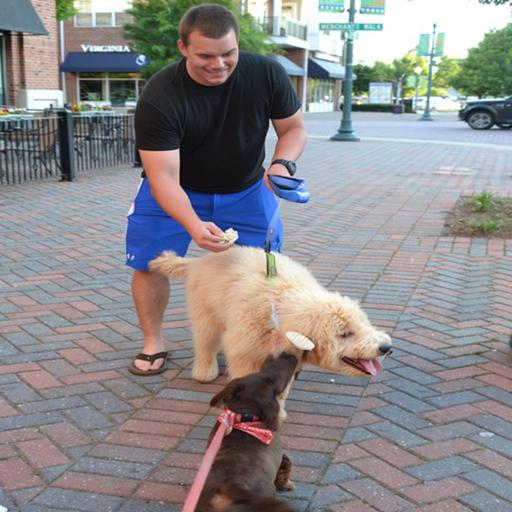}
& \includegraphics[width=.160\linewidth]{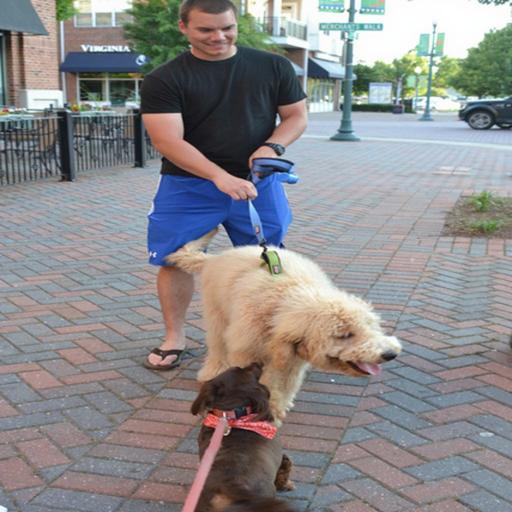}
& \includegraphics[width=.160\linewidth]{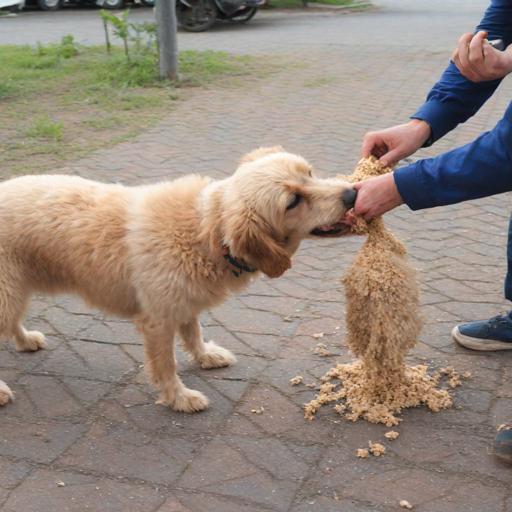}
& \includegraphics[width=.160\linewidth]{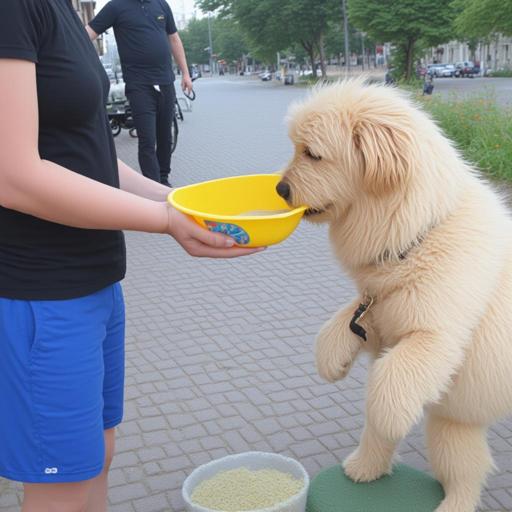} \\
\multicolumn{6}{c}{walk \textrightarrow feed dog}
\end{tabular}
}
\vspace{-11pt}
\caption{Qualitative comparison for layout-free HOI editing. Our method successfully renders the new interaction while preserving identity. In contrast, baseline methods often produce artifacts, fail to change the pose, or lose the subject's identity.}
\label{fig:editing}
\vspace{-8pt} % 5
\end{figure*}
\begin{figure*}
\centering
\setlength{\tabcolsep}{1pt} % General space between cols (6pt standard)
\renewcommand{\arraystretch}{1} % General space between rows (1 standard) 3.6 / 3.3 
\resizebox{0.9\linewidth}{!}{%
\begin{tabular}{ccccccc}
 & \multicolumn{4}{c}{Object-level methods} & \multicolumn{2}{c}{HOI-level methods} \\ \cmidrule(l){2-5} \cmidrule(l){6-7}
HOI Layout & GLIGEN & MIGC & InstanceDiff & Eligen & InteractDiff & Ours\\
\frame{\includegraphics[width=.138\linewidth]{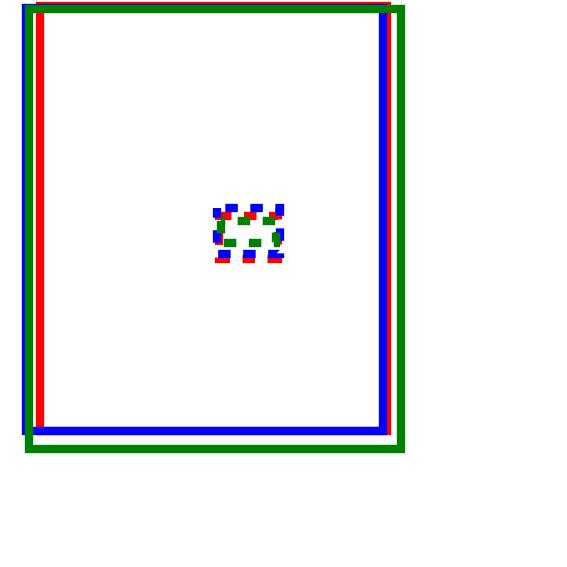}}
& \includegraphics[width=.138\linewidth]{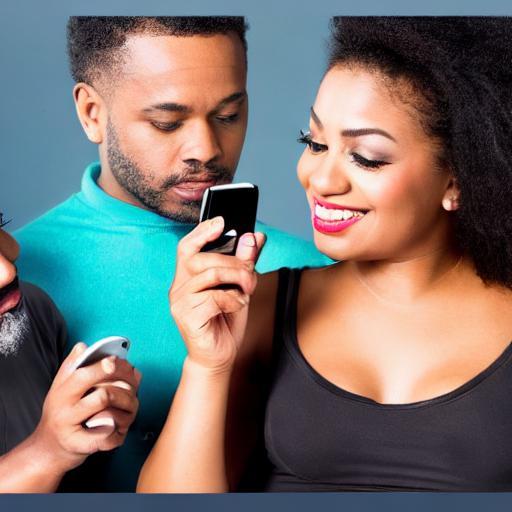}
& \includegraphics[width=.138\linewidth]{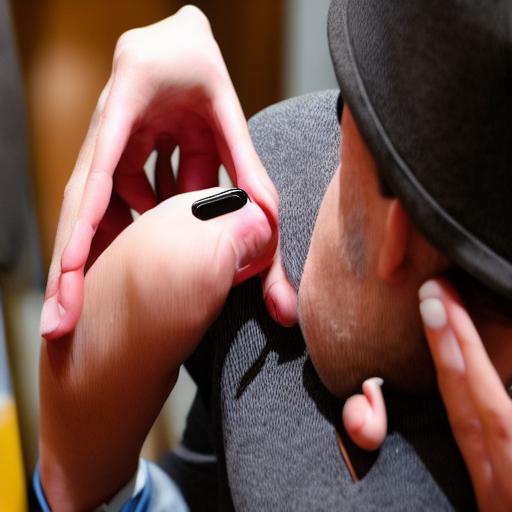}
& \includegraphics[width=.138\linewidth]{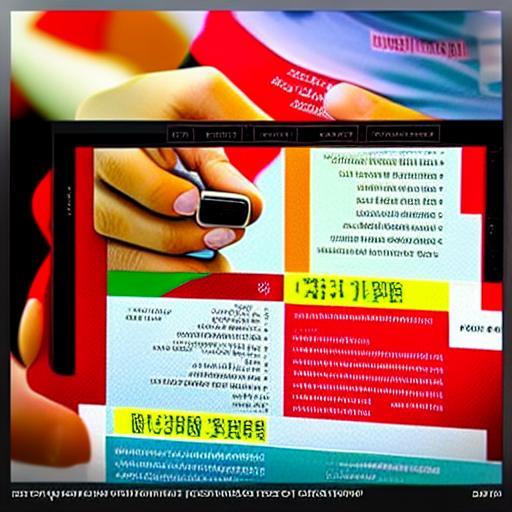}
& \includegraphics[width=.138\linewidth]{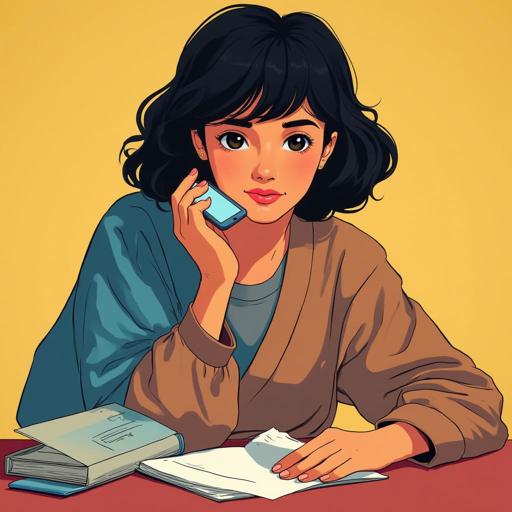}
& \includegraphics[width=.138\linewidth]{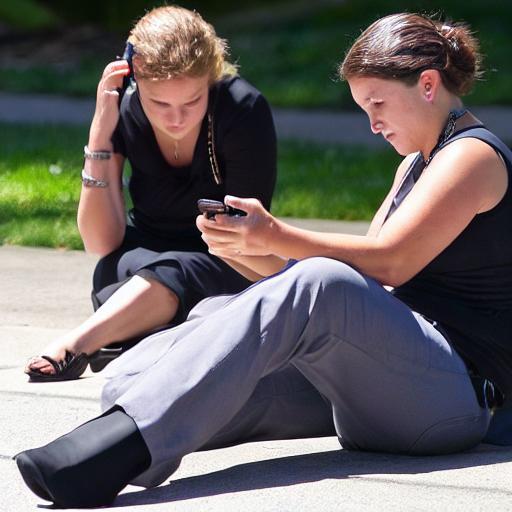}
& \includegraphics[width=.138\linewidth]{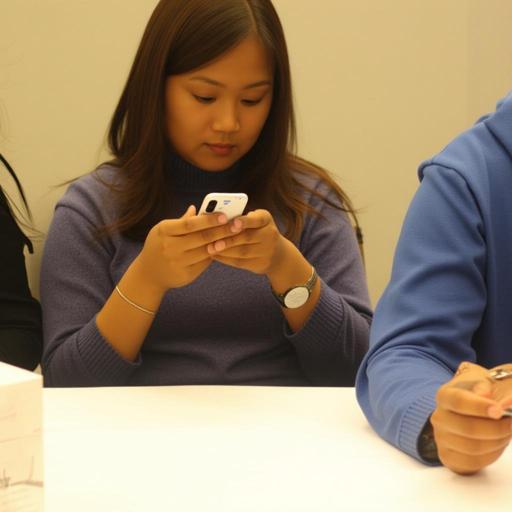}\\
\multicolumn{7}{c}{A person is \textcolor{red}{holding}, \textcolor{blue}{reading}, and \textcolor{OliveGreen}{texting} on a cell phone}\\
\frame{\includegraphics[width=.138\linewidth]{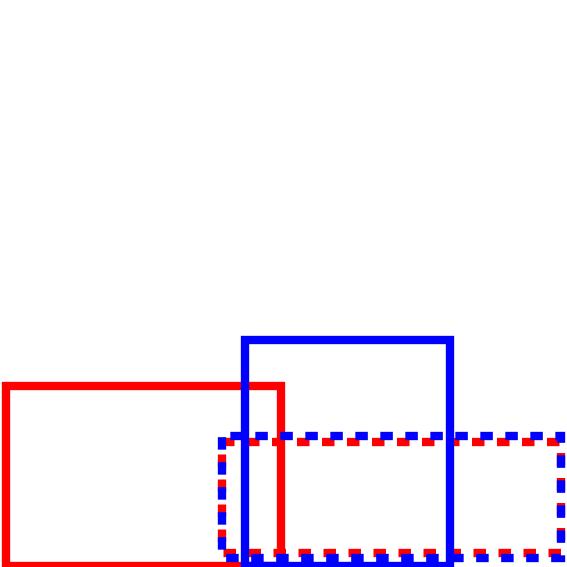}}
& \includegraphics[width=.138\linewidth]{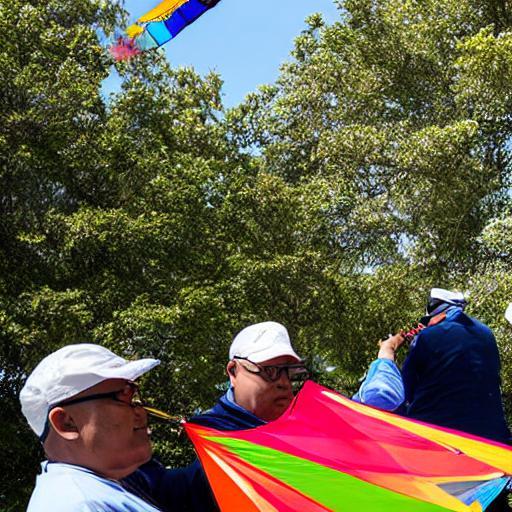}
& \includegraphics[width=.138\linewidth]{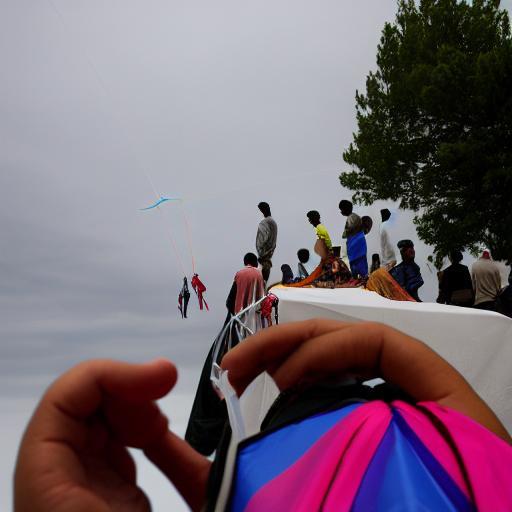}
& \includegraphics[width=.138\linewidth]{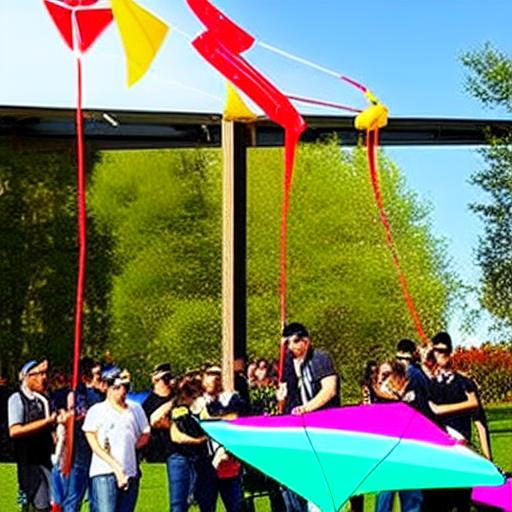}
& \includegraphics[width=.138\linewidth]{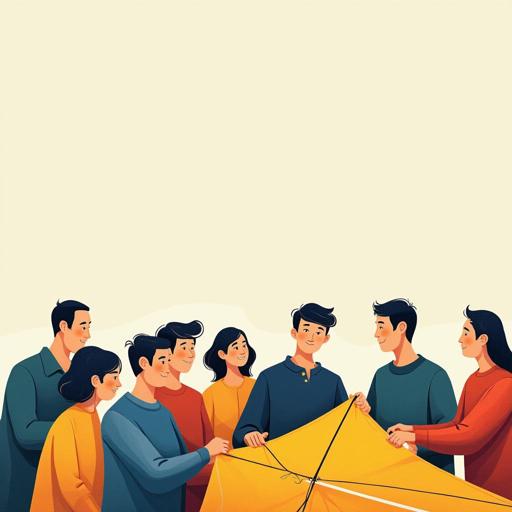}
& \includegraphics[width=.138\linewidth]{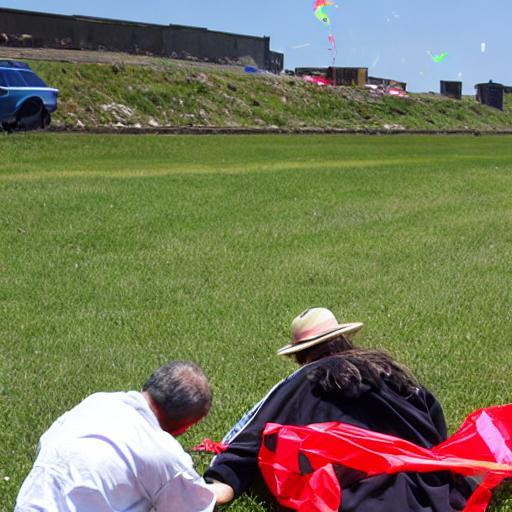}
& \includegraphics[width=.138\linewidth]{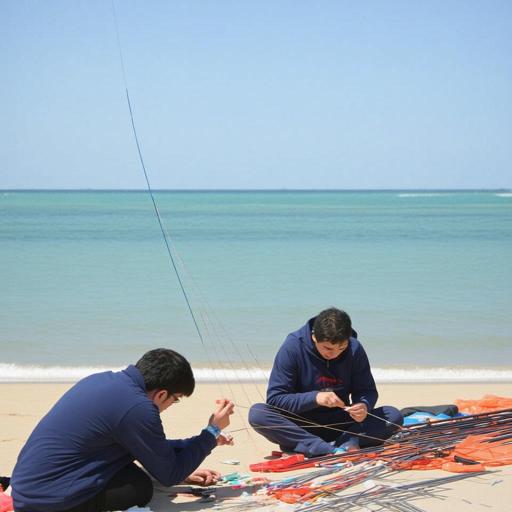} \\
\multicolumn{7}{c}{Two persons are assembling a kite together}
\end{tabular}
}
\vspace{-10pt}
\caption{Qualitative comparison for HOI generation. While object-level methods correctly place entities, they fail to synthesise specified interactions. Ours renders semantically and geometrically consistent interactions, demonstrating a deeper relational understanding.}
\label{fig:generation}
\vspace{-14pt} % 8pt
\end{figure*}

\subsection{Metrics}
\noindent\textbf{Image Quality.} We report standard human-preference aligned perceptual metrics: \textbf{PickScore}\cite{kirstain2023pickscore}, \textbf{HPSv2} \cite{wu2023hpsv2} (Human Preference Score), and \textbf{ImageReward}\cite{xu2023imagereward}. Higher scores indicate better quality and prompt alignment.

% \textbf{PickScore} \cite{kirstain2023pickscore}, a learned image-text preference model; \textbf{HPSv2} \cite{wu2023hpsv2} (Human Preference Score), a proxy model trained on human comparisons; and \textbf{ImageReward} \cite{xu2023imagereward}, a reward model for text-to-image quality.

\noindent\textbf{HOI Editing.} As to \cite{hoe2025interactedit}, we report \textbf{HOI Editability} (success of the target verb–object being realised, as detected) and \textbf{Editability–Identity} (a composite that balances HOI success with ID preservation). See Appen.~\ref{supp-subsec:iebench} for details.

\noindent\textbf{Spatial Score.} 
For layout-guided tasks, we run PViC \cite{zhang2023pvic} to detect subject and object instances for each target triplet. We compute the mean IoU between the target boxes ($b^s, b^o$) and the best-matching detected boxes ($\hat{b}^s, \hat{b}^o$), defined as $\mathrm{mIoU}=\tfrac12\big(\mathrm{IoU}(b^s,\hat b^s)+\mathrm{IoU}(b^o,\hat b^o)\big)$. Results are averaged over all targets, higher means better spatial alignment.

\noindent\textbf{HOI Accuracy.} Using PViC, a success is recorded when the target HOI is detected within their specified regions. We report the mean success rate across targets (higher is better).

\subsection{Tasks and Evaluations}
We evaluate three HOI tasks differ by available controls:

\noindent\textbf{Layout-free HOI editing.} 
Modifying interactions in an image using only HOI triplets (no layout), while preserving identity and image quality.
We generate 1000 samples for 100 target edits in IEBench \cite{hoe2025interactedit} and report Editability–Identity, HOI Editability, and image quality metrics (PickScore, HPS, ImageReward).

\noindent\textbf{Layout-guided HOI editing.} 
Modifying interactions in an image using HOI triplets and target layouts.
With layout guidance, it enable editing multiple HOIs at once, which was challenging due to limited expressibility in natural language. For single-HOI edits, we use IEBench with synthesised target layouts, detailed in \cref{subsec:iebench_layout}. For Multi-HOI edits, we \textit{propose a new MultiHOIEdit benchmark} (detailed in \cref{subsec:multihoiedit}), comprises of 200 target edits spanning 2–3 interactions per image, where we generates total 1000 samples for evaluation. In addition to the layout-free metrics, we also report Spatial Score.

\noindent\textbf{HOI generation.} Synthesising images from HOI triplets and layouts. We evaluate on 2000 HICO-DET test targets and report HOI accuracy, Spatial score, and image quality.
% We evaluate 2000 HOI targets sampled from HICO-DET test set and report HOI accuracy, Spatial score, and image quality metrics.
% (Detector mAP tables are provided for completeness in the appendix; our primary controllability metrics are mIoU and IA.)

\subsection{Quantitative Results}

\noindent\textbf{Layout-free HOI editing.} \Cref{tab:hoiedit} compares our method with recent editing baselines. We achieve the best Editability–Identity (0.638) and HOI Editability (0.596), improving over the strongest priors by +10.0\% and +16.0\%, respectively, while also attaining the best HPS, ImageReward and PickScore. These results indicate that, even without layout input, our unified formulation reliably edits the interaction while maintaining subject identity intact.

\noindent\textbf{Layout-guided HOI editing.} \Cref{tab:layout-hoiedit} reports single- and multi-HOI edits with layout guidance. For single-HOI, we establish a baseline by adapting InteractEdit\cite{hoe2025interactedit} and InteractDiffusion \cite{hoe2024interactdiffusion} (see \cref{supp-subsec:intedit+intdiff}). Our method achieves a high Spatial score (0.822), strong HOI Editability (0.570), and good perceptual quality. For much harder multi-HOI (2–3 HOIs across 1-3 persons), Spatial remains strong (0.675) and quality scores are maintained.
% for layout-conditioned editing
% The modest drop from single to multi indicates robust instance separation with limited cross-interaction interference.

\noindent\textbf{HOI generation.} \Cref{tab:hoigen} reports controllability and perceptual quality. Our method slightly surpasses \cite{hoe2024interactdiffusion} on Spatial and HOI accuracy, while also achieving the best perceptual scores, PickScore 21.41 (+0.7\%), HPS 0.2617 (+4.8\%) and ImageReward 0.5524 (+33.2\%) over the strongest prior. Thus, unifying editing and generation does not compromise HOI generation; instead, it improves it.

\subsection{Qualitative Results}
\cref{fig:editing} compares \textbf{layout-free HOI editing}. HOIEdit \cite{xu2025hoiedit} often corrupts the image. For \emph{hold→ride skateboard}, Qwen leaves the pose essentially unchanged and \cite{hoe2025interactedit} drifts in identity; others have an incorrect riding stance. Contrary, \paper renders the intended interaction while preserving identity. This stems from two separate factors: (i) HOI semantics learned during generation (contact patterns, verb–object geometry) transfer to editing, and (ii) structured HOI attention steers the edit to the correct roles/regions. Baselines without such HOI knowledge tend to keep poses unchanged or misrender contact.%For \emph{walk→feed dog}, Qwen/Flux mostly keep the original scene, while \cite{hoe2025interactedit} only partially realises the feeding interaction. In both cases, \paper renders the intended interaction while preserving identity. This stems from two separate factors: (i) HOI semantics learned during generation (contact patterns, verb–object geometry) transfer to editing, and (ii) structured HOI attention steers the edit to the correct roles/regions. Baselines without such HOI knowledge tend to keep poses unchanged or misrender contact.

\cref{fig:generation} compares \textbf{HOI generation}. \textit{Object-level methods} (GLIGEN, MIGC, InstanceDiff, Eligen) correctly place entities but rarely realise the relations: not texting on the phone. For \textit{HOI-level}, \cite{hoe2024interactdiffusion} improves relation plausibility but often produces less convincing, semantically off interactions. Our model, \paper yields superior semantic faithfulness, \eg, hands grasp the phone for `holding/reading/texting,'. We attribute these gains to: (i) \textbf{HOI tokens} that encode the interaction semantics, (ii) \textbf{structured HOI attention} that constrains HOI tokens to their regions while models the relation, and (iii) \textbf{HOI RoPE} that separates instances to avoid mix-ups. This yields spatially compliant and semantically faithful multi-HOI scenes. %and not assembling the kite. For \textit{HOI-level}, \cite{hoe2024interactdiffusion} improves relation plausibility but often produces less convincing, semantically off interactions. Our model, \paper yields superior semantic faithfulness, \eg, hands grasp the phone for `holding/reading/texting,' and `assembling a kite together' produces coordinated poses with clear instance separation and minimal spillover. We attribute these gains to: (i) \textbf{HOI tokens} that encode the interaction semantics, (ii) \textbf{structured HOI attention} that constrains HOI tokens to their regions while models the relation, and (iii) \textbf{HOI RoPE} that separates instances to avoid mix-ups. This yields spatially compliant and semantically faithful multi-HOI scenes.

\begin{figure}
\centering
\setlength{\tabcolsep}{1pt} % General space between cols (6pt standard)
\renewcommand{\arraystretch}{1} % General space between rows (1 standard) 3.6 / 3.3 
\begin{tabular}{cccc}
\includegraphics[width=.240\linewidth]{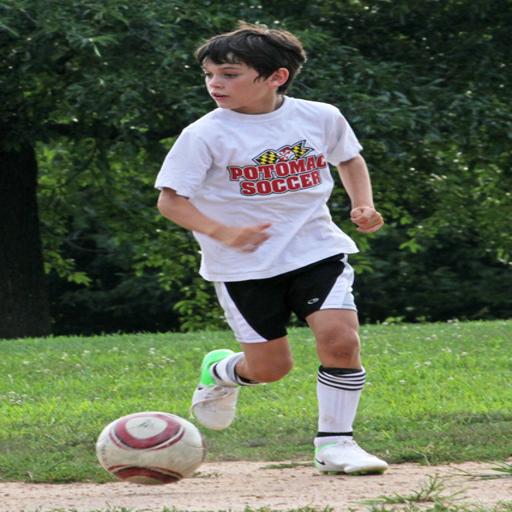}
& \includegraphics[width=.240\linewidth]{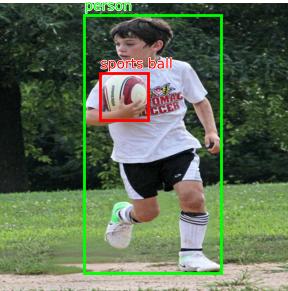}
& \includegraphics[width=.240\linewidth]{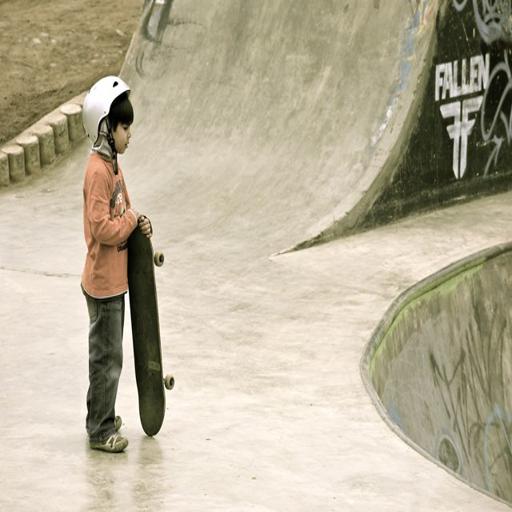}
& \includegraphics[width=.240\linewidth]{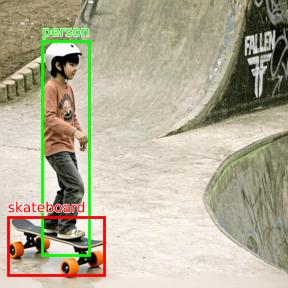}\\[-4pt]
\multicolumn{2}{c}{\small kick{\textrightarrow}hold ball}&
\multicolumn{2}{c}{\small hold{\textrightarrow}ride skateboard}\\ [-2pt]
\includegraphics[width=.240\linewidth]{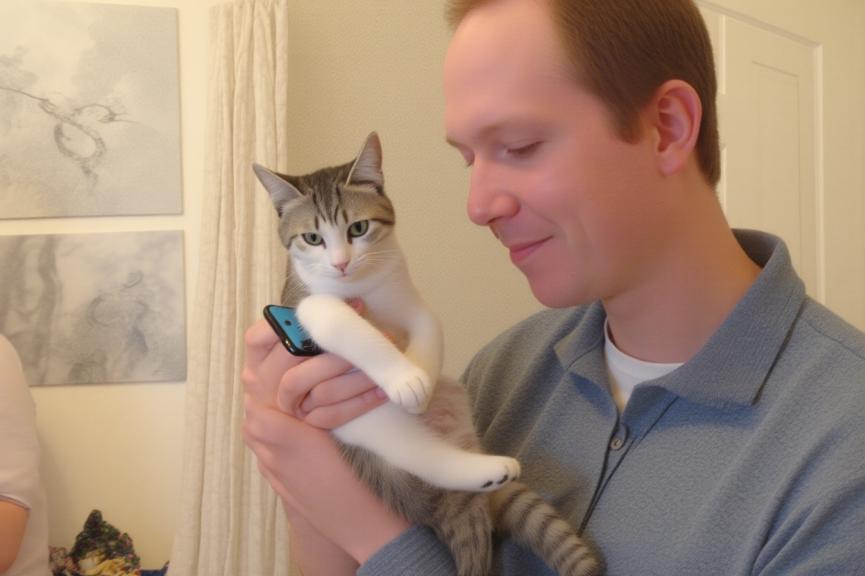}
& \includegraphics[width=.240\linewidth]{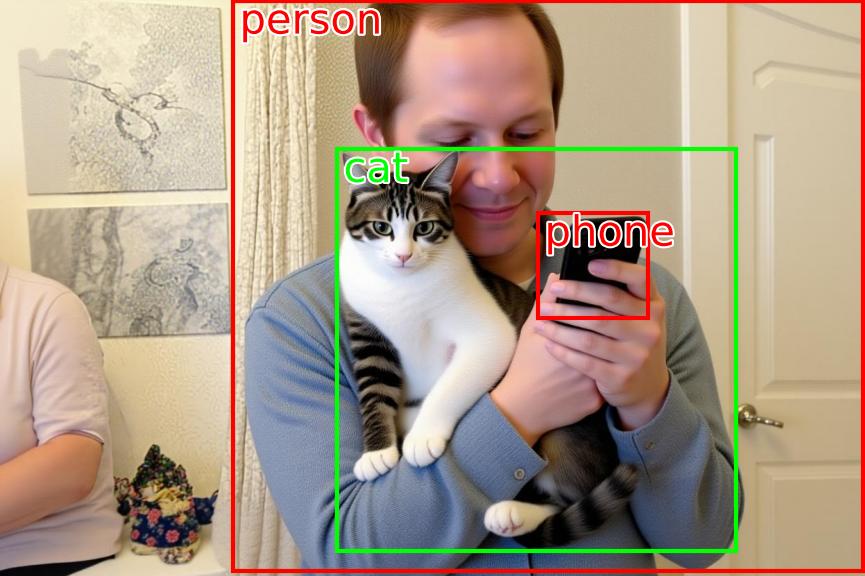}
& \includegraphics[width=.240\linewidth]{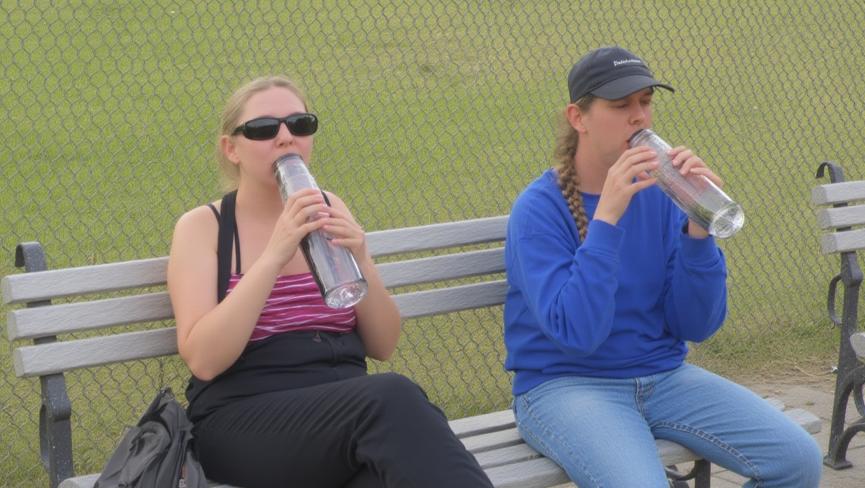}
& \includegraphics[width=.240\linewidth]{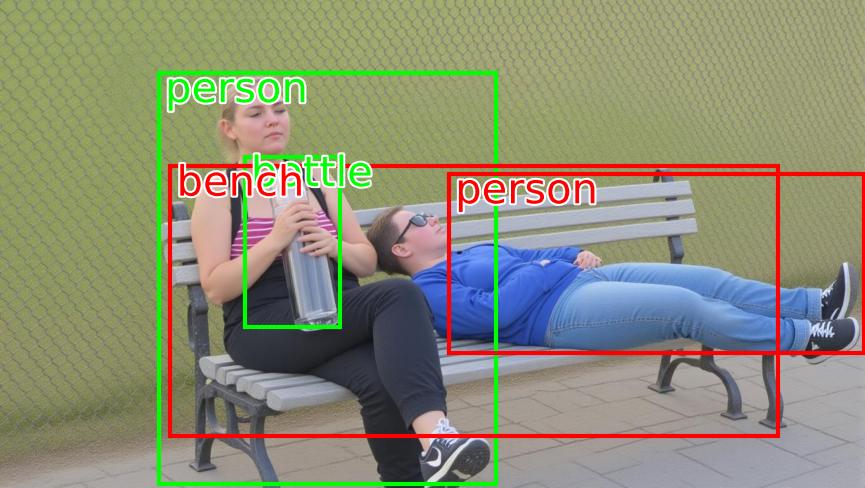} \\[-5pt]
\multicolumn{2}{c}{\small hold{\textrightarrow}hug cat}&
\multicolumn{2}{c}{\small drink with{\textrightarrow}carry bottle}\\[-4pt]
\multicolumn{2}{c}{\small hold{\textrightarrow}text on phone}&
\multicolumn{2}{c}{\small sit on{\textrightarrow}lie on bench}
\end{tabular}
\vspace{-10pt} % 10
\caption{Layout-guided editing examples. Our model supports single-HOI (top) and multi-HOI edits (bottom), limiting changes to target layouts while preserving scene consistency.}
\label{fig:layout-editing}
\vspace{-8pt} % 10
\end{figure}

\cref{fig:layout-editing} shows \textbf{layout-guided HOI edits}. For single-HOI scene: the edits are confined to the layout. The ball is firmly grasped, and the person shifts into a riding pose on the skateboard, while their identity and background remain intact. For multi-HOI scene, natural language alone is too ambiguous to specify multiple edits; layout resolves this. Our model simultaneously executes \emph{drink with{→}carry bottle} and \emph{sit on{→}lie on bench}, updating each person only within their regions. One holding the bottle and the other reclining on the bench, without spillover or mix-ups. This stems from joint training with multi-HOI generation, which teaches to compose and disentangle interactions. Combined with HOI attention and HOI RoPE, this enables reliable multi-HOI edits even without multi-HOI edit training pairs.

\begin{figure}
\centering
\setlength{\tabcolsep}{1pt} % General space between cols (6pt standard)
\renewcommand{\arraystretch}{1} % General space between rows (1 standard) 3.6 / 3.3 
\begin{tabular}{cccc}
\frame{\includegraphics[width=.240\linewidth]{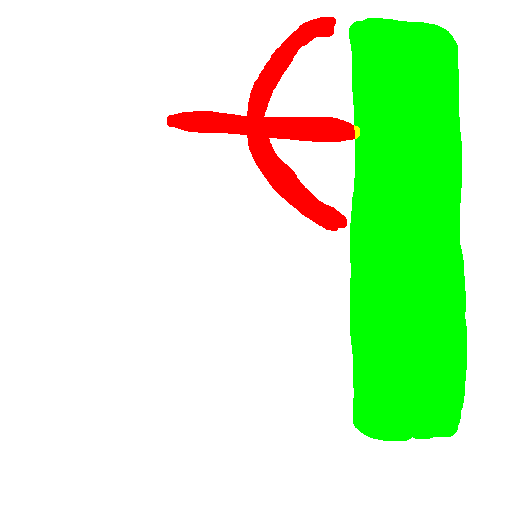}}
& \includegraphics[width=.240\linewidth]{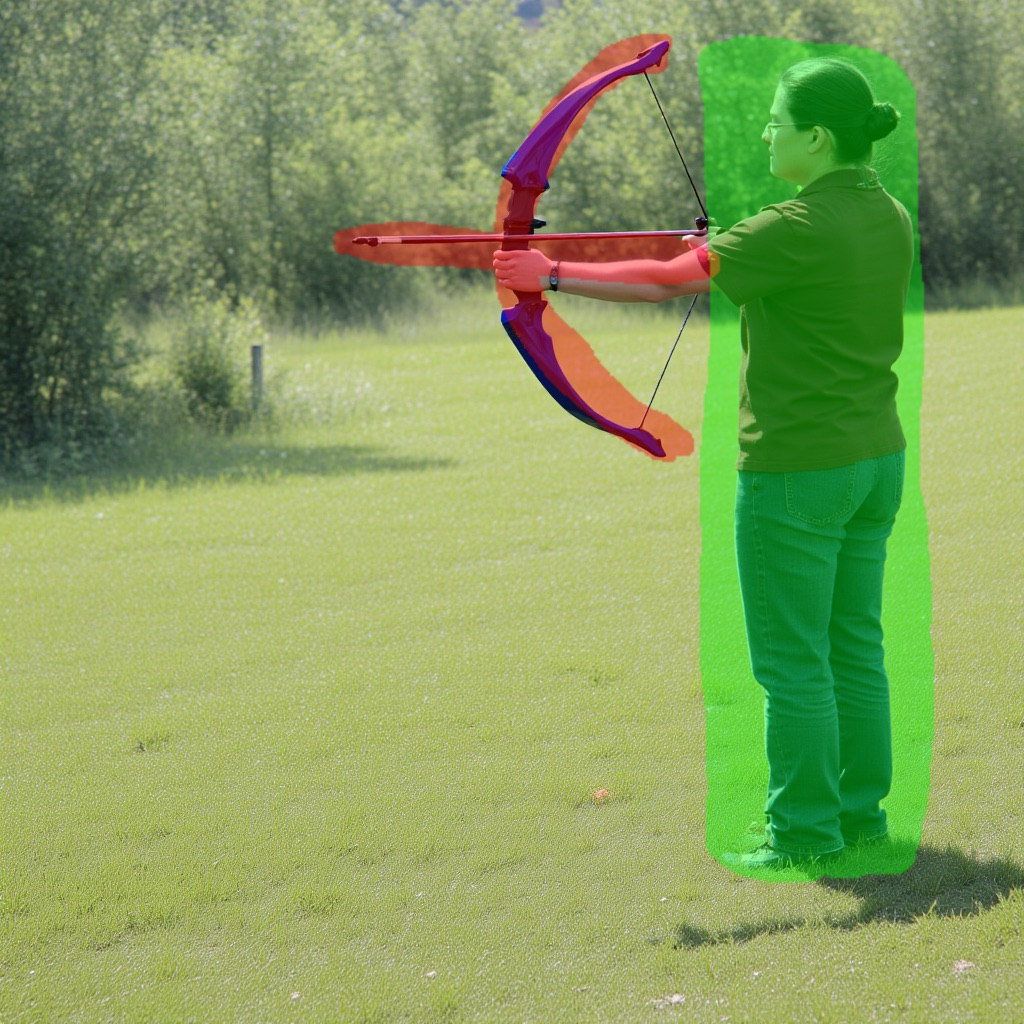} %png
& \frame{\includegraphics[width=.240\linewidth]{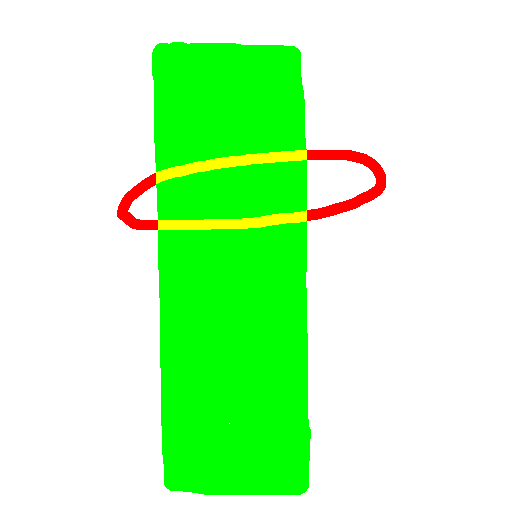}}
& \includegraphics[width=.240\linewidth]{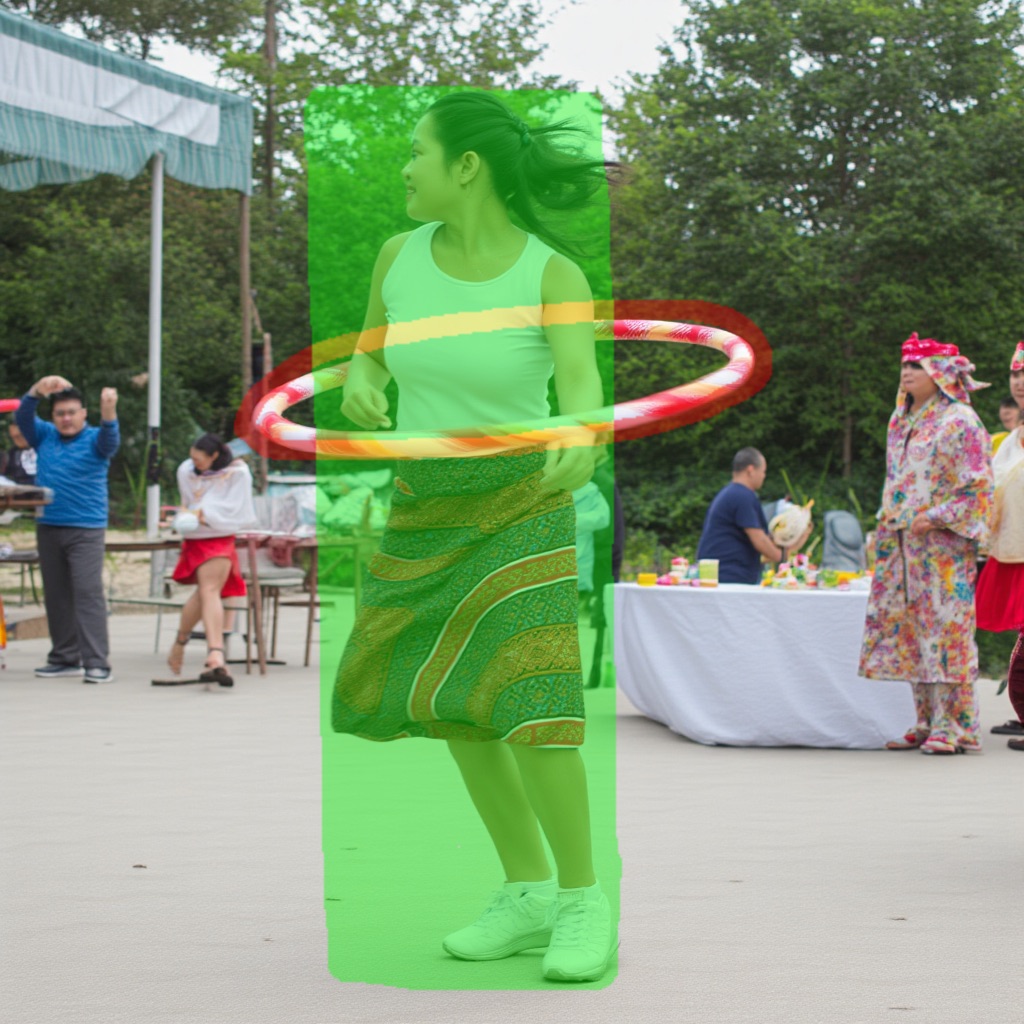}\\ [-3pt] %png
\multicolumn{2}{c}{\small \textlangle\textcolor{ForestGreen}{person} drawing \textcolor{red}{bow}\textrangle} &
\multicolumn{2}{c}{\small\textlangle\textcolor{ForestGreen}{person} spinning \textcolor{red}{hula loop}\textrangle} \\
\frame{\includegraphics[width=.240\linewidth]{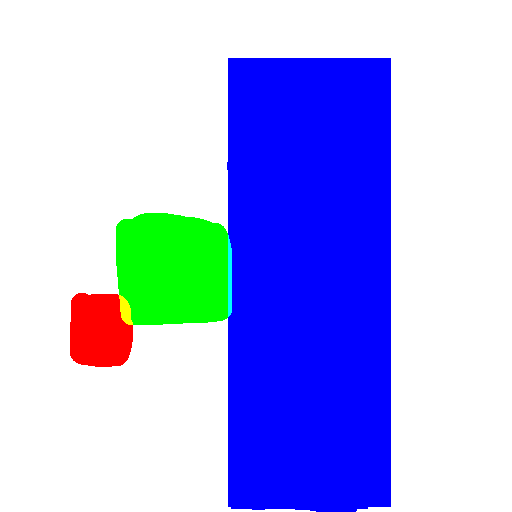}}
& \includegraphics[width=.240\linewidth]{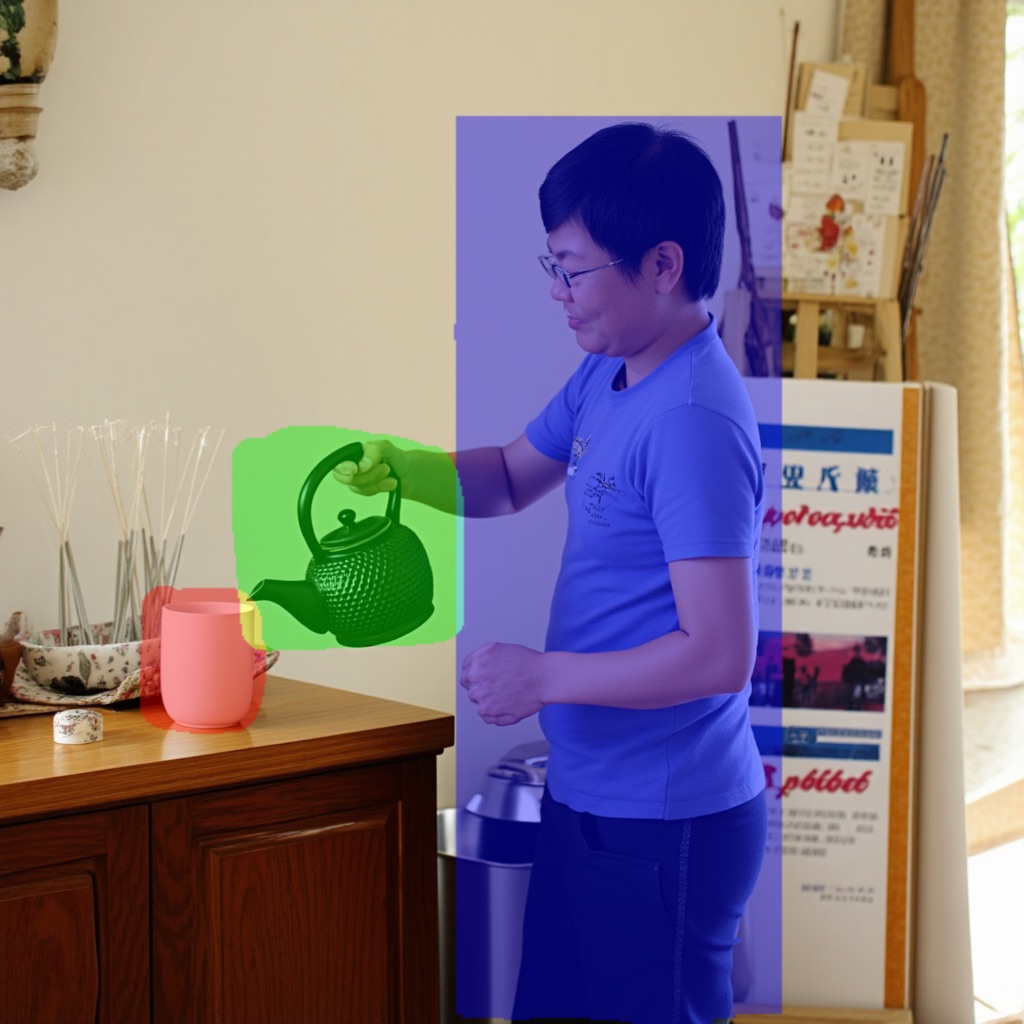}%png
& \frame{\includegraphics[width=.240\linewidth]{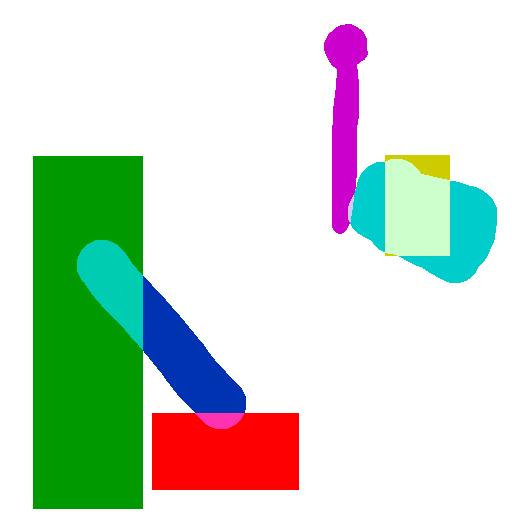}}
& \includegraphics[width=.240\linewidth]{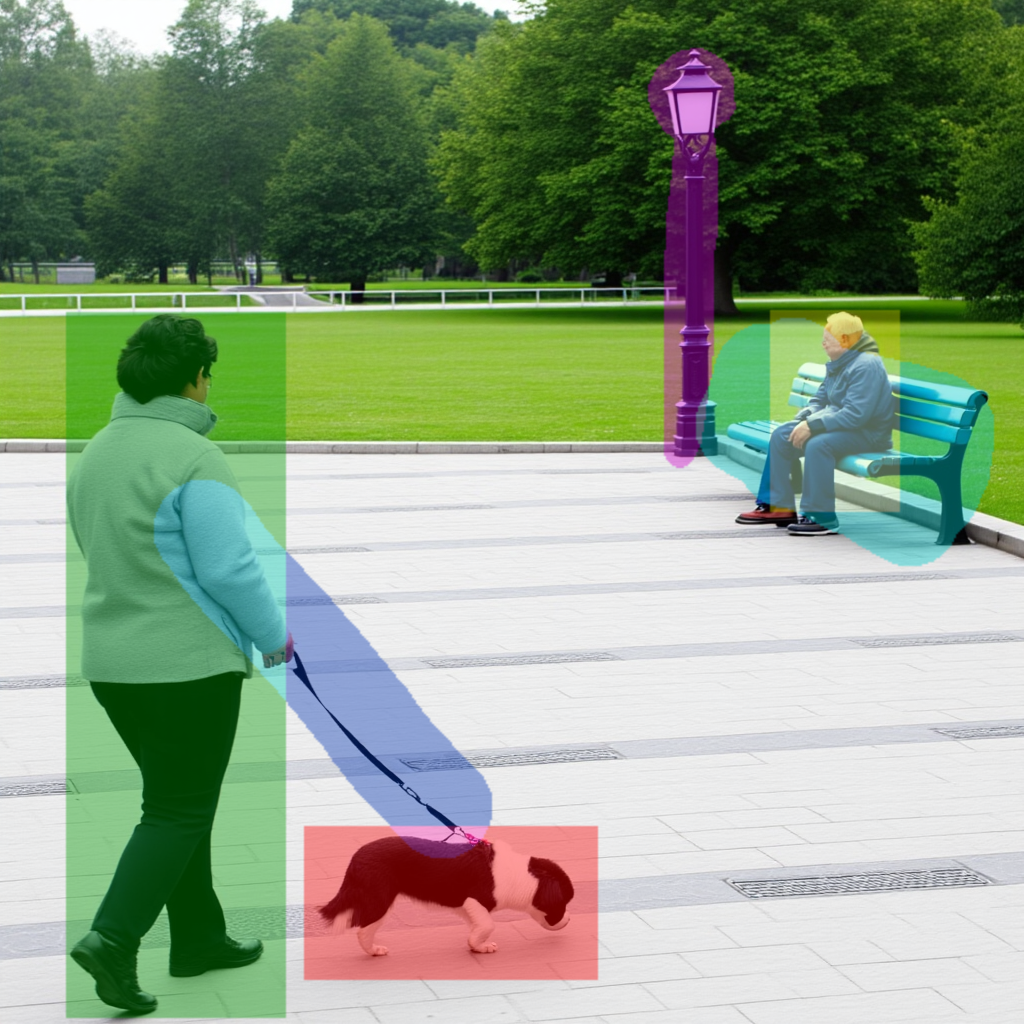} \\[-3pt]
\multicolumn{2}{c}{\small\textlangle\textcolor{blue}{person} pouring \textcolor{ForestGreen}{teapot}\textrangle} &
\multicolumn{2}{c}{\small\textlangle\textcolor{ForestGreen}{person} walking \textcolor{red}{dog}\textrangle} \\[-3pt]
\multicolumn{2}{c}{\small\{\textcolor{red}{cup}\}} &
\multicolumn{2}{c}{\small\textlangle\textcolor{YellowOrange}{person} sitting on \textcolor{cyan}{bench}\textrangle} \\[-3pt]
& & \multicolumn{2}{c}{\small\{\textcolor{Fuchsia}{lamp post}\};\{\textcolor{blue}{leash}\}} \\[-3pt]
% & & \multicolumn{2}{c}{}
\end{tabular}
\vspace{-8pt} % 8
\caption{Versatile control in HOI generation. Our model supports conditioning on both arbitrary-shape masks (top) and a mix of HOI and object-only inputs within a single scene (bottom), demonstrating its compositional capabilities.}
\label{fig:mixed-generation}
\vspace{-16pt} % 10 -- latest: 15
\end{figure}

\Cref{fig:mixed-generation} showcases \textbf{arbitrary-shape masks} and \textbf{mixed-modality control}. Irregular masks (strokes/polygons) provide fine-grained shape control for subject/object regions. We combine layout-guided HOIs and object-only entities, \eg, adding background props with object-only masks while generating foreground interactions. These behaviours stem from modality-dropout training and our layout-aware HOI attention. Overall, the unified interface supports flexible modality combinations in a single generation.

% attribute the gains to the module designs... 


\subsection{Ablation Studies}\label{subsec:ablation}
\begin{figure}[t]
\centering
\setlength{\tabcolsep}{0.5pt} % General space between cols (6pt standard)
\renewcommand{\arraystretch}{1} % General space between rows (1 standard) 3.6 / 3.3 
\begin{tabular}{ccccc}
{\footnotesize Layout} & {\footnotesize (1)} & {\footnotesize (2)} & {\footnotesize (3)} & {\footnotesize (4)}\\
% \includegraphics[width=.190\linewidth]{example-image-duck}
% & \includegraphics[width=.190\linewidth]{example-image-duck}
% & \includegraphics[width=.190\linewidth]{example-image-duck}
% & \includegraphics[width=.190\linewidth]{example-image-duck}
% & \includegraphics[width=.190\linewidth]{example-image-duck}\\
% \multicolumn{5}{c}{HOI Editing: target prompt 1}\\
\includegraphics[width=.193\linewidth]{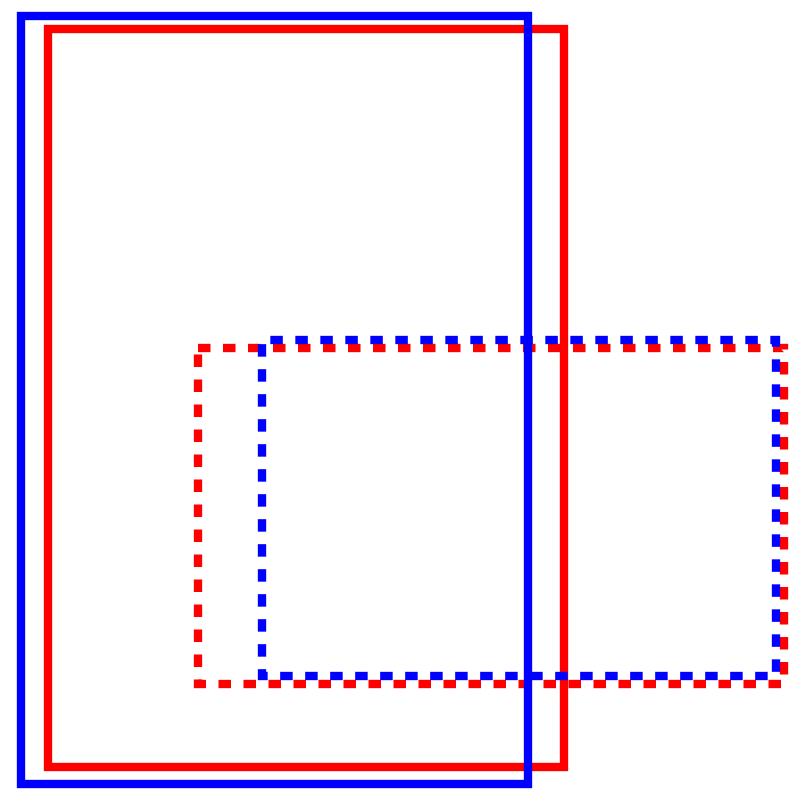}
& \includegraphics[width=.193\linewidth]{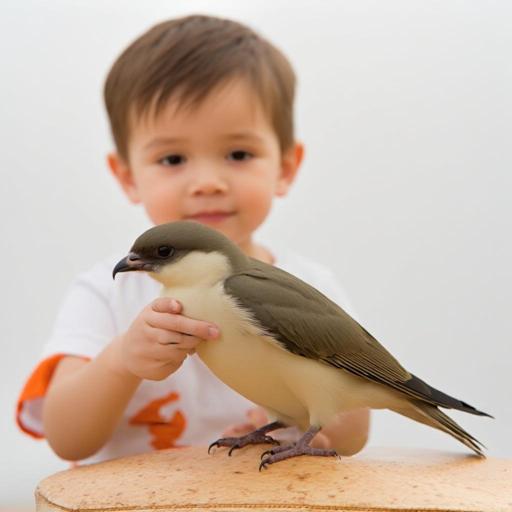}
& \includegraphics[width=.193\linewidth]{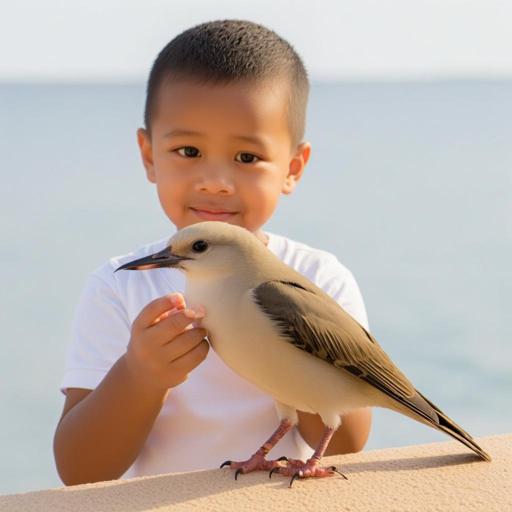}
& \includegraphics[width=.193\linewidth]{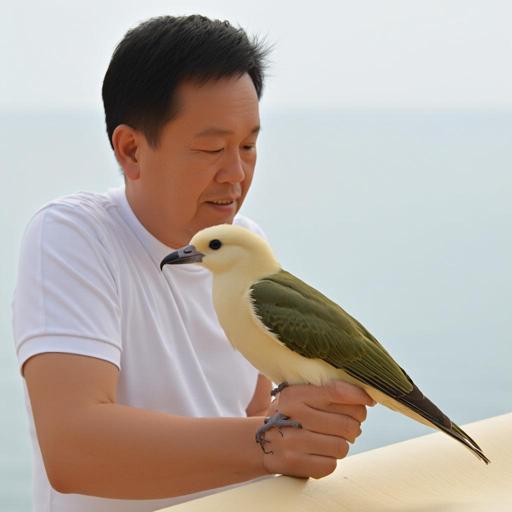}
& \includegraphics[width=.193\linewidth]{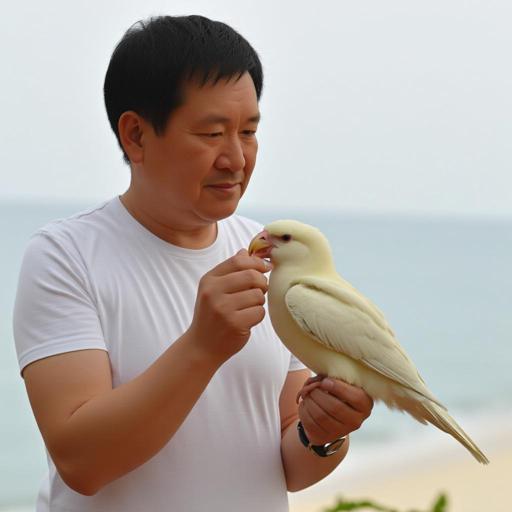} \\[-3pt]
\multicolumn{5}{c}{\footnotesize A person is \textcolor{red}{holding} and \textcolor{blue}{petting} bird}
\end{tabular}
\vspace{-10pt}
\caption{Progressively adding components improves the interaction's plausibility, only the full model (4) successfully rendering the complex, two-handed action of both ``holding'' and ``petting.''}
\label{fig:ablation}
\vspace{-8pt}
\end{figure}

We conduct a comprehensive ablation study to validate the contribution of each component, summarized in \cref{tab:ablation} and visualised in \cref{fig:ablation}. We perform an additive analysis, starting from a strong baseline (BL), which is the Eligen \cite{zhang2025eligen}.
% It provides robust object-level spatial grounding but lacks explicit HOI modelling.

Introducing \textbf{Action Grounding (AG)} establishes a foundational understanding of interactions that the object-level model lacks. This is evident in the large gains across both generation and editing tasks.
Layering on the \textbf{HOI Encoder (Enc)} further improves performance, particularly boosting the perceptual quality (IR) by providing the model with explicit role and instance cues.
The subsequent addition of \textbf{Structured HOI Attention (Attn)} yields another major improvement in correctness metrics (HOI Acc. and EI), confirming its critical role in enforcing the relational structure of the interaction and adhering to layouts. 
Finally, incorporating \textbf{HOI RoPE (HRoPE)} provides the last refinement step by helping to disentangle instance identities, significantly enhancing perceptual quality (IR).

\begin{table}[t]
\caption{
% Ablation study on core components. AG, Enc, Attn,  HRoPE, IR, EI represent Action Grounding, HOI Encoder, HOI Attention, HOI RoPE, ImageReward and Editability-Identity scores.
Ablation study on core components. AG: Action Grounding, Enc: HOI Encoder, Attn: HOI Attention, HRoPE: HOI RoPE, EI: Editability-Identity, IR: ImageReward.
}
\label{tab:ablation}
\vspace{-10pt}
\centering
\resizebox{\columnwidth}{!}{%
\begin{tabular}{@{}lcccccccc@{}}
\toprule
\multirow{2}{*}{} & \multicolumn{4}{c}{Components}                    & \multicolumn{2}{c}{HOI Generation} & \multicolumn{2}{c}{Multi-HOI Edit} \\ \cmidrule(l){2-5} \cmidrule(l){6-7} \cmidrule(l){8-9} 
                        & AG         & Enc        & Attn       & HRoPE      & HOI Acc.         & IR              & EI              & IR               \\ \midrule
BL                &            &            &            &            & 0.3061           & 0.3921          & -               & -                \\
(1)               & \checkmark &            &            &            & 0.4138           & 0.3156          & 0.423           & 0.1118           \\
(2)               & \checkmark & \checkmark &            &            & 0.4254           & 0.4602          & 0.422           & 0.1306           \\
(3)               & \checkmark & \checkmark & \checkmark &            & 0.4504           & 0.4861          & 0.433           & 0.1944           \\
\textbf{(4)}      & \checkmark & \checkmark & \checkmark & \checkmark & \textbf{0.4528}  & \textbf{0.5224} & \textbf{0.435}  & \textbf{0.2046}  \\ \bottomrule
\end{tabular}%
}
\vspace{-10pt}
\end{table}

This progressive improvement is visualised in \Cref{fig:ablation} on the multi-action prompt \textit{``A person is holding and petting bird.''} (i) With only Action Grounding (AG), the model renders only a simple \textit{`pet'} action. (ii) Adding HOI Encoder provides explicit role cues, yielding a more plausible \textit{`petting'} pose. (iii) Introducing HOI Attention enables the \textit{`holding'} pose but \textit{`petting'} remains entangled with the \textit{`holding'} gesture. (iv) Adding HRoPE separates the two action concepts and correctly depicts both \textit{`hold'} and \textit{`pet'}. This confirms all components are complementary in \paper for a deep relational understanding. % of complex scenes.

\Cref{supp-sec:ablation-unification} shows our \textbf{unified} model outperforms \textbf{task-specific} ones via a ``synergy effect'', where generative priors enhance editing robustness and vice-versa.
\section{Conclusion}
\label{sec:conclusion}
We introduced \paper, a single DiT-based framework that unifies Human-Object Interaction (HOI) generation and editing by \textbf{explicitly modelling interaction structure}. This is realised through three core components: a dedicated \textbf{HOI Encoder} providing fine-grained role and instance identity, \textbf{Structured HOI Attention} enforcing a verb-mediated relational topology constrained by layout, and \textbf{HOI RoPE} ensuring clear instance separation. Our approach bridges the gap between layout-guided generation and layout-free editing, supports flexible control, and enables, for the first time, the challenging \textbf{multi-HOI editing} task. % in a single image. 
\paper achieves state-of-the-art controllability and perceptual quality, delivering physically plausible interactions across both editing and generation benchmarks. 
By effectively integrating relational structure into DiTs, our work pushes generative models beyond simple entity placement toward synthesising semantically coherent HOI scenes.

\section*{Acknowledgement} This research is supported in part by the National Research Foundation, Singapore, under the NRF Medium Sized Centre Scheme (CARTIN).
Any opinions, findings and conclusions expressed in this material are those of the authors and do not reflect the views of National Research Foundation, Singapore.
This research is also supported in part by the ASEAN-China Cooperation Fund (ACCF) under project “Deep Ensemble Under Non-Ideal Conditions and Its Typical Applications in Computer Vision.”
{
    \small
    \bibliographystyle{ieeenat_fullname}
    \bibliography{main}

@String(CVPR= {IEEE Conf. Comput. Vis. Pattern Recog.})

@String(ICCV= {Int. Conf. Comput. Vis.})

@String(ECCV= {Eur. Conf. Comput. Vis.})

@String(ICLR = {Int. Conf. Learn. Represent.})

@String(CVPR  = {CVPR})

@String(ICCV  = {ICCV})

@String(ECCV  = {ECCV})

@String(ICLR  = {ICLR})

@inproceedings{gligen2023,
  title={Gligen: Open-set grounded text-to-image generation},
  author={Li, Yuheng and Liu, Haotian and Wu, Qingyang and Mu, Fangzhou and Yang, Jianwei and Gao, Jianfeng and Li, Chunyuan and Lee, Yong Jae},
  booktitle=CVPR,
  pages={22511--22521},
  year={2023}
}

@inproceedings{qahoi2021,
  title={Qahoi: Query-based anchors for human-object interaction detection},
  author={Chen, Junwen and Yanai, Keiji},
  booktitle={2023 18th International Conference on Machine Vision and Applications (MVA)},
  pages={1--5},
  year={2023},
  organization={IEEE}
}

@inproceedings{hicodet2018,
  title={Learning to detect human-object interactions},
  author={Chao, Yu-Wei and Liu, Yunfan and Liu, Xieyang and Zeng, Huayi and Deng, Jia},
  booktitle={2018 ieee winter conference on applications of computer vision (wacv)},
  pages={381--389},
  year={2018},
  organization={IEEE}
}

@misc{stablediffusion2021,
      title={High-Resolution Image Synthesis with Latent Diffusion Models}, 
      author={Robin Rombach and Andreas Blattmann and Dominik Lorenz and Patrick Esser and Björn Ommer},
      year={2021},
      eprint={2112.10752},
      archivePrefix={arXiv},
      primaryClass={cs.CV}
}

@article{ddpm2020,
  title={Denoising diffusion probabilistic models},
  author={Ho, Jonathan and Jain, Ajay and Abbeel, Pieter},
  journal={NeurIPS},
  volume={33},
  pages={6840--6851},
  year={2020}
}

@inproceedings{unet2015,
  title={U-net: Convolutional networks for biomedical image segmentation},
  author={Ronneberger, Olaf and Fischer, Philipp and Brox, Thomas},
  booktitle={Medical Image Computing and Computer-Assisted Intervention--MICCAI 2015: 18th International Conference, Munich, Germany, October 5-9, 2015, Proceedings, Part III 18},
  pages={234--241},
  year={2015},
  organization={Springer}
}

@article{nerf2022,
	title = {{NeRF}: representing scenes as neural radiance fields for view synthesis},
	volume = {65},
	issn = {0001-0782, 1557-7317},
	url = {https://dl.acm.org/doi/10.1145/3503250},
	doi = {10.1145/3503250},
	journal = {Communications of the ACM},
	author = {Mildenhall, Ben and Srinivasan, Pratul P. and Tancik, Matthew and Barron, Jonathan T. and Ramamoorthi, Ravi and Ng, Ren},
	month = jan,
	year = {2022},
	pages = {99--106}
}

@inproceedings{adam2014,
  title={Adam: A method for stochastic optimization},
  author={Kingma, Diederik P and Ba, Jimmy},
  booktitle    = {3rd International Conference on Learning Representations, {ICLR} 2015,
                  San Diego, CA, USA, May 7-9, 2015, Conference Track Proceedings},
  year         = {2015},
}

@inproceedings{ddim2021,
  title={Denoising Diffusion Implicit Models},
  author={Song, Jiaming and Meng, Chenlin and Ermon, Stefano},
  booktitle={ICLR},
  year={2021}
}

@InProceedings{hoe2024interactdiffusion,
    author    = {Hoe, Jiun Tian and Jiang, Xudong and Chan, Chee Seng and Tan, Yap-Peng and Hu, Weipeng},
    title     = {InteractDiffusion: Interaction Control in Text-to-Image Diffusion Models},
    booktitle = {CVPR},
    month     = {June},
    year      = {2024},
    pages     = {6180-6189}
}

@inproceedings{zhang2023pvic,
  author    = {Zhang, Frederic Z. and Yuan, Yuhui and Campbell, Dylan and Zhong, Zhuoyao and Gould, Stephen},
  title     = {Exploring Predicate Visual Context in Detecting Human–Object Interactions},
  booktitle = {ICCV},
  month     = {October},
  year      = {2023},
  pages     = {10411-10421},
}

@inproceedings{luo2024sichoi,
  title={Discovering Syntactic Interaction Clues for Human-Object Interaction Detection},
  author={Luo, Jinguo and Ren, Weihong and Jiang, Weibo and Chen, Xi'ai and Wang, Qiang and Han, Zhi and Liu, Honghai},
  booktitle={CVPR},
  pages={28212--28222},
  year={2024}
}

@inproceedings{cao2023masactrl,
  title={Masactrl: Tuning-free mutual self-attention control for consistent image synthesis and editing},
  author={Cao, Mingdeng and Wang, Xintao and Qi, Zhongang and Shan, Ying and Qie, Xiaohu and Zheng, Yinqiang},
  booktitle={ICCV},
  pages={22560--22570},
  year={2023}
}

@inproceedings{hertz2022p2p,
  title={Prompt-to-prompt image editing with cross attention control},
  author={Hertz, Amir and Mokady, Ron and Tenenbaum, Jay and Aberman, Kfir and Pritch, Yael and Cohen-Or, Daniel},
  booktitle={ICLR},
  year={2023}
}

@InProceedings{brooks2022instructpix2pix,
    author     = {Brooks, Tim and Holynski, Aleksander and Efros, Alexei A.},
    title      = {InstructPix2Pix: Learning to Follow Image Editing Instructions},
    booktitle  = {CVPR},
    year       = {2023},
}

@inproceedings{mokady2023nti,
  title={Null-text inversion for editing real images using guided diffusion models},
  author={Mokady, Ron and Hertz, Amir and Aberman, Kfir and Pritch, Yael and Cohen-Or, Daniel},
  booktitle={CVPR},
  pages={6038--6047},
  year={2023}
}

@inproceedings{
hu2022lora,
title={Lo{RA}: Low-Rank Adaptation of Large Language Models},
author={Edward J Hu and Yelong Shen and Phillip Wallis and Zeyuan Allen-Zhu and Yuanzhi Li and Shean Wang and Lu Wang and Weizhu Chen},
booktitle={ICLR},
year={2022},
url={https://openreview.net/forum?id=nZeVKeeFYf9}
}

@article{liu2023groundingdino,
  title={Grounding dino: Marrying dino with grounded pre-training for open-set object detection},
  author={Liu, Shilong and Zeng, Zhaoyang and Ren, Tianhe and Li, Feng and Zhang, Hao and Yang, Jie and Li, Chunyuan and Yang, Jianwei and Su, Hang and Zhu, Jun and others},
  journal={ECCV},
  year={2024}
}

@inproceedings{kirillov2023segmentanything,
  title={Segment anything},
  author={Kirillov, Alexander and Mintun, Eric and Ravi, Nikhila and Mao, Hanzi and Rolland, Chloe and Gustafson, Laura and Xiao, Tete and Whitehead, Spencer and Berg, Alexander C and Lo, Wan-Yen and others},
  booktitle={ICCV},
  pages={4015--4026},
  year={2023}
}

@article{oquab2023dinov2,
  title={Dinov2: Learning robust visual features without supervision},
  author={Oquab, Maxime and Darcet, Timoth{\'e}e and Moutakanni, Th{\'e}o and Vo, Huy and Szafraniec, Marc and Khalidov, Vasil and Fernandez, Pierre and Haziza, Daniel and Massa, Francisco and El-Nouby, Alaaeldin and others},
  journal={arXiv preprint arXiv:2304.07193},
  year={2023}
}

@inproceedings{deutch2024turboedit,
  title={Turboedit: Text-based image editing using few-step diffusion models},
  author={Deutch, Gilad and Gal, Rinon and Garibi, Daniel and Patashnik, Or and Cohen-Or, Daniel},
  booktitle={SIGGRAPH Asia 2024 Conference Papers},
  pages={1--12},
  year={2024}
}

@inproceedings{huberman2024editfriendly,
  title={An edit friendly {DDPM} noise space: Inversion and manipulations},
  author={Huberman-Spiegelglas, Inbar and Kulikov, Vladimir and Michaeli, Tomer},
  booktitle={CVPR},
  pages={12469--12478},
  year={2024}
}

@article{xu2025hoiedit,
  title={HOIEdit: Human--object interaction editing with text-to-image diffusion model},
  author={Xu, Tang and Wang, Wenbin and Zhong, Alin},
  journal={The Visual Computer},
  pages={1--13},
  year={2025},
  publisher={Springer}
}

@inproceedings{cha2025verbdiff,
  title={VerbDiff: Text-Only Diffusion Models with Enhanced Interaction Awareness},
  author={Cha, SeungJu and Lee, Kwanyoung and Kim, Ye-Chan and Oh, Hyunwoo and Kim, Dong-Jin},
  booktitle={Proceedings of the Computer Vision and Pattern Recognition Conference},
  pages={8041--8050},
  year={2025}
}

@article{labs2025flux1kontext,
  title={FLUX. 1 Kontext: Flow Matching for In-Context Image Generation and Editing in Latent Space},
  author={Labs, Black Forest and Batifol, Stephen and Blattmann, Andreas and Boesel, Frederic and Consul, Saksham and Diagne, Cyril and Dockhorn, Tim and English, Jack and English, Zion and Esser, Patrick and others},
  journal={arXiv preprint arXiv:2506.15742},
  year={2025}
}

@inproceedings{xiao2025omnigen,
  title={Omnigen: Unified image generation},
  author={Xiao, Shitao and Wang, Yueze and Zhou, Junjie and Yuan, Huaying and Xing, Xingrun and Yan, Ruiran and Li, Chaofan and Wang, Shuting and Huang, Tiejun and Liu, Zheng},
  booktitle={CVPR},
  pages={13294--13304},
  year={2025}
}

@book{rijsbergen1979fscore,
author = {Rijsbergen, C. J. Van},
title = {Information Retrieval},
year = {1979},
isbn = {0408709294},
publisher = {Butterworth-Heinemann},
address = {USA},
edition = {2nd}
}

@article{zhang2025eligen,
  title={Eligen: Entity-level controlled image generation with regional attention},
  author={Zhang, Hong and Duan, Zhongjie and Wang, Xingjun and Chen, Yingda and Zhang, Yu},
  journal={arXiv preprint arXiv:2501.01097},
  year={2025}
}

@article{hoe2025interactedit,
  title={InteractEdit: Zero-Shot Editing of Human-Object Interactions in Images},
  author={Hoe, Jiun Tian and Hu, Weipeng and Zhou, Wei and Xie, Chao and Wang, Ziwei and Chan, Chee Seng and Jiang, Xudong and Tan, Yap-Peng},
  journal={arXiv preprint arXiv:2503.09130},
  year={2025}
}

@InProceedings{Peebles2023DiT,
    author    = {Peebles, William and Xie, Saining},
    title     = {Scalable Diffusion Models with Transformers},
    booktitle = {Proceedings of the IEEE/CVF International Conference on Computer Vision (ICCV)},
    month     = {October},
    year      = {2023},
    pages     = {4195-4205}
}

@misc{wu2025qwenimagetechnicalreport,
      title={Qwen-Image Technical Report}, 
      author={Chenfei Wu and Jiahao Li and Jingren Zhou and Junyang Lin and Kaiyuan Gao and Kun Yan and Sheng-ming Yin and Shuai Bai and Xiao Xu and Yilei Chen and Yuxiang Chen and Zecheng Tang and Zekai Zhang and Zhengyi Wang and An Yang and Bowen Yu and Chen Cheng and Dayiheng Liu and Deqing Li and Hang Zhang and Hao Meng and Hu Wei and Jingyuan Ni and Kai Chen and Kuan Cao and Liang Peng and Lin Qu and Minggang Wu and Peng Wang and Shuting Yu and Tingkun Wen and Wensen Feng and Xiaoxiao Xu and Yi Wang and Yichang Zhang and Yongqiang Zhu and Yujia Wu and Yuxuan Cai and Zenan Liu},
      year={2025},
      eprint={2508.02324},
      archivePrefix={arXiv},
      primaryClass={cs.CV},
      url={https://arxiv.org/abs/2508.02324}, 
}

@inproceedings{esser2024sd3mmdit,
  title={Scaling rectified flow transformers for high-resolution image synthesis},
  author={Esser, Patrick and Kulal, Sumith and Blattmann, Andreas and Entezari, Rahim and M{\"u}ller, Jonas and Saini, Harry and Levi, Yam and Lorenz, Dominik and Sauer, Axel and Boesel, Frederic and others},
  booktitle={Forty-first international conference on machine learning},
  year={2024}
}

@inproceedings{zhou2024migc,
  title={Migc: Multi-instance generation controller for text-to-image synthesis},
  author={Zhou, Dewei and Li, You and Ma, Fan and Zhang, Xiaoting and Yang, Yi},
  booktitle={Proceedings of the IEEE/CVF Conference on Computer Vision and Pattern Recognition},
  pages={6818--6828},
  year={2024}
}

@inproceedings{lipman2023flowmatching,
  title={Flow Matching for Generative Modeling},
  author={Lipman, Yaron and Chen, Ricky TQ and Ben-Hamu, Heli and Nickel, Maximilian and Le, Matthew},
  booktitle={The Eleventh International Conference on Learning Representations},
  year={2023}
}

@article{cao2023unihoi,
  title={Detecting any human-object interaction relationship: Universal hoi detector with spatial prompt learning on foundation models},
  author={Cao, Yichao and Tang, Qingfei and Su, Xiu and Chen, Song and You, Shan and Lu, Xiaobo and Xu, Chang},
  journal={Advances in Neural Information Processing Systems},
  volume={36},
  pages={739--751},
  year={2023}
}

@inproceedings{lian2023llmgrounded,
  title={Llm-grounded diffusion: Enhancing prompt understanding of text-to-image diffusion models with large language models},
  author={Lian, Long and Li, Boyi and Yala, Adam and Darrell, Trevor},
  booktitle={arXiv preprint arXiv:2305.13655},
  year={2023}
}

@article{zhou2024migc++,
  title={Migc++: Advanced multi-instance generation controller for image synthesis},
  author={Zhou, Dewei and Li, You and Ma, Fan and Yang, Zongxin and Yang, Yi},
  journal={IEEE Transactions on Pattern Analysis and Machine Intelligence},
  year={2024},
  publisher={IEEE}
}

@inproceedings{deng2024fireflow,
  title={Fireflow: Fast inversion of rectified flow for image semantic editing},
  author={Deng, Yingying and He, Xiangyu and Mei, Changwang and Wang, Peisong and Tang, Fan},
  booktitle={ICML},
  year={2025}
}

@article{kirstain2023pickscore,
  title={Pick-a-pic: An open dataset of user preferences for text-to-image generation},
  author={Kirstain, Yuval and Polyak, Adam and Singer, Uriel and Matiana, Shahbuland and Penna, Joe and Levy, Omer},
  journal={Advances in neural information processing systems},
  volume={36},
  pages={36652--36663},
  year={2023}
}

@article{wu2023hpsv2,
  title={Human preference score v2: A solid benchmark for evaluating human preferences of text-to-image synthesis},
  author={Wu, Xiaoshi and Hao, Yiming and Sun, Keqiang and Chen, Yixiong and Zhu, Feng and Zhao, Rui and Li, Hongsheng},
  journal={arXiv preprint arXiv:2306.09341},
  year={2023}
}

@article{xu2023imagereward,
  title={Imagereward: Learning and evaluating human preferences for text-to-image generation},
  author={Xu, Jiazheng and Liu, Xiao and Wu, Yuchen and Tong, Yuxuan and Li, Qinkai and Ding, Ming and Tang, Jie and Dong, Yuxiao},
  journal={Advances in Neural Information Processing Systems},
  volume={36},
  pages={15903--15935},
  year={2023}
}

@inproceedings{wang2024instancediffusion,
  title={Instancediffusion: Instance-level control for image generation},
  author={Wang, Xudong and Darrell, Trevor and Rambhatla, Sai Saketh and Girdhar, Rohit and Misra, Ishan},
  booktitle={Proceedings of the IEEE/CVF conference on computer vision and pattern recognition},
  pages={6232--6242},
  year={2024}
}

@article{su2024rope,
  title={Roformer: Enhanced transformer with rotary position embedding},
  author={Su, Jianlin and Ahmed, Murtadha and Lu, Yu and Pan, Shengfeng and Bo, Wen and Liu, Yunfeng},
  journal={Neurocomputing},
  volume={568},
  pages={127063},
  year={2024},
  publisher={Elsevier}
}

@article{wu2025omnigen2,
  title={OmniGen2: Exploration to Advanced Multimodal Generation},
  author={Chenyuan Wu and Pengfei Zheng and Ruiran Yan and Shitao Xiao and Xin Luo and Yueze Wang and Wanli Li and Xiyan Jiang and Yexin Liu and Junjie Zhou and Ze Liu and Ziyi Xia and Chaofan Li and Haoge Deng and Jiahao Wang and Kun Luo and Bo Zhang and Defu Lian and Xinlong Wang and Zhongyuan Wang and Tiejun Huang and Zheng Liu},
  journal={arXiv preprint arXiv:2506.18871},
  year={2025}
}

@article{ho2022cfg,
  title={Classifier-free diffusion guidance},
  author={Ho, Jonathan and Salimans, Tim},
  journal={arXiv preprint arXiv:2207.12598},
  year={2022}
}

@article{2020t5,
  author  = {Colin Raffel and Noam Shazeer and Adam Roberts and Katherine Lee and Sharan Narang and Michael Matena and Yanqi Zhou and Wei Li and Peter J. Liu},
  title   = {Exploring the Limits of Transfer Learning with a Unified Text-to-Text Transformer},
  journal = {Journal of Machine Learning Research},
  year    = {2020},
  volume  = {21},
  number  = {140},
  pages   = {1-67},
  url     = {http://jmlr.org/papers/v21/20-074.html}
}

@misc{flux2024,
    author={Black Forest Labs},
    title={FLUX},
    year={2024},
    note={GitHub repository},
}
}

\clearpage
\setcounter{page}{1}
\maketitlesupplementary
\appendix

\section{Implementation Details}\label{supp-sec:implementation-details}
We build our model by adapting Flux.1 Kontext \cite{labs2025flux1kontext}, Eligen \cite{zhang2025eligen} and Flux.1 Dev \cite{flux2024} backbone. The text encoder weights are kept frozen during training, and we applied LoRA \cite{hu2022lora} fine-tuning on the linear layers of each block in the DiT, with a rank of 64, resulting in 0.3 billion trainable parameters (2.5\% of the frozen 12B base model). The HOI Encoder (17M) is trained from scratch, while the backbone is adapted via 344M trainable LoRA parameters. We train our model on two NVIDIA RTX 6000 ADA GPUs, with constant learning rate of $1\times10^{-4}$ and bf16 precision. We train on resolution buckets, randomly sampling from the following resolutions (height, width) at each step: (1024, 1024), (768, 1360), (1360, 768), (880, 1168), (1168, 880), (1248, 832), and (832, 1248). For the editing task, we follow Flux.1 Kontext \cite{labs2025flux1kontext} to separate the source image from the noisy latent. The VAE-encoded source image latent patches are assigned RoPE indexes of $(1,x,y)$ while the noise latents are assigned $(0,x,y)$, respectively. For the arbitrary shape, the Fourier embedding $e_{\text{box}}(b_n^r)$ is obtained using the shape's minimum enclosing bounding box. During inference, we use 28 sampling steps and set the classifier-free guidance scale \cite{ho2022cfg} to 3.5. 
% The global text prompt allows users to provide descriptive details (\eg, background, style, or specific object attributes) that cannot be captured by HOI triplets alone.

\subsection{Sequence length and budgeting.}
Each HOI interaction yields \emph{role sequences} of HOI tokens: subject \(\mathcal{S}_n\)\, object \(\mathcal{O}_n\) and action \(\mathcal{A}_n)\); an object-only case contributes just \((\mathcal{O}_n)\). We cap the total HOI-token budget at \(K_{\text{HOI}}\) (default to \(4608\) for 48GB GPU memory) and per-sequence length at \(L_{\max}\) (default \(512\)).
Let \(M\) be the number of active role sequences, we assign the same length $L$ to every active sequence,
{\setlength{\abovedisplayskip}{-1pt}
\setlength{\belowdisplayskip}{1pt}\[
L=\min\!\Big(L_{\max},\ \Big\lfloor \tfrac{K_{\text{HOI}}}{M}\Big\rfloor\Big), \]}
so that the total HOI-token count satisfy \(M\,L\le K_{\text{HOI}}\). Practically, each role sequence is padded or truncated to length $L$ for batching.

\subsection{Nano Banana.} We compare our method against Nano Banana as a representative closed-source baseline. We access the model via the Gemini API \footnote{\url{https://aistudio.google.com/}} using the \texttt{gemini-2.5-flash-image} variant. For fairness, we employ the identical text prompts and source images defined in our editing task (\cref{fig:editing}). Since the Gemini API does not currently expose parameters for seed control or stochasticity, we report results from a single inference trial per prompt to evaluate its default zero-shot performance.

\subsection{InteractEdit + InteractDiffusion Baseline} \label{supp-subsec:intedit+intdiff} To establish a rigorous baseline for layout-guided HOI editing, we integrate the state-of-the-art InteractEdit \cite{hoe2025interactedit} and InteractDiffusion \cite{hoe2024interactdiffusion} frameworks. We adapt the original SDXL-based InteractEdit backbone to the InteractDiffusion-XL variant. Our implementation follows a two-stage inversion process for each source image in the IEBench benchmark. In a departure from the standard text-only inversion used in InteractEdit, we leverage InteractDiffusion's native support for structural guidance by incorporating HOI triplets and bounding boxes throughout the inversion stages. Specifically, we execute the inversion for 1000 steps in Stage 1 and 200 steps in Stage 2, adhering to the default configurations of InteractEdit. During the editing phase, we synthesize the final image by conditioned generation using the inverted weights and a structured prompt: ``a photo of ⟨subject⟩ ⟨target action⟩ ⟨object⟩ at ⟨background⟩''. This process is further guided by the target HOI triplet and the specified HOI layout, ensuring the baseline is evaluated under identical conditioning to our proposed method. Finally, we apply the standard IEBench evaluation strategy to ensure a fair and consistent comparison across all reported metrics.

\section{Dataset Details}
\subsection{Synthesis of Target Layouts for IEBench} \label{subsec:iebench_layout}
The IEBench benchmark \cite{hoe2025interactedit} is designed for layout-free editing and thus does not provide target bounding boxes for edits. First, we built a statistical geometry bank from the HICO-DET training set. For each HOI class ⟨action,object⟩, we computed a 5-dimensional \textbf{multivariate Gaussian distribution}. This distribution models the object's geometry relative to the subject, using a 5D vector that captures the relative centre displacement $(dx,dy)$, relative object size $(rw,rh)$, all scaled by the subject's height, and the Intersection-over-Union (IoU).

To generate a target layout for a specific edit in IEBench, we used this statistical model along with a manually specified heuristic. We categorised objects as \textit{``large/stable''} (\eg, bed, bus) or \textit{``small/movable''} (\eg, skateboard, cell phone). For edits involving \textbf{large objects}, we fixed the object's bounding box $(b_o)$ from the source image and sampled a new subject box $(b_s)$ from the learned relative distribution. Conversely, for \textbf{small objects}, we fixed the subject's box $(b_s)$ and sampled a new object box $(b_o)$. In some ambiguous cases, both boxes were sampled.

We generated proposals for all 100 edits in IEBench. These proposals were then manually inspected to filter out any implausible layouts, such as those with unreasonable aspect ratios, sizes, or positions.

\subsection{MultiHOIEdit}\label{subsec:multihoiedit}
To evaluate the novel task of multi-HOI editing, for which no existing benchmark exists to our knowledge, we introduce \textbf{MultiHOIEdit}. The process began by creating a set of high-quality source images. We used the Flux.1 model to synthesise images containing two or three distinct HOIs, focusing on scenes with different objects to ensure complexity. The generation of plausible multi-HOI images proved to be exceptionally challenging; to ensure correctness, we verified each synthesised image using the PViC HOI detector \cite{zhang2023pvic} and retained only those where all target interactions were successfully detected. This rigorous filtering process had a very low yield, with only \textbf{200 valid source images} being selected from an initial pool of 8,942 generations (a 2.2\% success rate), underscoring the difficulty of the task.

From this curated set of source images, we then defined the target edits. For each source image, we created one to three distinct editing tasks, where each task involved modifying two or more of the existing HOIs simultaneously. The target layouts for these new interactions were proposed by extending the statistical geometry bank method described in \cref{subsec:iebench_layout} and were then manually filtered for quality and plausibility. The final MultiHOIEdit benchmark comprises \textbf{103 unique source images} and a total of \textbf{200 distinct multi-interaction editing tasks}. Qualitative examples of these complex edits are provided in \cref{fig:example_multi_supp}.

\begin{figure*}[p]
    \centering
    \begin{minipage}[b]{0.48\textwidth} % Adjust width as needed
        \centering
        \includegraphics[width=1\linewidth]{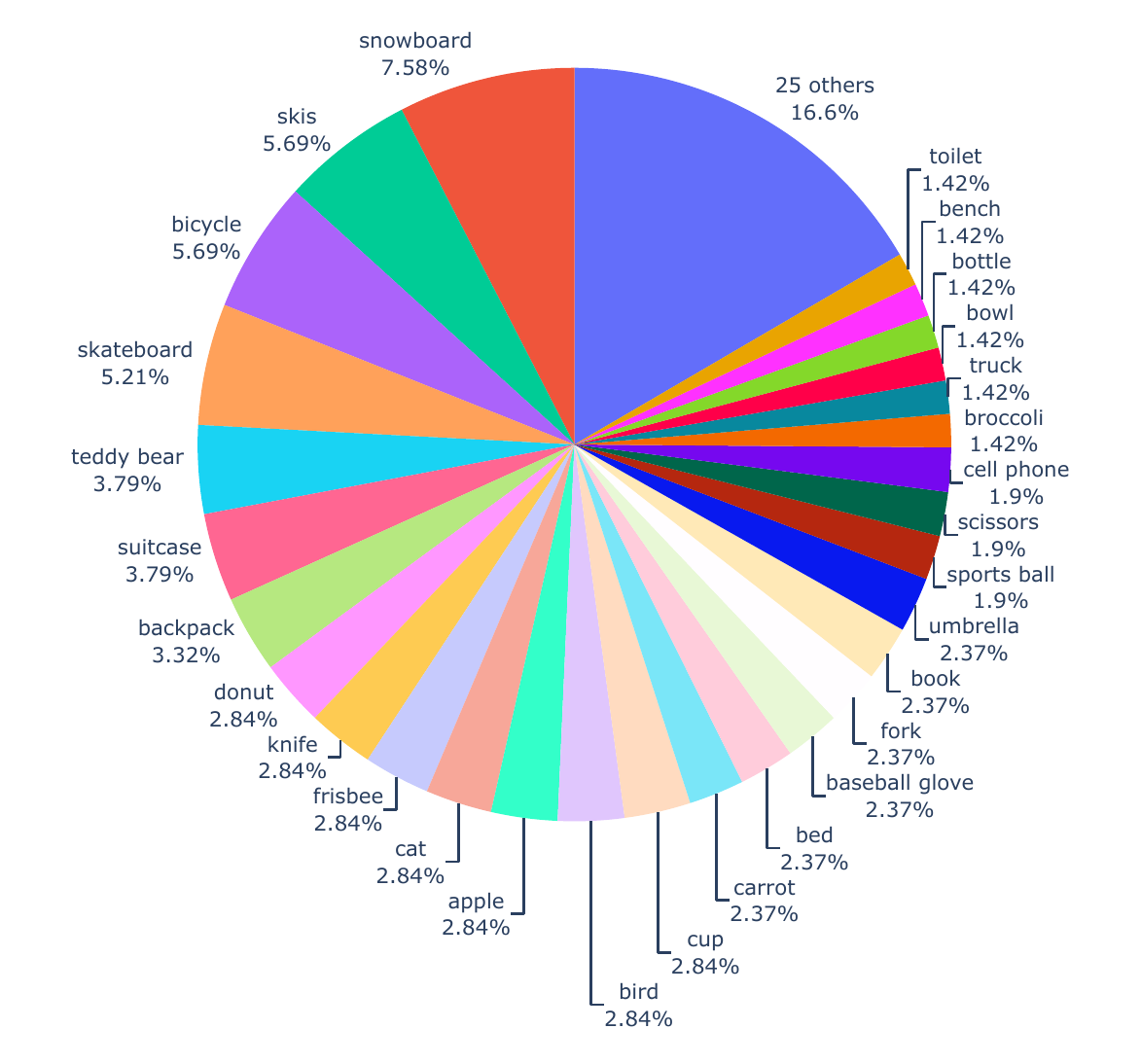}
        \vspace{-22pt}
        \caption{Distribution of the 54 object categories within the \nobreak{MultiHOIEdit}. The ``25 others'' aggregates the least frequent categories with 2 or fewer appearance.}
        \label{fig:multihoiedit_object_cat}
        \vspace{-0pt}

        \includegraphics[width=1\linewidth]{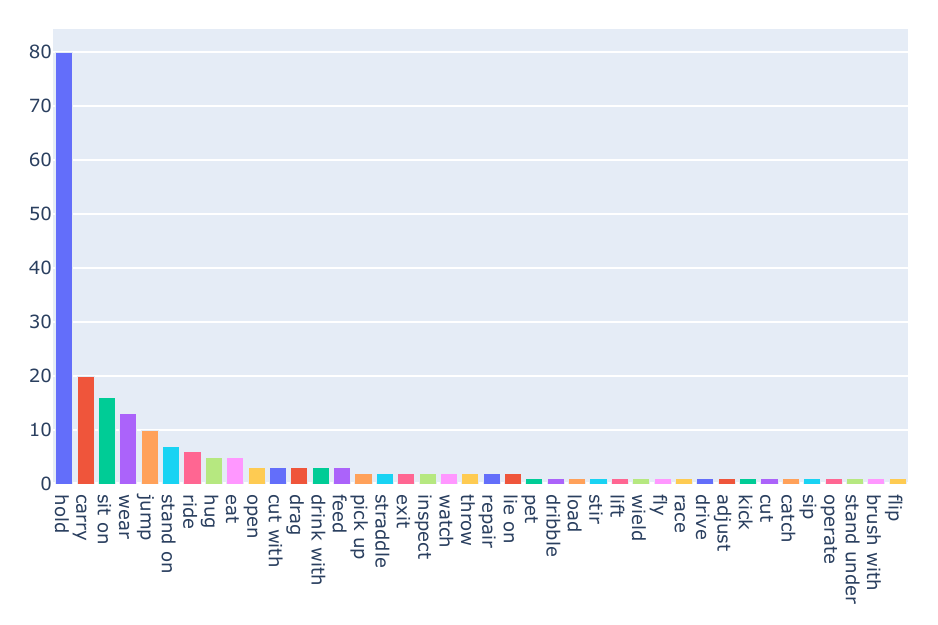}
        \vspace{-24pt}
        \caption{Distribution of source (pre-edit) actions in MultiHOIEdit.}
        \label{fig:multihoiedit_source_actions}
        \vspace{-0pt}

        \includegraphics[width=1\linewidth]{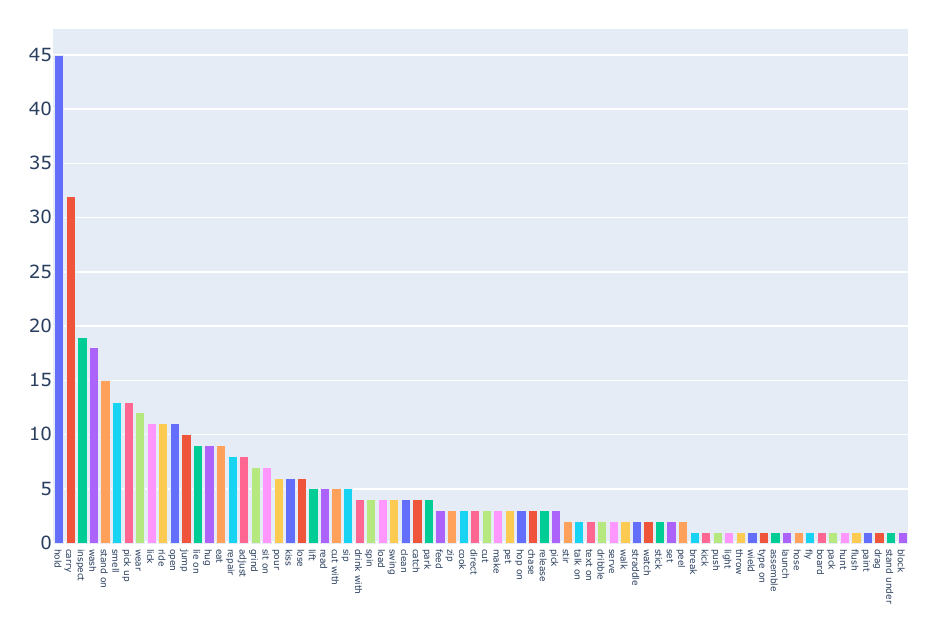}
        \vspace{-24pt}
        \caption{Distribution of 74 target (post-edit) actions in MultiHOIEdit.}
        \label{fig:multihoiedit_target_actions}
        \vspace{-5pt}
    \end{minipage}
    \hfill
    \begin{minipage}[b]{0.48\textwidth} % Adjust width as needed
        \centering
        \includegraphics[width=0.98\linewidth]{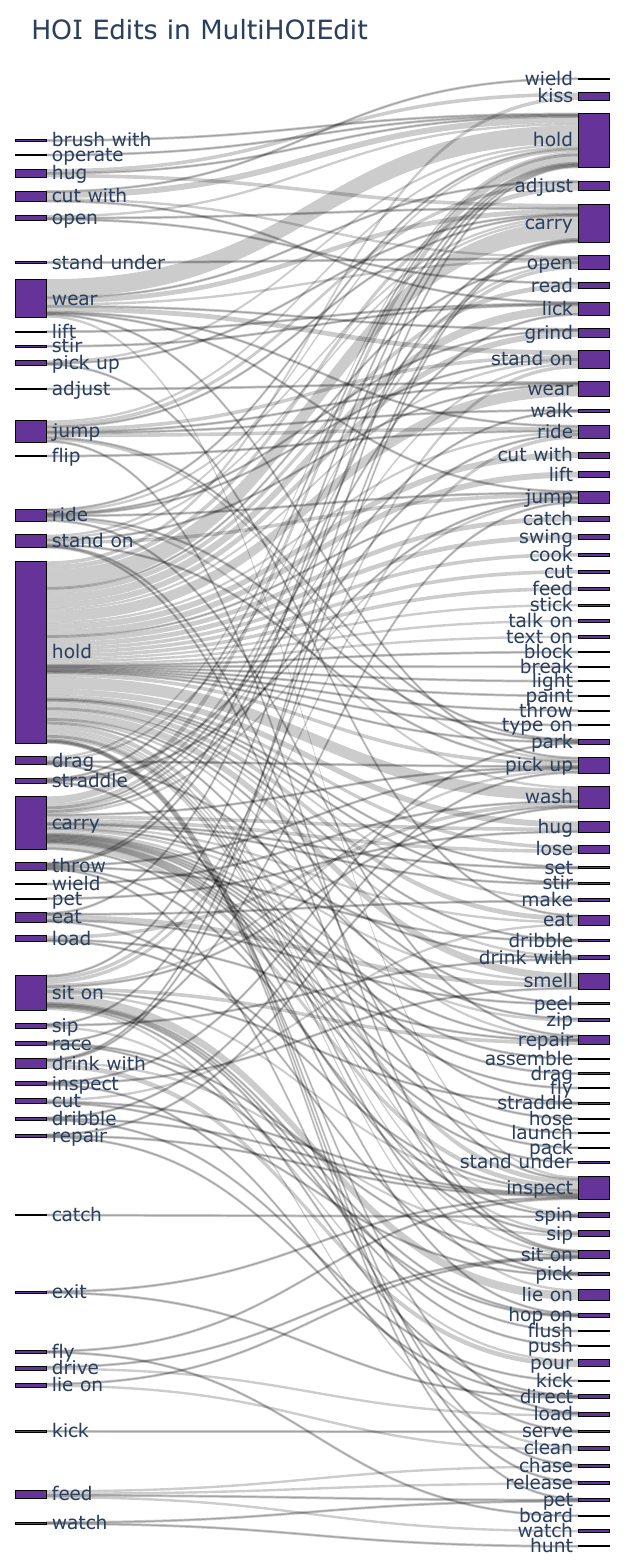}
        \vspace{-12pt}
        \caption{Sankey diagram visualising the action transitions in the MultiHOIEdit benchmark. The flows illustrate the mapping from source actions (left) to target actions (right), detailing the full range of edits.}
        \label{fig:multihoiedit_sankey}
    \end{minipage}
\end{figure*}

The benchmark is diverse, covering \textbf{54 object categories} (\cref{fig:multihoiedit_object_cat}) and a total of \textbf{40 source actions} (\cref{fig:multihoiedit_source_actions}) and \textbf{74 target actions} (\cref{fig:multihoiedit_target_actions}). Overall, the tasks involve transitions between \textbf{112 source HOI-object pairs} and \textbf{252 target HOI-object pairs}, with the full range of edits detailed in \cref{fig:multihoiedit_sankey}. We will release \textbf{MultiHOIEdit} publicly.

\subsection{HOI-Edit-44K} \label{supp-subsec:hoiedit44k}
The HOI-Edit-44K dataset addresses the critical scarcity of large-scale, paired data for the task of human-object interaction editing. 
The final dataset consists of 44,117 high-quality, paired HOI editing examples. Each sample in the dataset includes (1) the source image, (2) the target interaction triplet (subject, object, action), (3) the edited image and (4) the corresponding HOI layout for the edited image.

\begin{figure}[tp!]
    \centering
    \includegraphics[width=\linewidth]{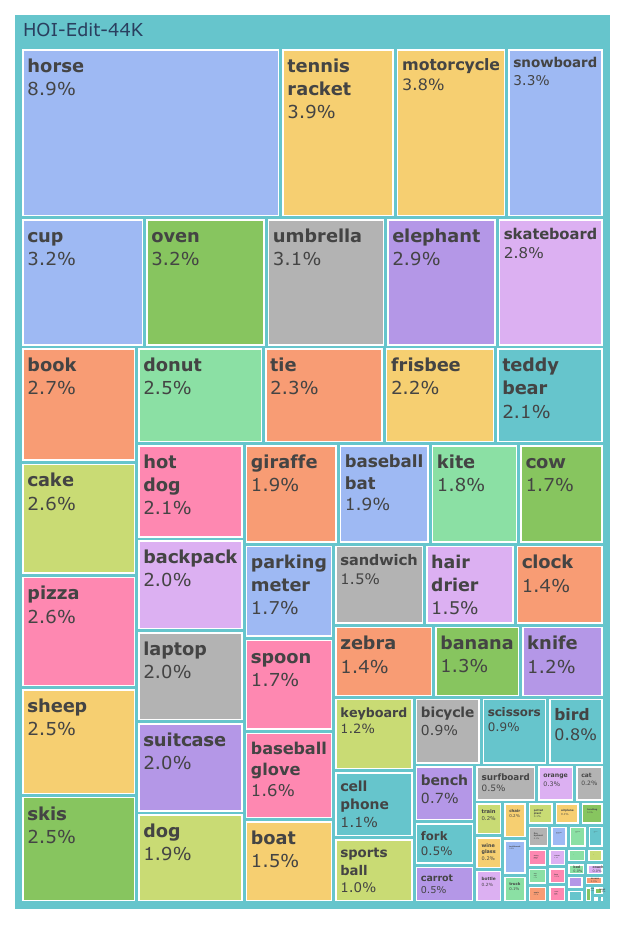}
    \vspace{-25pt}
    \caption{Treemap visualising the distribution of the interacting object categories in the HOI-Edit-44K dataset. The size of each block corresponds to the category's frequency.}
    \label{fig:hoiedit44k_object_cat}
    \vspace{-10pt}
\end{figure}

\begin{figure}[tp!]
    \centering
    \includegraphics[width=1\linewidth]{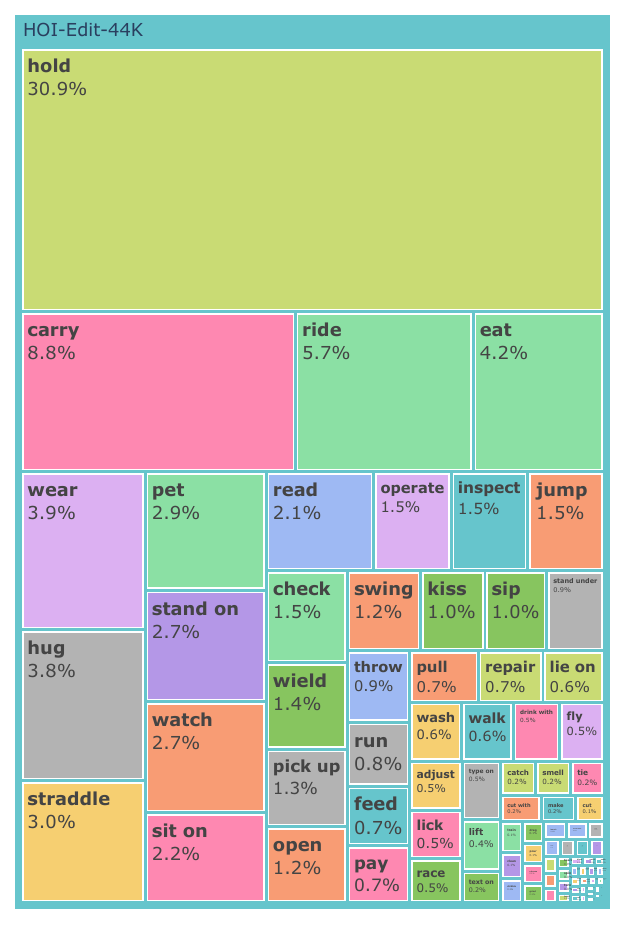}
    \vspace{-25pt}
    \caption{Treemap visualising the distribution of target action categories in the HOI-Edit-44K dataset.}
    \label{fig:hoiedit44k_action_cat}
    \vspace{-30pt}
\end{figure}

The dataset is diverse, containing 79 unique object categories (\cref{fig:hoiedit44k_object_cat}) and 92 unique target actions (\cref{fig:hoiedit44k_action_cat}), which combine to form 372 unique HOI triplets. See \cref{fig:example_hoiedit44k_supp} for qualitative examples. This resource was critical for jointly training our unified model, providing the necessary supervision for robust, identity-preserving HOI editing .
% Total samples: 44117
% Total unique target actions: 92
% Total unique objects: 79
% Total unique HOIs: 372

\noindent\textbf{Generalization and reliability.} Identity-preserving HOI edit pairs are scarce, necessitating our strictly curated HOI-Edit-44K. Our sources images are not purely synthetic as they come from both Flux.1 generations and real HICO-DET photos. We retain pairs only if they satisfy two rigorous criteria: HOI correctness via PViC and identity consistency via DINOv2 ($\ge$ 0.75). This strict quality control yields a $\sim$90\% rejection rate, which ensures the high reliability and physical plausibility of the final 44K curated pairs. Crucially, we also jointly train on HOI generation using real HICO-DET images. This exposes the model to real-scene statistics and interaction distributions beyond synthetic edits, anchoring the learned representation in real-image distributions and effectively mitigating potential teacher-model bias.

\section{Evaluation}
\subsection{Interaction Editing}\label{supp-subsec:iebench}
For Interaction Editing task, we follow the evaluation protocol of InteractEdit \cite{hoe2025interactedit} in their proposed IEBench. These evaluation metrics specifically designed for HOI editing task and quantify the trade-off between intended interaction transformation correctness and identity preservation.

\textbf{(i) HOI Editability, \text{\textsc{he}}} quantifies editing success by determining whether the target interaction is present in the edited image. Leveraging PViC \cite{zhang2023pvic}, a state-of-the-art HOI detector, each generated image is assigned a score of one if the target interaction is detected, and zero otherwise. The final \text{\textsc{he}} score is computed as the mean detection rate over all edited samples.

\textbf{(ii) Editability-Identity Score, \text{\textsc{ei}}} quantifies the trade-off between HE score and Identity Consistency via the harmonic mean, analogous to the $F_1$ score \cite{rijsbergen1979fscore}. This formulation ensures a balanced evaluation by penalizing low performance in either dimension:
\begin{align}
    \text{\textsc{ei}} &= \frac{2 \times \text{HOI Editability} \times \text{Identity Consistency}} {\text{HOI Editability} + \text{Identity Consistency}}.
\end{align}
Here, Identity Consistency assesses how well the subject and object identities are preserved after editing.  To compute this, GroundingDINO \cite{liu2023groundingdino} and SAM \cite{kirillov2023segmentanything} is used to detect and segment the subject and object in both the source and edited images. Then, DINOv2 \cite{oquab2023dinov2} is used for extracting feature embeddings, and the cosine similarity between embeddings of source-subject and edited-subject (and similarly for the object) is calculated and aggregated over images and seeds.

\subsection{Human Evaluation Study}\label{supp-sec:human-study} 
\begin{figure}[ht]
    \centering
    \includegraphics[width=1\linewidth]{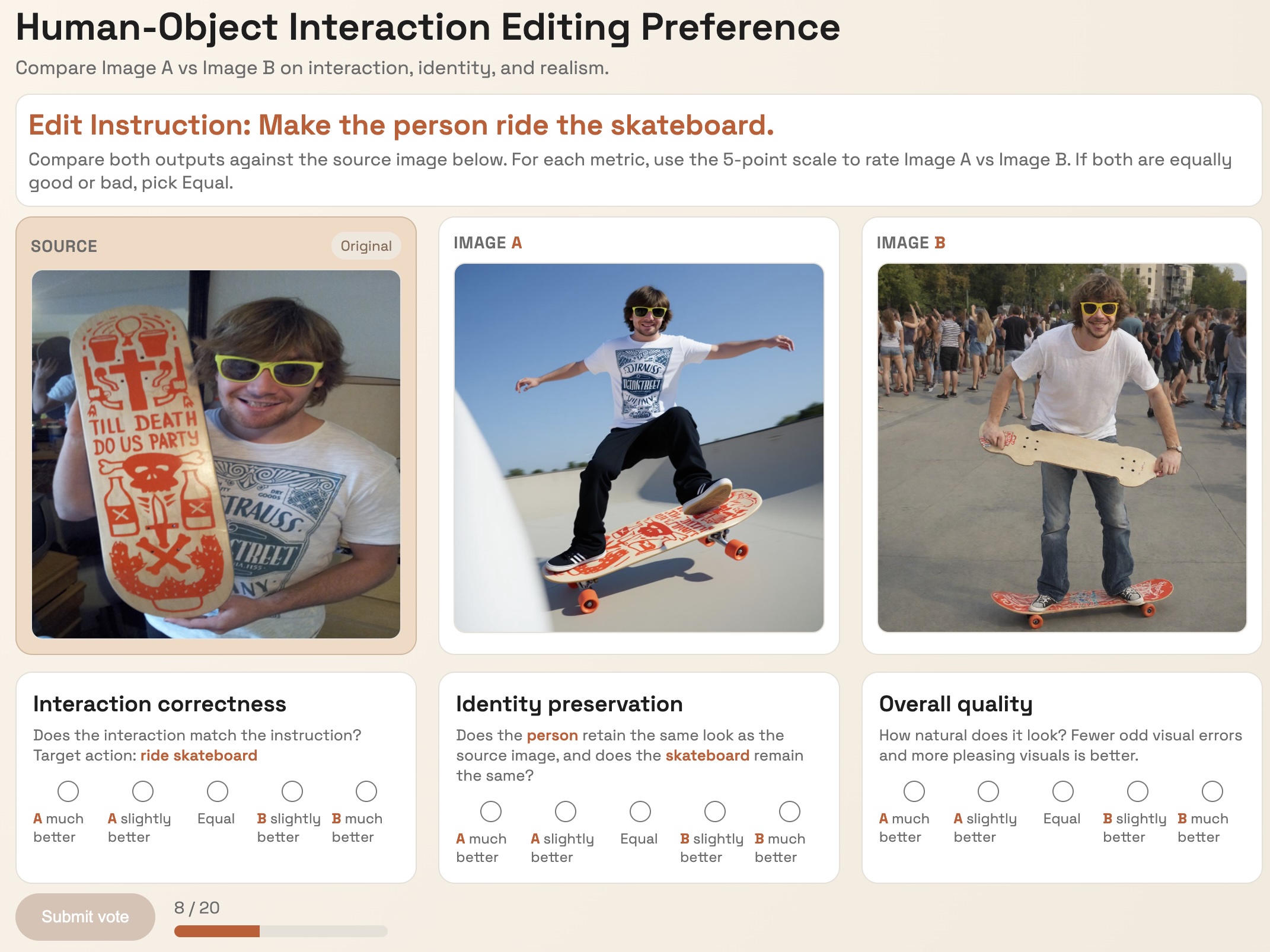}
    \vspace{-18pt}
    \caption{Evaluation Interface. Web-based survey used for data collection. Participants performed side-by-side comparisons of two models based on a source image and target edit instruction.}
    \label{fig:survey-site}
    \vspace{-10pt}
\end{figure}

To complement the quantitative results, we conducted a rigorous human preference study evaluating \textbf{HOI Correctness}, \textbf{Identity Preservation}, and \textbf{Overall Quality}. The study utilized a blind, randomized side-by-side comparison format where 26 unique respondents evaluated a total of \textbf{450 trials} ($N$=450). As illustrated in the provided survey interface in \cref{fig:survey-site}, participants were presented with a source image and a specific edit instruction, such as "Make the person ride the skateboard". For each trial, respondents rated two anonymized outputs: our model versus a baseline, using a 5-point Likert scale ranging from "A much better" to "B much better," with an "Equal" option for ties.

\begin{figure}[ht!]
    \centering
    \includegraphics[width=1\linewidth]{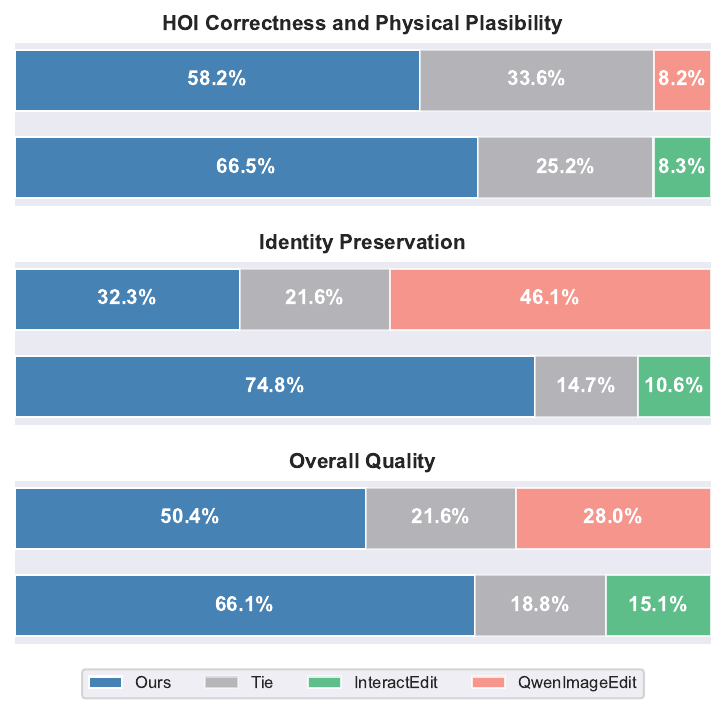}
    \vspace{-25pt} % 10
    \caption{Results of the Human Preference Study. Aggregated preference percentages for HOI Correctness and Physical Plausibility, Identity Preservation, and Overall Quality. The top bar in each category compares OneHOI (Ours) against QwenImageEdit, while the bottom bar compares it against InteractEdit.}
    \label{fig:survey-result}
    \vspace{-12pt}
\end{figure}

The results, summarized in \cref{fig:survey-result}, demonstrate that our method significantly outperforms leading baselines in physical plausibility and structural coherence. When compared against QwenImageEdit, our model was preferred in 58.2\% of cases for HOI Physics Plausibility, while the baseline was favoured in only 8.2\% of trials. Furthermore, our approach achieved a commanding 72.0\% win/tie rate in Overall Quality, consisting of a 50.4\% outright win rate and a 21.6\% tie rate. In comparisons with InteractEdit, our model maintained a superior win rate for Identity Preservation (74.8\%) and Overall Quality (66.1\%). These findings suggest that our unified representation effectively resolves the trade-off between executing complex interaction edits and maintaining the structural identity of the original scene.

% \section{Layout Grounding}
% \section{Community LoRA}

\section{Additional Qualitative Results}
\begin{figure*}[p] % t
    \centering
    \includegraphics[width=1\linewidth]{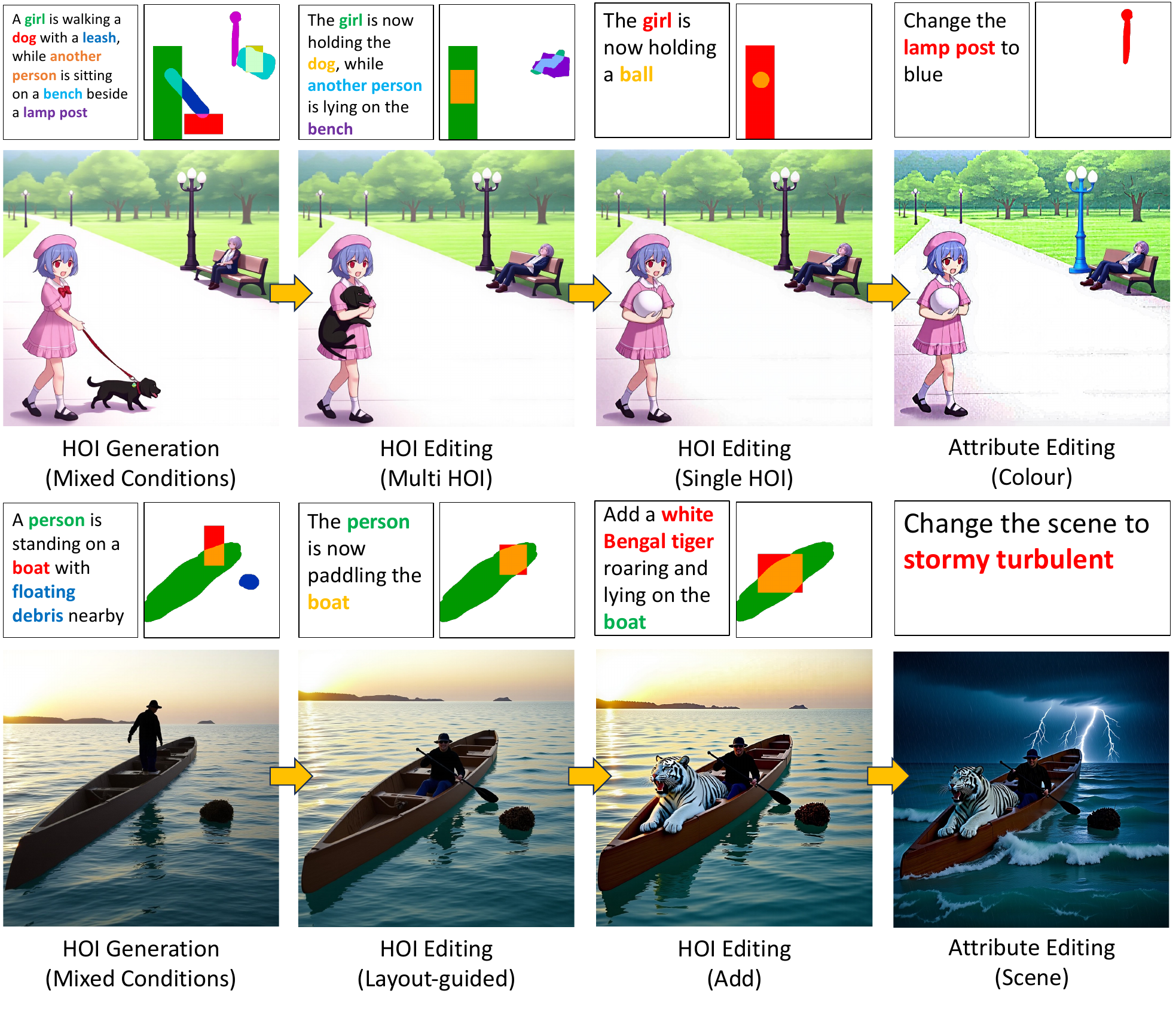}
    \vspace{-18pt}
    \caption{
    % \textbf{Versatile workflow for unified HOI generation and editing.} \paper enables a seamless, multi-step workflow within a single model. Starting with (1) \textbf{Mixed-Condition Generation}, it synthesises a complex scene from layout-guided HOIs (walking dog) and arbitrary shape-guided HOIs (sitting on bench) alongside independent objects (lamp post, leash). The model then supports (2) \textbf{Multi-HOI Editing}, simultaneously updating two distinct interactions, followed by (3) \textbf{Single-HOI Editing} and (4) \textbf{Attribute Editing} (changing object colour). 
    \textbf{Versatile workflow for unified HOI generation and editing using \paper.} \paper enables a seamless, multi-step workflow within a single model, showcasing diverse conditional control. Starting with: \\
    \textbf{Top Row: Urban Park Scene.} (1) \textbf{Mixed-Condition Generation} synthesises a complex scene from layout-guided HOIs (\ie, \textit{walking dog}) and arbitrary shape-guided independent objects (\ie, \textit{lamp post}, \textit{leash}), alongside another HOI (\ie, \textit{person sitting on bench}). (2) \textbf{Multi-HOI Editing} simultaneously updates two distinct interactions (\ie, \textit{holding dog}, \textit{person lying on bench}). (3) \textbf{Single-HOI Editing} modifies one interaction (\ie, \textit{holding ball}). (4) \textbf{Attribute Editing} changes an object's colour (\ie, \textit{black} $\rightarrow$ \textit{blue}). \\ 
    \textbf{Bottom Row: Ocean Survival Scene.} (1) \textbf{Mixed-Condition Generation} creates a challenging open-water scenario from a person standing on a boat and arbitrary shape-guided floating debris. (2) \textbf{Layout-guided HOI Editing} precisely changes the person's action (\ie, \textit{paddling the boat}). (3) \textbf{HOI Editing (Add)} introduces a new interaction (\ie, \textit{white Bengal tiger roaring and lying on the boat}). (4) \textbf{Attribute Editing (Scene)} transforms the entire environment (\ie, \textit{day} $\rightarrow$ \textit{stormy}, \textit{calm} $\rightarrow$ \textit{turbulent} ocean).
    }
    \label{fig:workflow_girl}
    \vspace{-10pt}
\end{figure*}

We provide additional qualitative results for the layout-free HOI editing task in \Cref{fig:editing_supp}. Likewise, \Cref{fig:generation-supp} presents additional qualitative results for HOI generation. Furthermore, \Cref{fig:workflow_girl} serves as a visual representation to the paper's core question, demonstrating that \textbf{HOI generation and editing are successfully unified within a single framework.} The step-by-step workflow showcases the seamless integration of initial HOI generation, multi-HOI editing, single-HOI editing, and attribute editing, thereby demonstrate the comprehensive and versatile control enabled by our method.

\begin{figure}[h!]
    \centering
    \vspace{-10pt} % 12
    \includegraphics[width=1\linewidth]{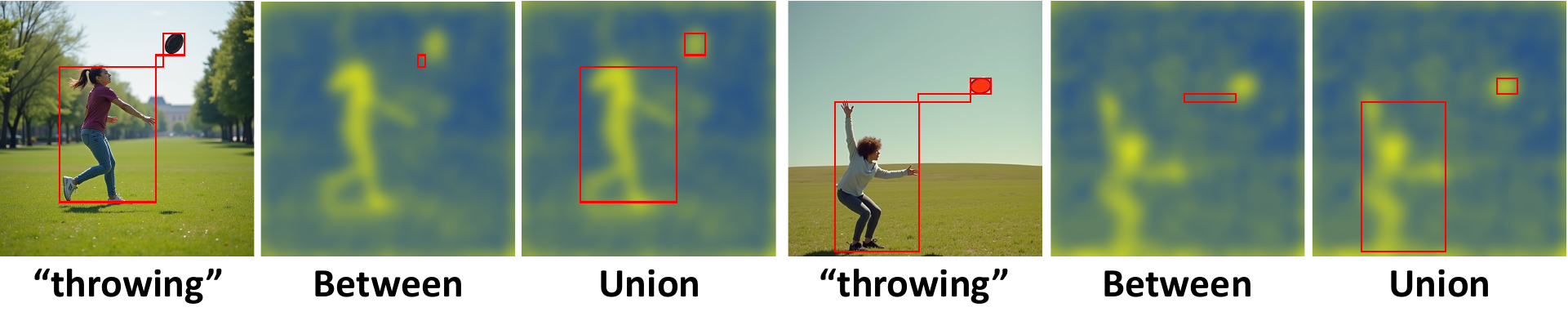}
    \vspace{-24pt} % 30
    \caption{Attention footprint of Flux.1. ``Union'' better matches the attention footprint compared to ``Between''.}
    \vspace{-14pt} % 12
    \label{fig:frisbee}
\end{figure}

\noindent\textbf{Spatial action region for remote action.} We use \text{subject}$~\cup$~\text{object} as an attention-aligned action grounding prior. \cref{fig:union} (main paper) shows that for disjoint interactions, action-token attention concentrates on entities, and \textit{union} matches this footprint better than the “\textit{Between}” band. We further validate this on a trajectory verb (“throwing frisbee” in \cref{fig:frisbee}). The action-token attention focuses on the thrower and frisbee, and the \textit{union} region matches this footprint, while the “\textit{Between}” band is often narrow/misplaced.

\section{Ablation on Unification vs. Task-specific}\label{supp-sec:ablation-unification} We unify HOI generation and editing by supporting mixed conditioning for real-world use cases (text-only, partial layouts, or multi-HOI). Separate training yields brittle, task-specific priors. Notably, the generation becomes strictly layout-dependent, while editing fails to scale to multi-HOI. As shown in \cref{tab:abl-uni-vs-specific}, \textbf{Unified} model consistently outperforms \textbf{task-specific} models trained under matched computation (1k steps), improving HOI Accuracy by 26.4\% in generation and HOI Editability by 21.1\% in layout-free editing. This confirms that joint training enables a \textit{``synergy effect''}, where generative priors enhance editing robustness and vice versa. (\textit{Note}: Task-specific = single-task models)

\begin{table*}[p] % ht
    \caption{Ablation on Unification.}
    \label{tab:abl-uni-vs-specific}
    \vspace{-5pt}
    \centering
    % \vspace{-9pt}
    \setlength{\tabcolsep}{6pt}
    \resizebox{0.9\linewidth}{!}{%
    \begin{tabular}{@{}llcc@{}} %  
        % \hline 
        \toprule
        Task Scenario & Metric & Task-specific & \textbf{Unified (Ours)} \\ 
        % \hline 
        \midrule
        Generation & Spatial $\uparrow$ / HOI Acc. $\uparrow$~~ & 0.422 / 0.177 & \textbf{0.443 / 0.224} \\ %\midrule
        Layout-Free Edit~~ & Editability-Identity $\uparrow$ / HOI Editability $\uparrow$ & 0.574 / 0.464 & \textbf{0.611 / 0.562} \\ %\midrule
        Multi-HOI Edit & Editability-Identity $\uparrow$ / HOI Editability $\uparrow$ & 0.391 / 0.287 & \textbf{0.435 / 0.329} \\ 
        % \hline 
        \bottomrule
    \end{tabular}
    }
\end{table*}

\begin{figure*}[t]
\centering
\setlength{\tabcolsep}{1pt} % General space between cols (6pt standard)
\renewcommand{\arraystretch}{1} % General space between rows (1 standard) 3.6 / 3.3 
\begin{tabular}{cccccc}
Source & Target & Source & Target & Source & Target\\
\includegraphics[width=.160\linewidth]{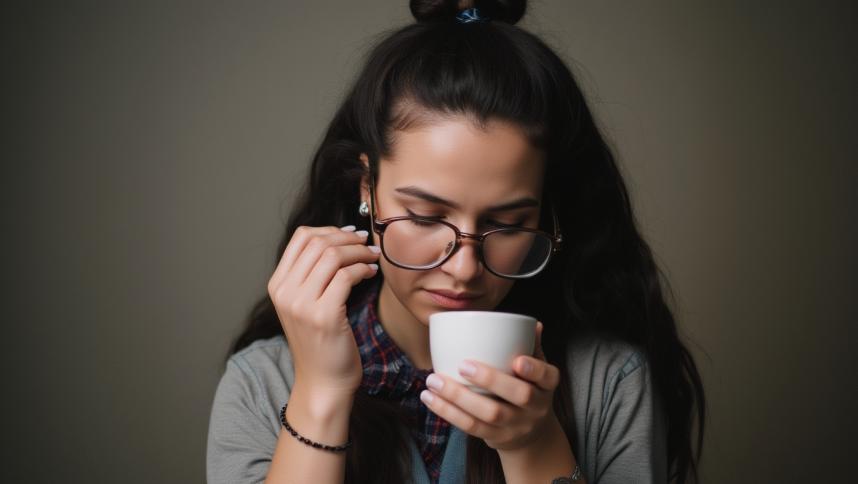}
& \includegraphics[width=.160\linewidth]{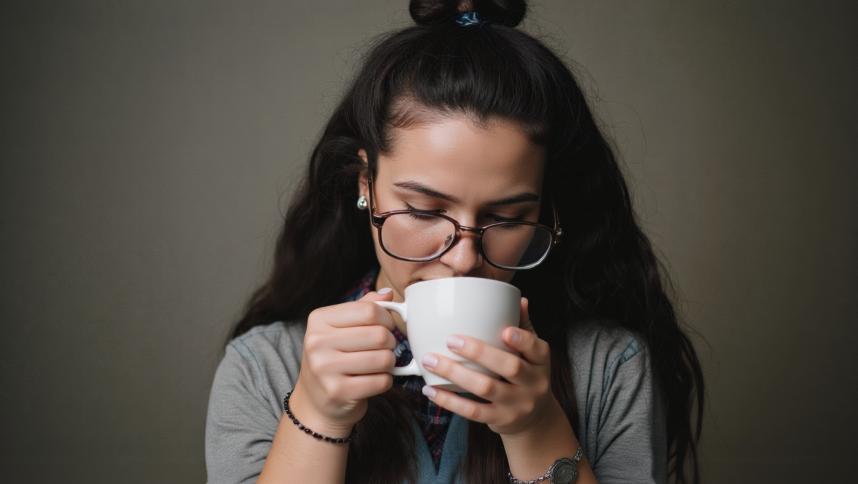}
& \includegraphics[width=.125\linewidth]{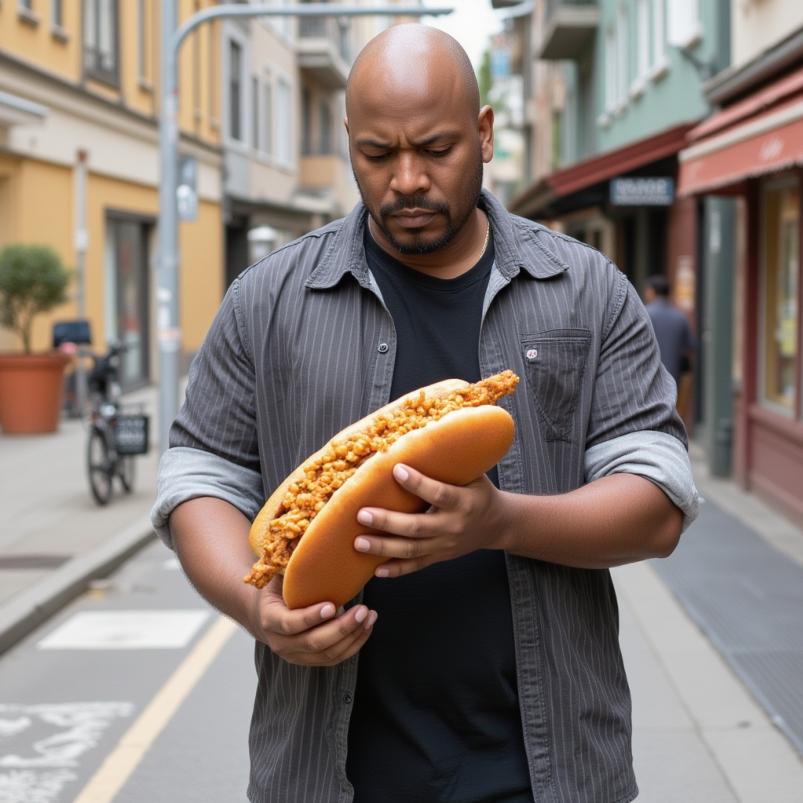}
& \includegraphics[width=.125\linewidth]{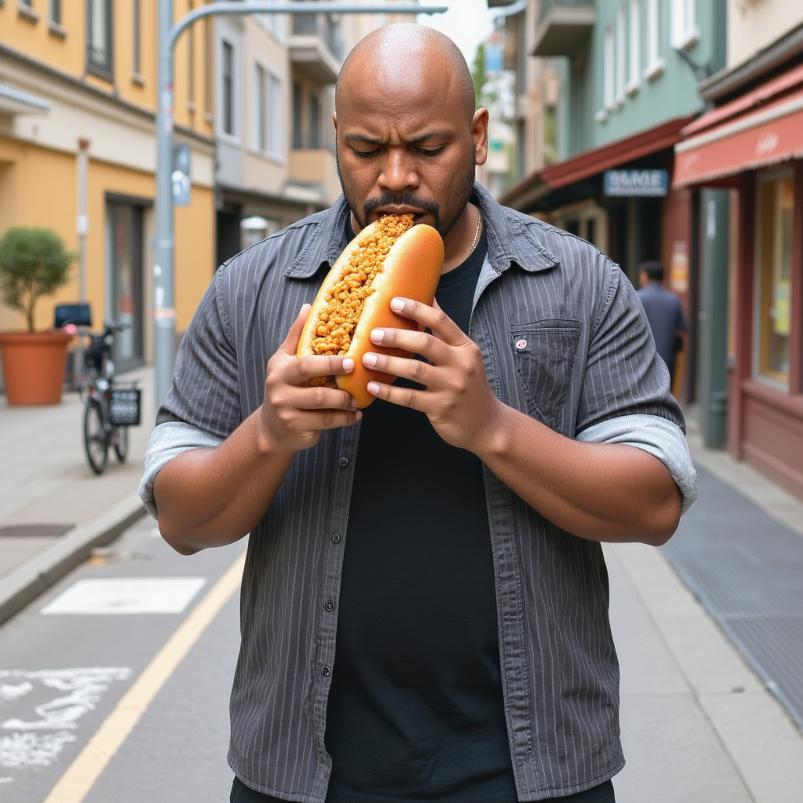}
& \includegraphics[width=.160\linewidth]{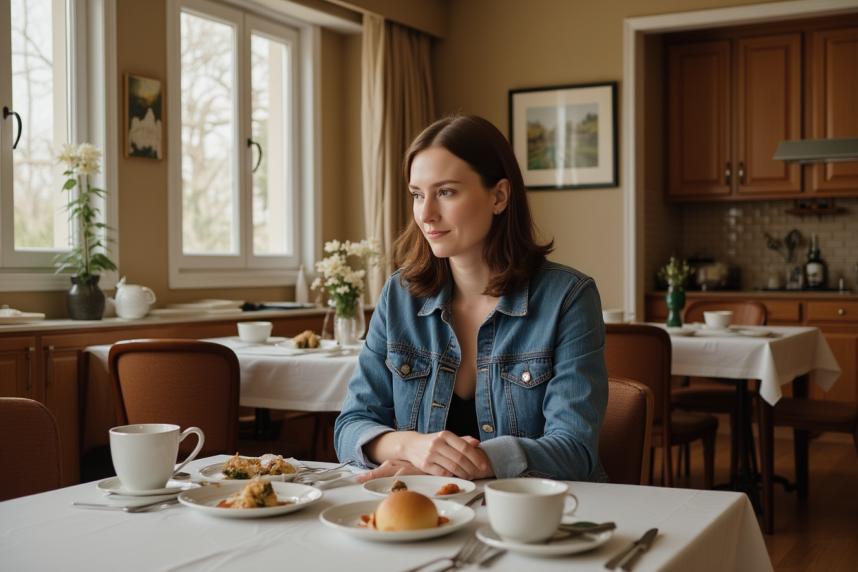}
& \includegraphics[width=.160\linewidth]{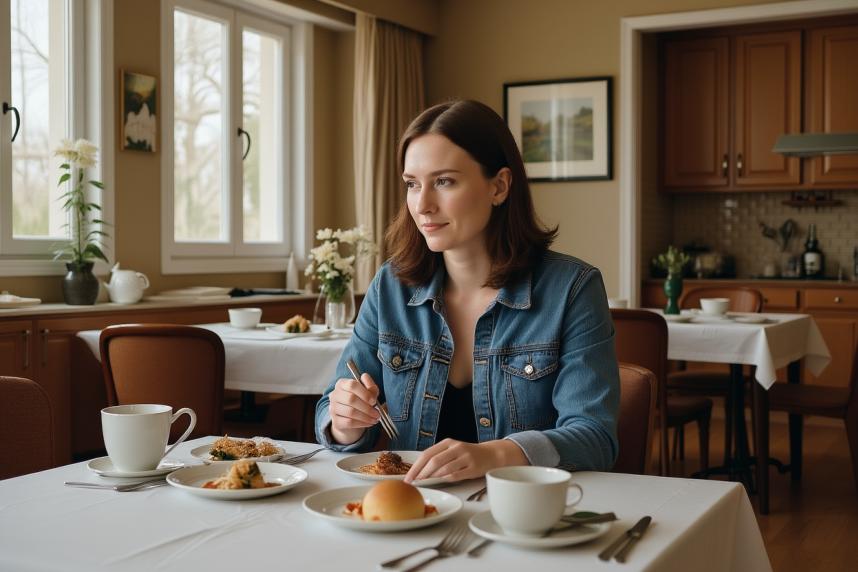}\\[-4pt]

\multicolumn{2}{c}{\small \textcolor{blue}{hold \textrightarrow sip cup}} 
& \multicolumn{2}{c}{\small \textcolor{blue}{hold \textrightarrow eat hotdog}} 
& \multicolumn{2}{c}{\small \textcolor{blue}{sit on \textrightarrow eat at dining table}}\\
\includegraphics[width=.160\linewidth]{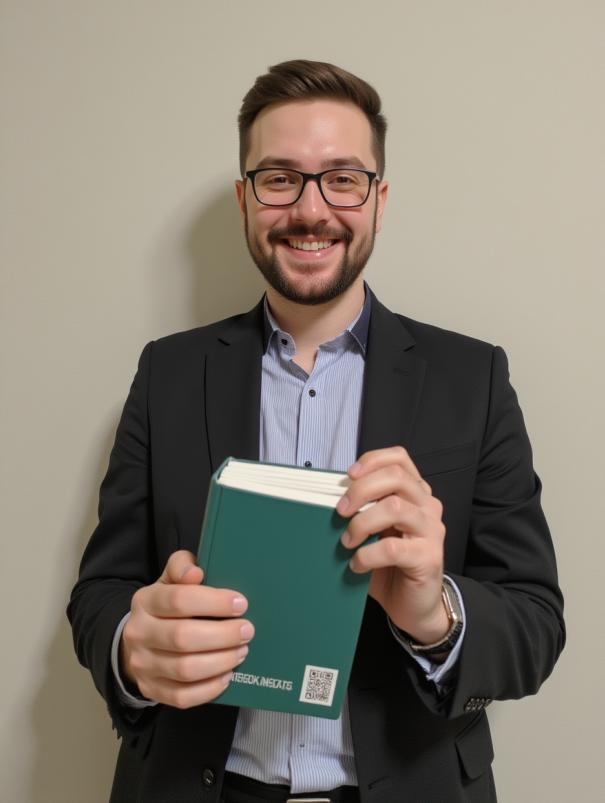}
& \includegraphics[width=.160\linewidth]{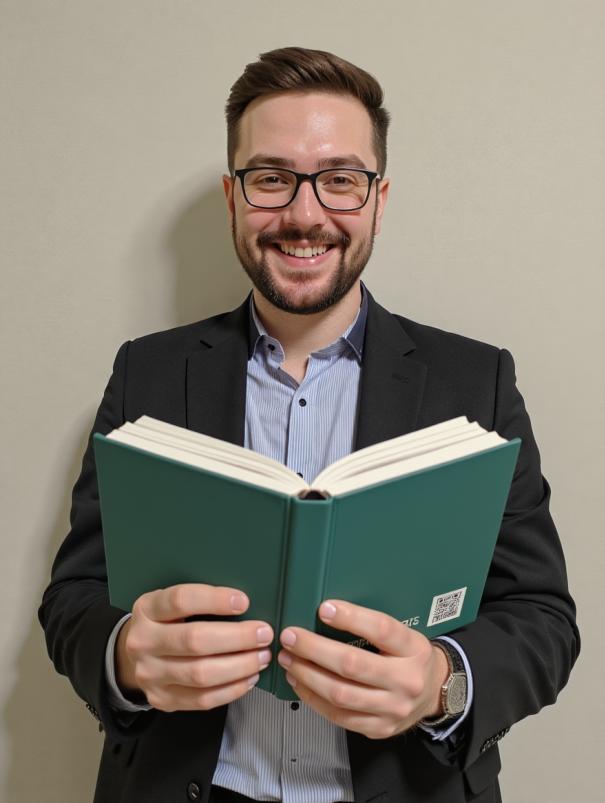}
& \includegraphics[width=.122\linewidth]{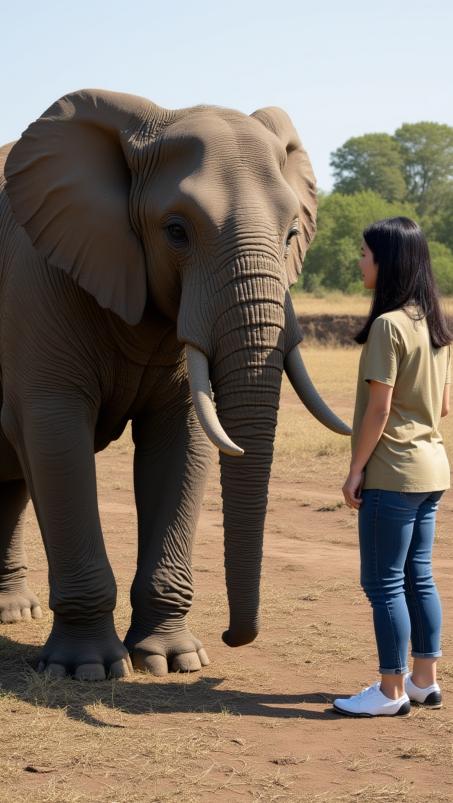}
& \includegraphics[width=.122\linewidth]{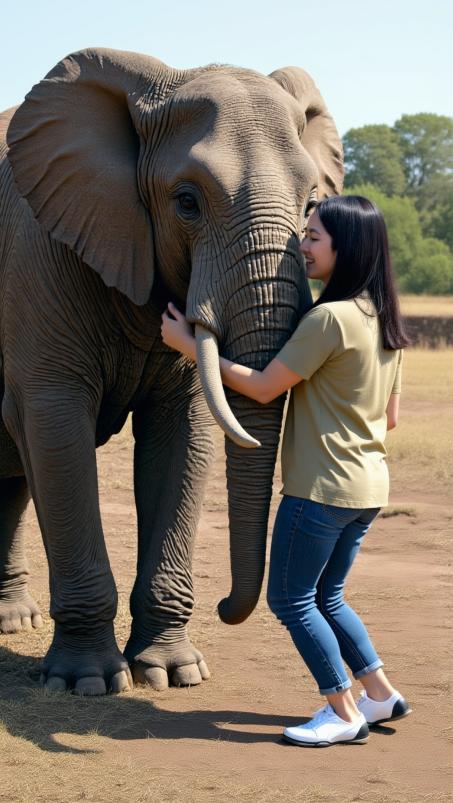}
& \includegraphics[width=.160\linewidth]{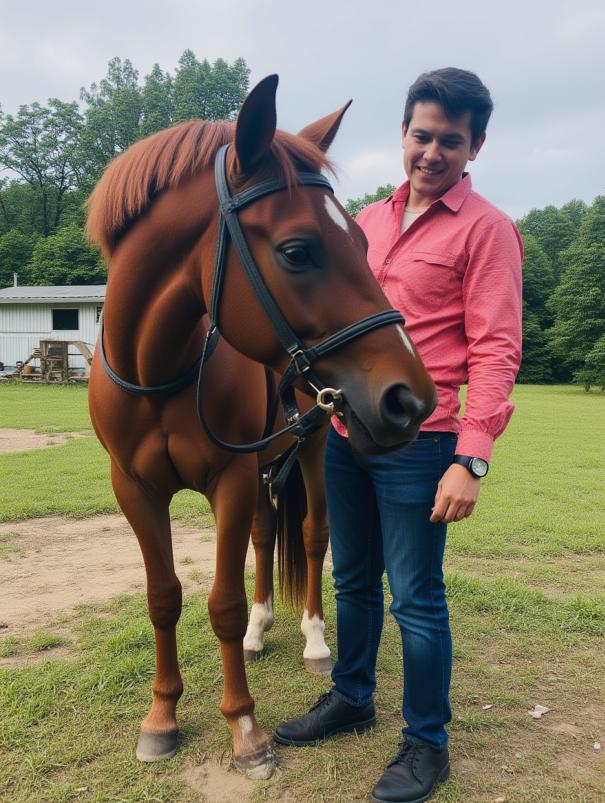}
& \includegraphics[width=.160\linewidth]{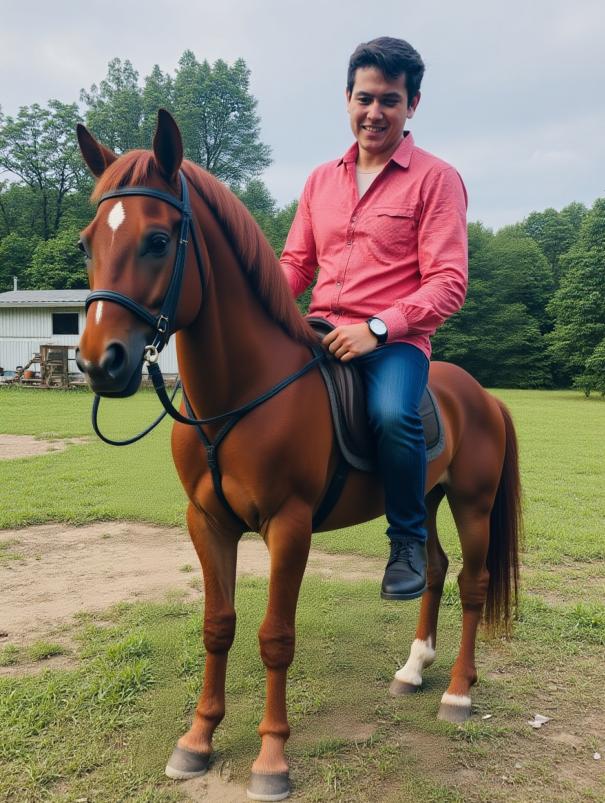} \\ [-4pt]
\multicolumn{2}{c}{\small \textcolor{blue}{hold \textrightarrow read book}} 
& \multicolumn{2}{c}{\small \textcolor{blue}{watch \textrightarrow hug elephant}} 
& \multicolumn{2}{c}{\small \textcolor{blue}{hold \textrightarrow ride horse}}
\end{tabular}
\vspace{-10pt}
\caption{Examples from the HOI-Edit-44K dataset.}
\label{fig:example_hoiedit44k_supp}
\vspace{-10pt}
\end{figure*}
%\vspace{-50pt}

\clearpage

\clearpage

\begin{figure*}[!b]
\centering
\setlength{\tabcolsep}{1pt} % General space between cols (6pt standard)
\renewcommand{\arraystretch}{1} % General space between rows (1 standard) 3.6 / 3.3 
\begin{tabular}{ccccccc}
 & \multicolumn{4}{c}{Object-level methods} & \multicolumn{2}{c}{HOI-level methods} \\ \cmidrule(l){2-5} \cmidrule(l){6-7}
HOI Layout & GLIGEN & MIGC & InstanceDiff & Eligen & InteractDiff & Ours\\
\frame{\includegraphics[width=.138\linewidth]{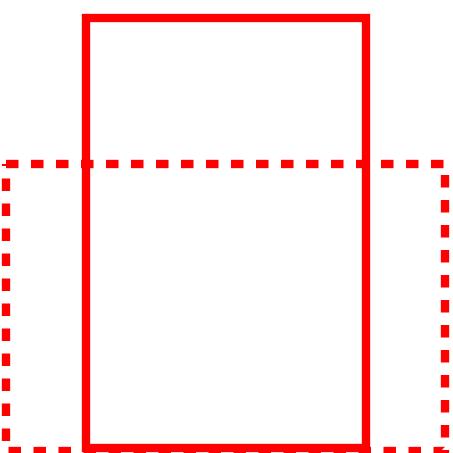}}
& \includegraphics[width=.138\linewidth]{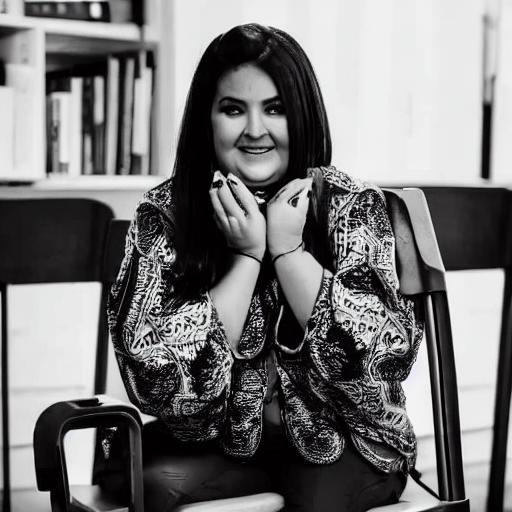}
& \includegraphics[width=.138\linewidth]{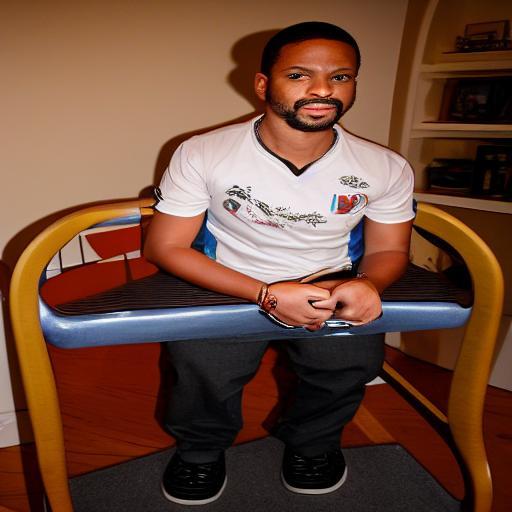}
& \includegraphics[width=.138\linewidth]{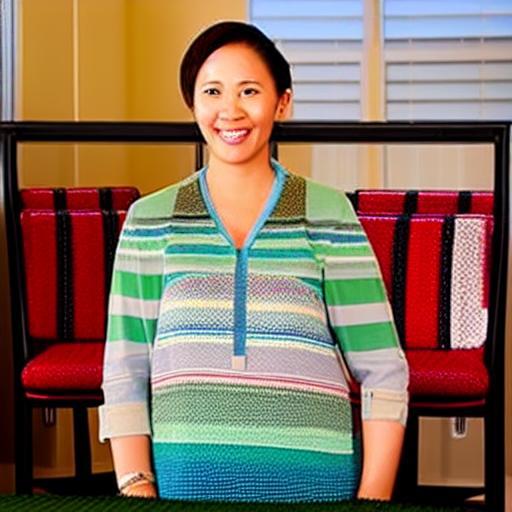}
& \includegraphics[width=.138\linewidth]{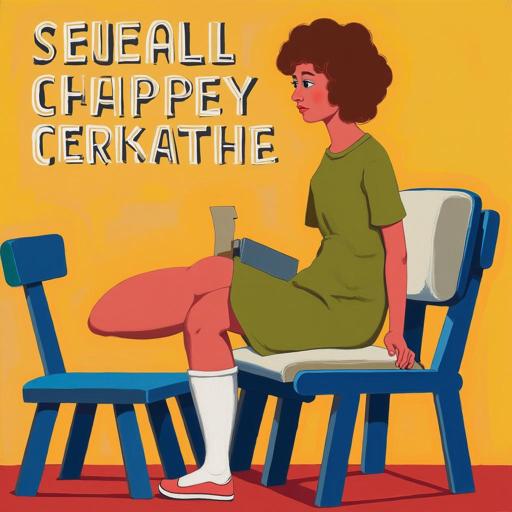}
& \includegraphics[width=.138\linewidth]{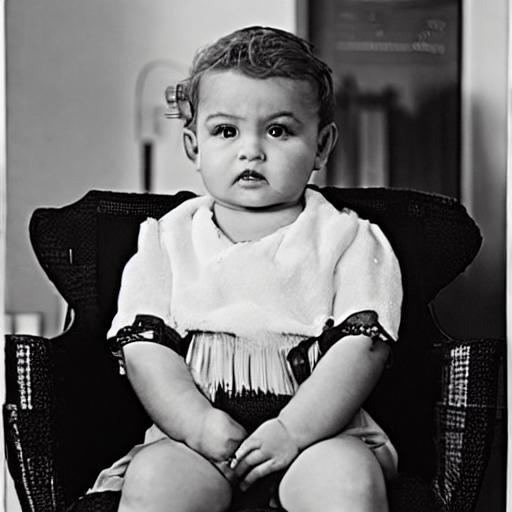}
& \includegraphics[width=.138\linewidth]{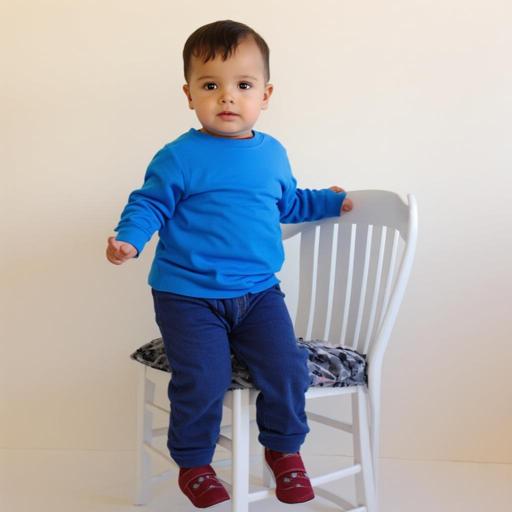}\\[-3pt]
\multicolumn{7}{c}{A child is \textcolor{red}{standing} on a chair}\\
\frame{\includegraphics[width=.138\linewidth]{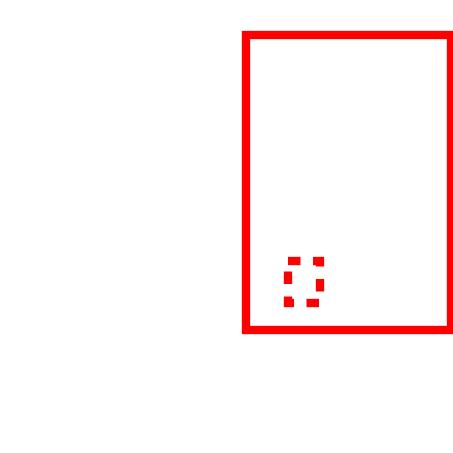}}
& \includegraphics[width=.138\linewidth]{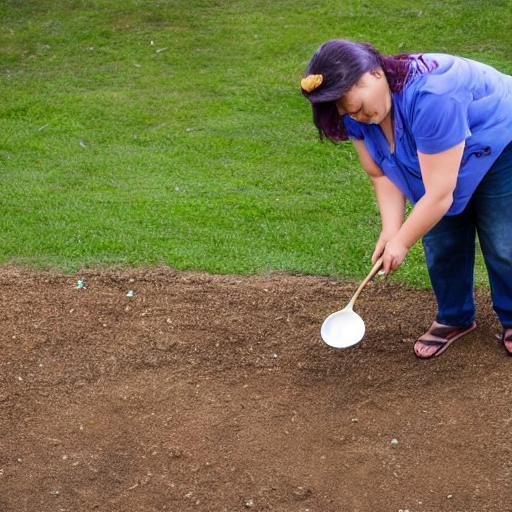}
& \includegraphics[width=.138\linewidth]{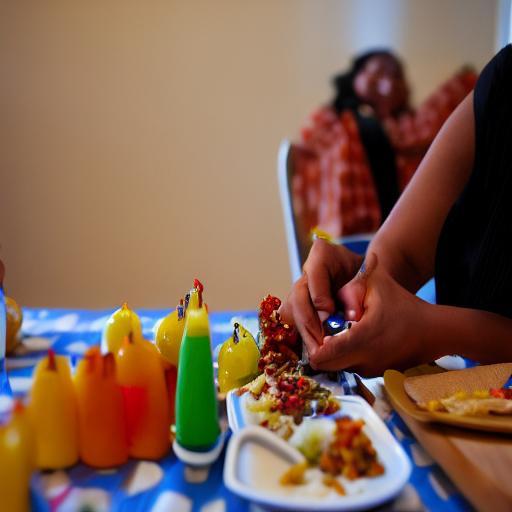}
& \includegraphics[width=.138\linewidth]{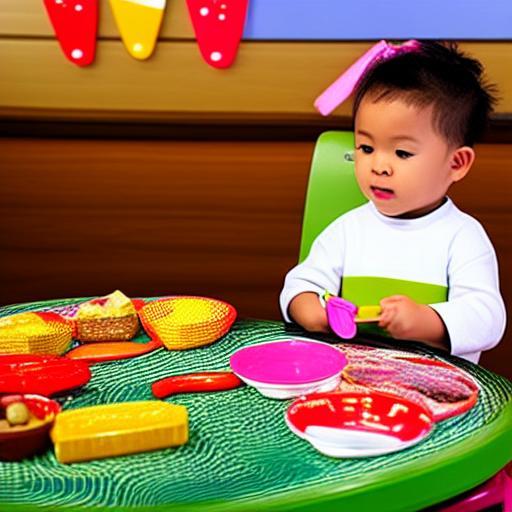}
& \includegraphics[width=.138\linewidth]{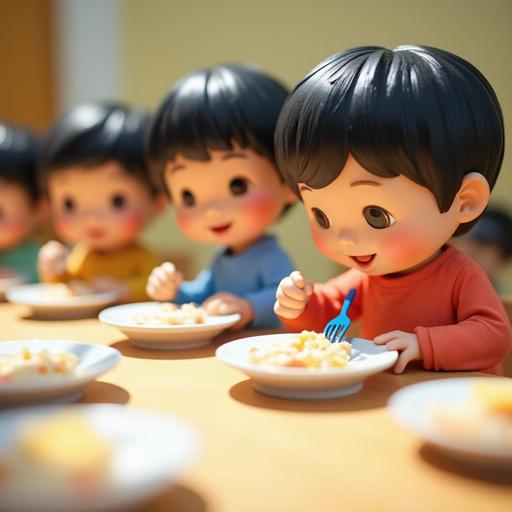}
& \includegraphics[width=.138\linewidth]{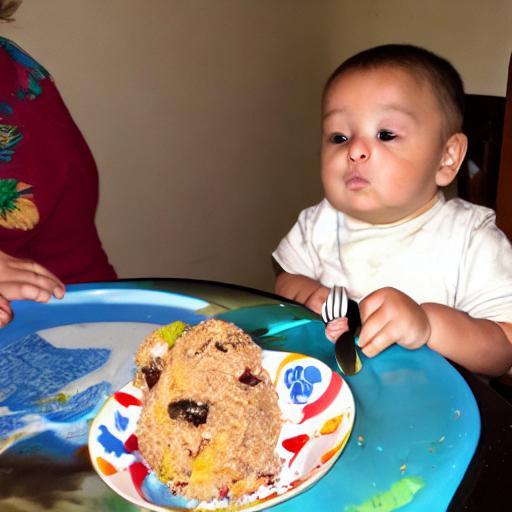}
& \includegraphics[width=.138\linewidth]{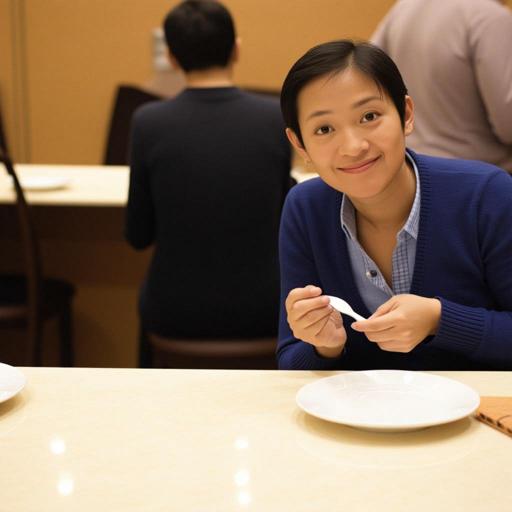} \\ [-3pt]
\multicolumn{7}{c}{A person is \textcolor{red}{holding} a spoon}\\
\frame{\includegraphics[width=.138\linewidth]{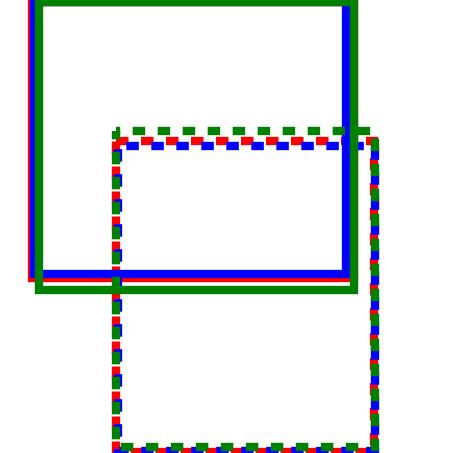}}
& \includegraphics[width=.138\linewidth]{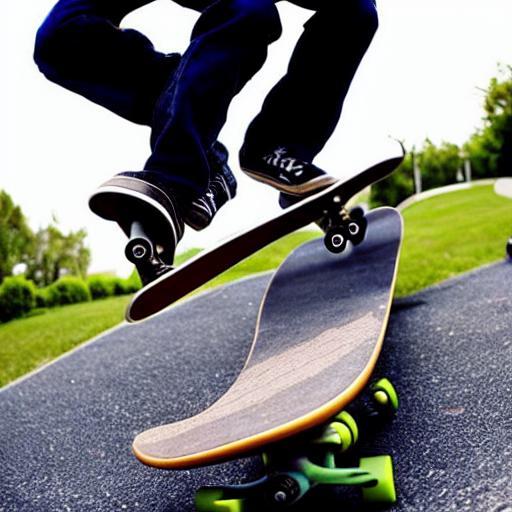}
& \includegraphics[width=.138\linewidth]{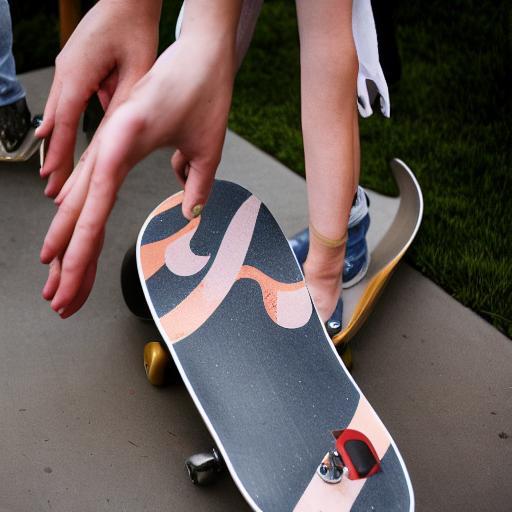}
& \includegraphics[width=.138\linewidth]{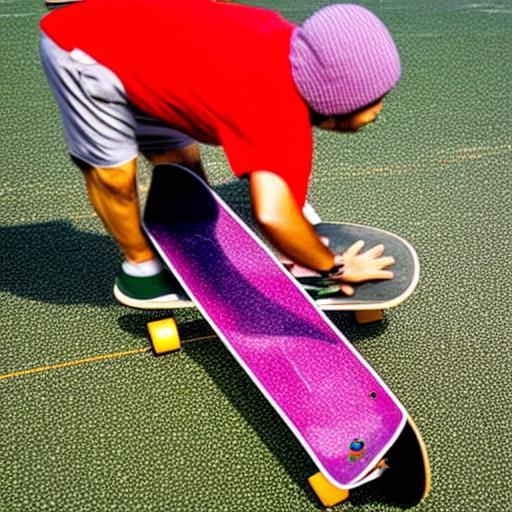}
& \includegraphics[width=.138\linewidth]{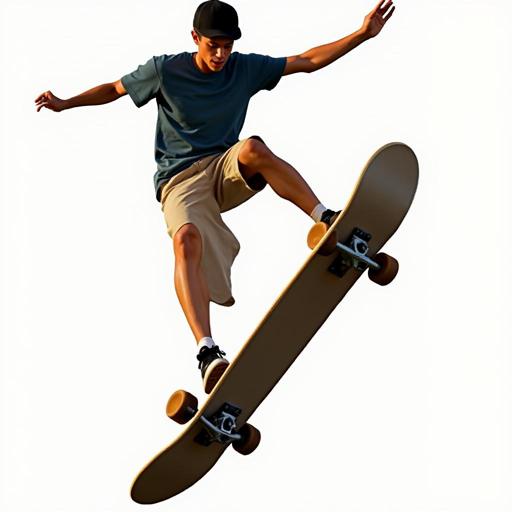}
& \includegraphics[width=.138\linewidth]{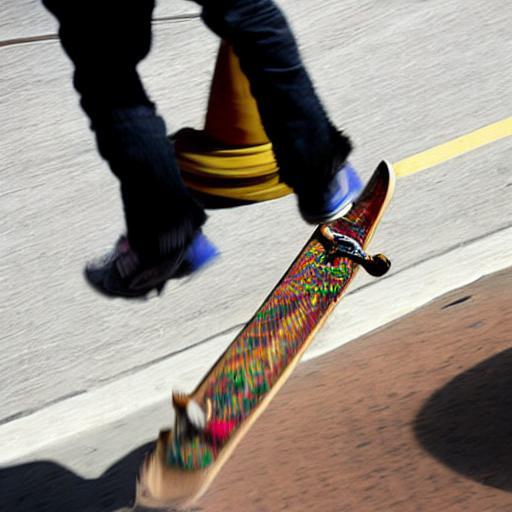}
& \includegraphics[width=.138\linewidth]{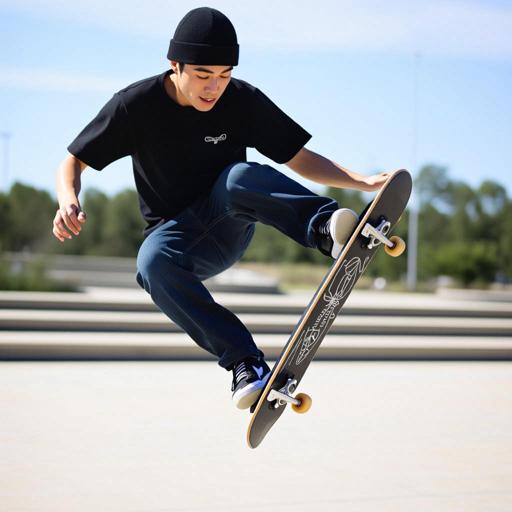} \\ [-3pt]
\multicolumn{7}{c}{A person is \textcolor{red}{flipping}, \textcolor{blue}{jumping} and \textcolor{OliveGreen}{riding} a skateboard}\\
\frame{\includegraphics[width=.138\linewidth]{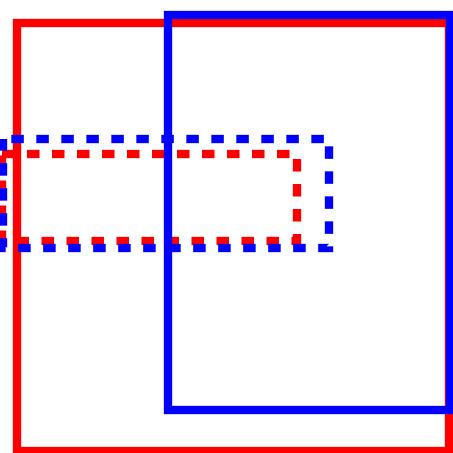}}
& \includegraphics[width=.138\linewidth]{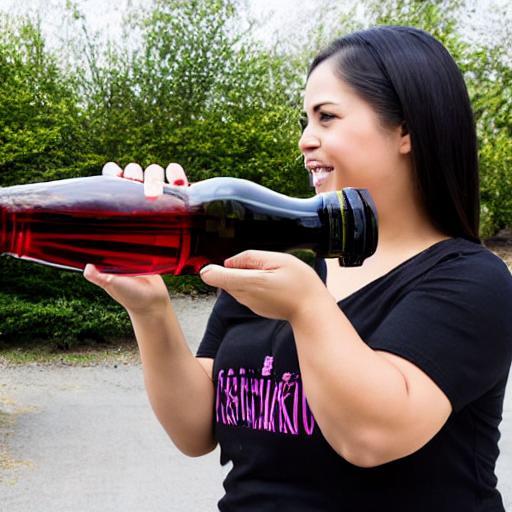}
& \includegraphics[width=.138\linewidth]{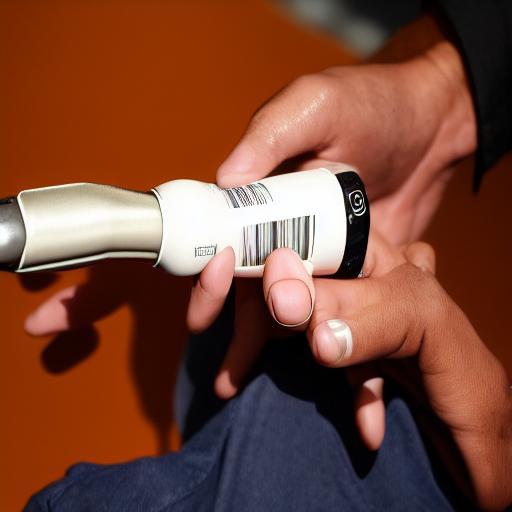}
& \includegraphics[width=.138\linewidth]{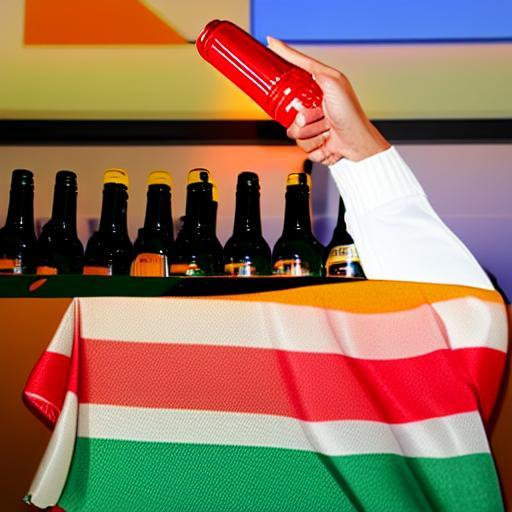}
& \includegraphics[width=.138\linewidth]{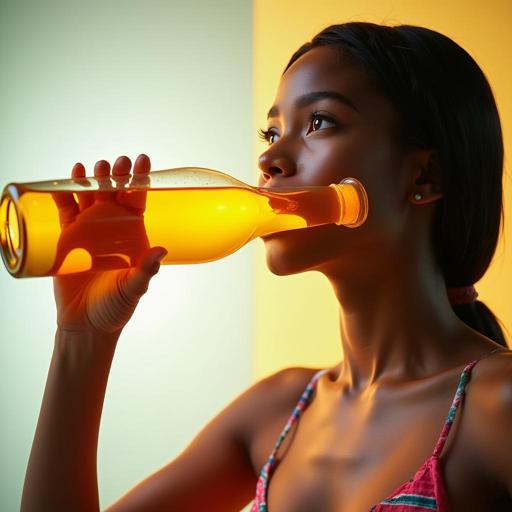}
& \includegraphics[width=.138\linewidth]{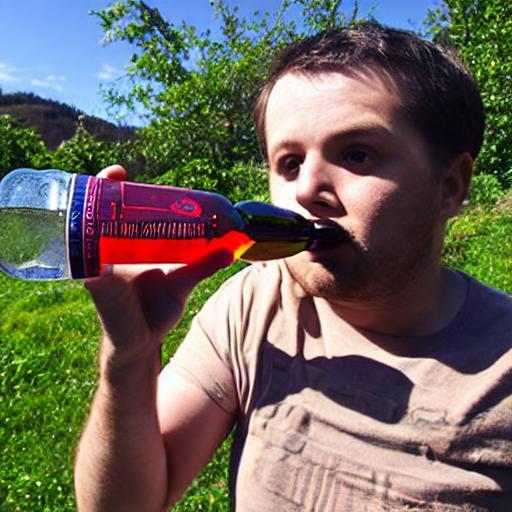}
& \includegraphics[width=.138\linewidth]{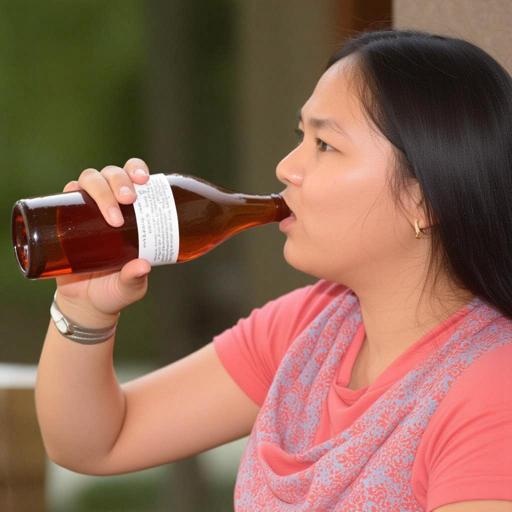} \\ [-3pt]
\multicolumn{7}{c}{A person is \textcolor{red}{drinking with} bottle while \textcolor{blue}{holding} it}
% \frame{\includegraphics[width=.138\linewidth]{assets/fig_gen_supp/x1_y5.jpg}}
% & \includegraphics[width=.138\linewidth]{assets/fig_gen_supp/x2_y5.jpg}
% & \includegraphics[width=.138\linewidth]{assets/fig_gen_supp/x3_y5.jpg}
% & \includegraphics[width=.138\linewidth]{assets/fig_gen_supp/x4_y5.jpg}
% & \includegraphics[width=.138\linewidth]{assets/fig_gen_supp/x5_y5.jpg}
% & \includegraphics[width=.138\linewidth]{assets/fig_gen_supp/x6_y5.jpg}
% & \includegraphics[width=.138\linewidth]{assets/fig_gen_supp/x7_y5.jpg} \\ [-3pt]
% \multicolumn{7}{c}{}\\
% \frame{\includegraphics[width=.138\linewidth]{assets/fig_gen_supp/x1_y6.jpg}}
% & \includegraphics[width=.138\linewidth]{assets/fig_gen_supp/x2_y6.jpg}
% & \includegraphics[width=.138\linewidth]{assets/fig_gen_supp/x3_y6.jpg}
% & \includegraphics[width=.138\linewidth]{assets/fig_gen_supp/x4_y6.jpg}
% & \includegraphics[width=.138\linewidth]{assets/fig_gen_supp/x5_y6.jpg}
% & \includegraphics[width=.138\linewidth]{assets/fig_gen_supp/x6_y6.jpg}
% & \includegraphics[width=.138\linewidth]{assets/fig_gen_supp/x7_y6.jpg} \\ [-3pt]
% \multicolumn{7}{c}{A person is \textcolor{red}{blowing} the cake while another person is \textcolor{blue}{carrying}, \textcolor{OliveGreen}{holding} and \textcolor{orange}{picking up} the same cake}
\end{tabular}
\vspace{-6pt}
\caption{Additional qualitative results for HOI generation. These examples further highlight the limitations of baselines, which often fail to render the specific action even when the objects are placed correctly. 
In the \textbf{first row (standing on a chair)}, all baseline methods incorrectly generate a child sitting on a chair, while our model is the only one that correctly synthesises the `standing on' pose.
Similarly, for \textbf{holding a spoon (row 2)}, baselines produce general eating scenes, with Eligen and InteractDiff showing a fork instead. Our model, in contrast, correctly renders the person holding a spoon.
% This trend continues in the \textbf{third row (washing a train)}, only Eligen and ours successfully generate the "washing" action.
This challenge is more pronounced in complex multi-HOI prompts. For \textbf{row 3 (flipping, jumping, and riding a skateboard)}, baselines fail to capture the `flipping' or `jumping' motions, rendering a simple `riding' pose at best. In \textbf{row 4 (drinking with bottle while holding it)}, most methods fail to combine both `holding' or `drinking'. 
% Finally, for the complex multi-person prompt in \textbf{row 6 (one person blowing... another... carrying, holding, and picking up)}, baselines fail to depict both subjects performing their distinct, simultaneous interactions. 
In contrast, our model generates coherent images that plausibly reflects all specified interactions, demonstrating superior compositional understanding.
}
\label{fig:generation-supp}
\end{figure*}

\begin{figure*}
\centering
\setlength{\tabcolsep}{1pt} % General space between cols (6pt standard)
\renewcommand{\arraystretch}{1} % General space between rows (1 standard) 3.6 / 3.3 
\begin{tabular}{cccccc}
Source & HOIEdit & Qwen Image Edit & Flux.1 Kontext & InteractEdit & Ours\\
\includegraphics[width=.160\linewidth]{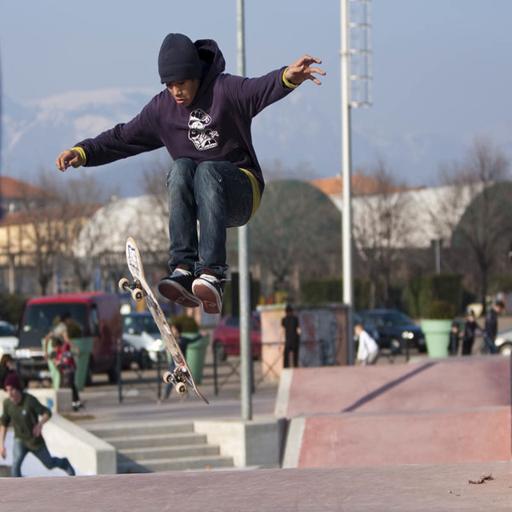}
& \includegraphics[width=.160\linewidth]{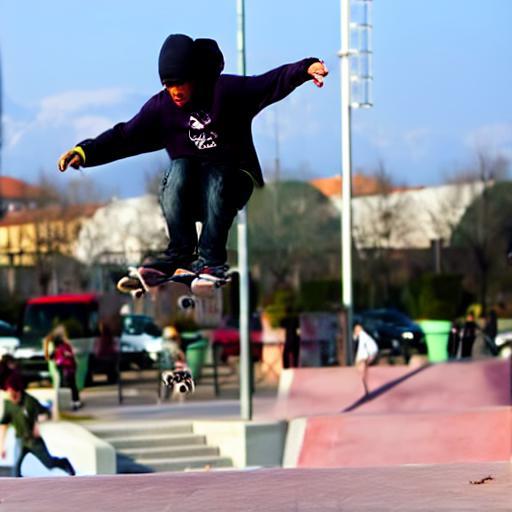}
& \includegraphics[width=.160\linewidth]{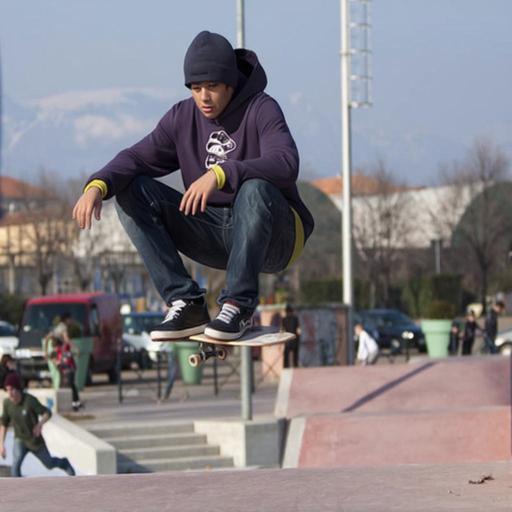}
& \includegraphics[width=.160\linewidth]{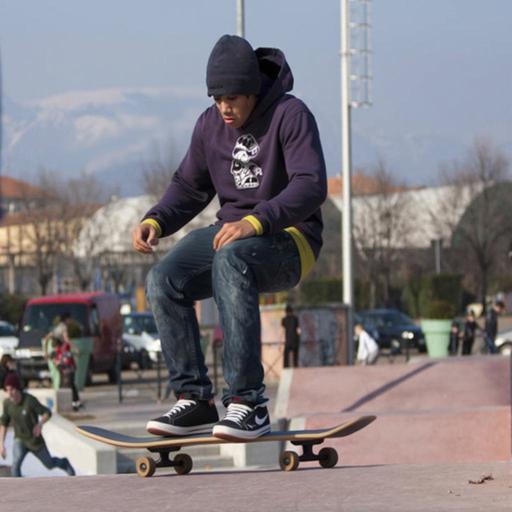}
& \includegraphics[width=.160\linewidth]{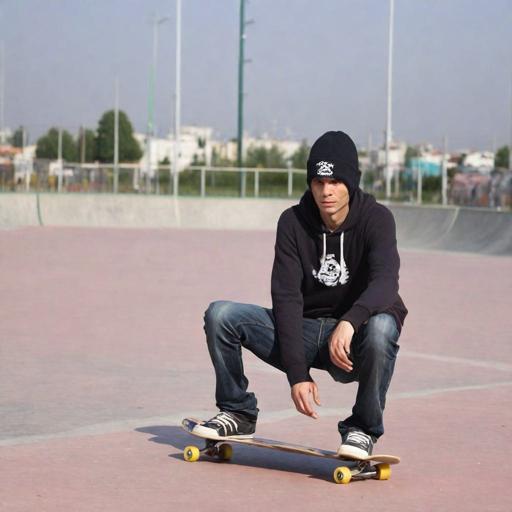}
& \includegraphics[width=.160\linewidth]{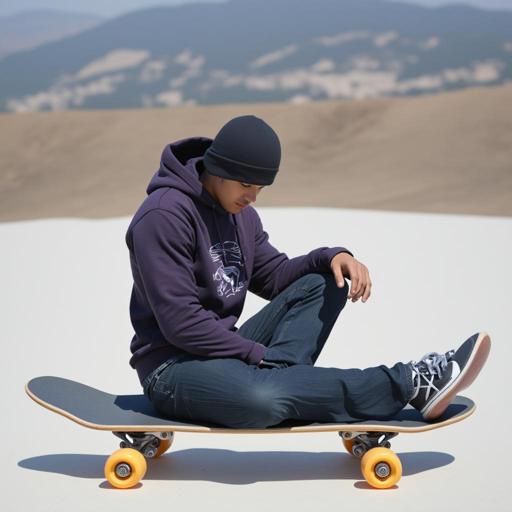}\\ [-3pt]
\multicolumn{6}{c}{jump \textrightarrow sit on skateboard}\\
\includegraphics[width=.160\linewidth]{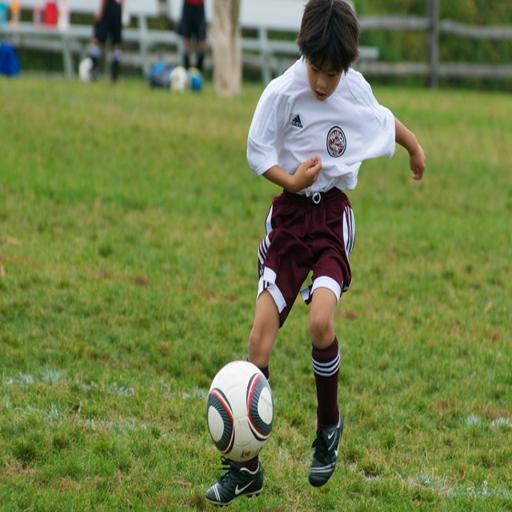}
& \includegraphics[width=.160\linewidth]{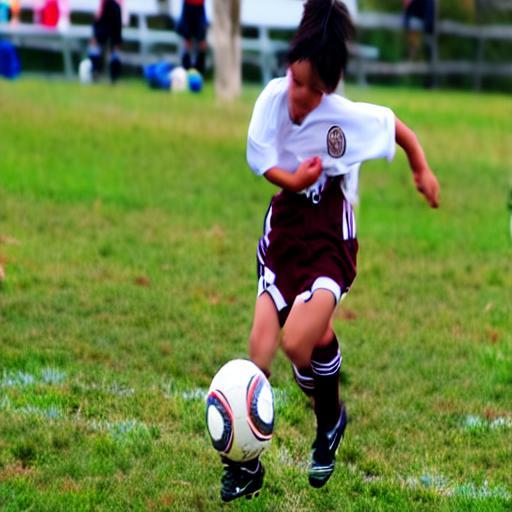}
& \includegraphics[width=.160\linewidth]{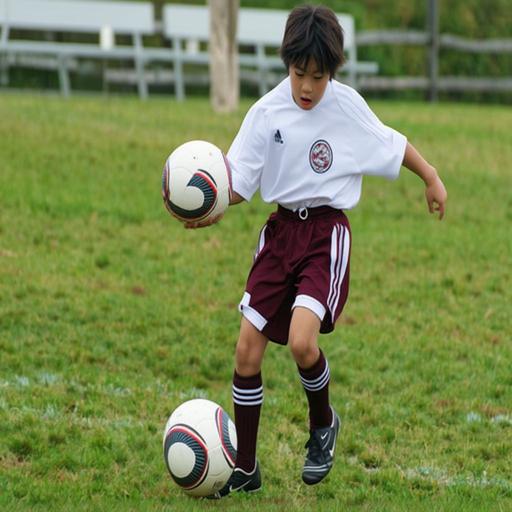}
& \includegraphics[width=.160\linewidth]{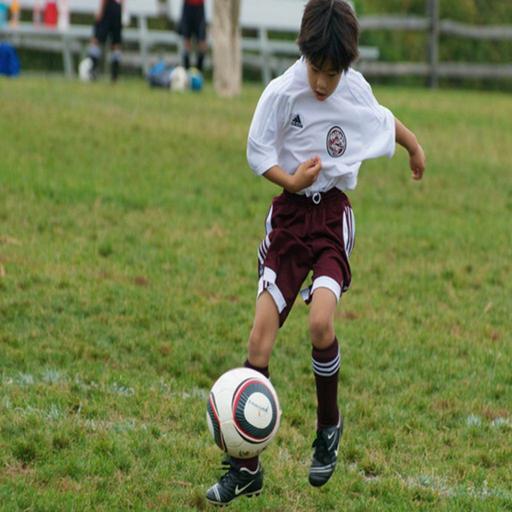}
& \includegraphics[width=.160\linewidth]{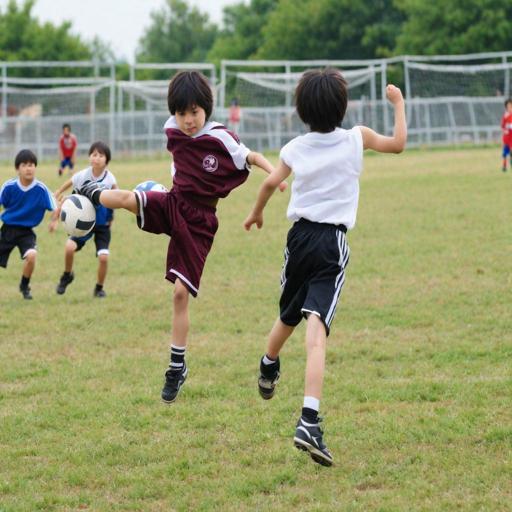}
& \includegraphics[width=.160\linewidth]{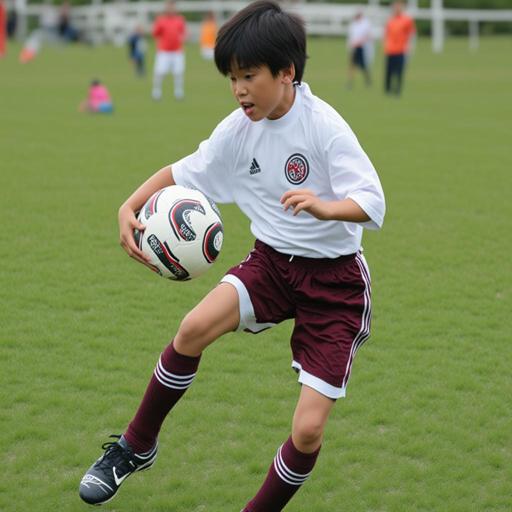} \\ [-3pt]
\multicolumn{6}{c}{kick \textrightarrow hold ball}\\
\includegraphics[width=.160\linewidth]{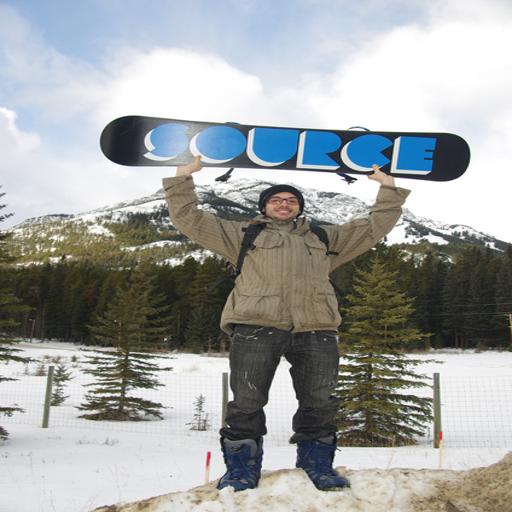}
& \includegraphics[width=.160\linewidth]{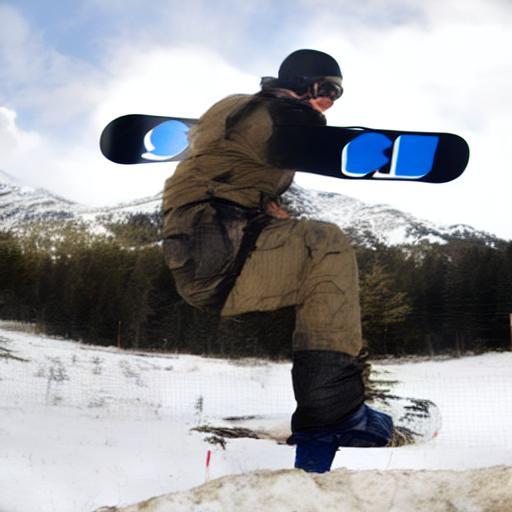}
& \includegraphics[width=.160\linewidth]{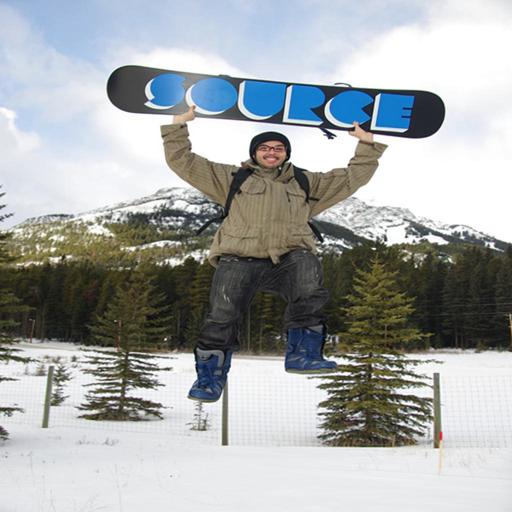}
& \includegraphics[width=.160\linewidth]{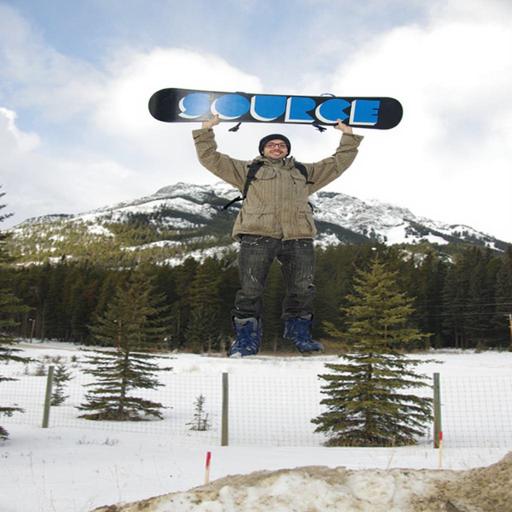}
& \includegraphics[width=.160\linewidth]{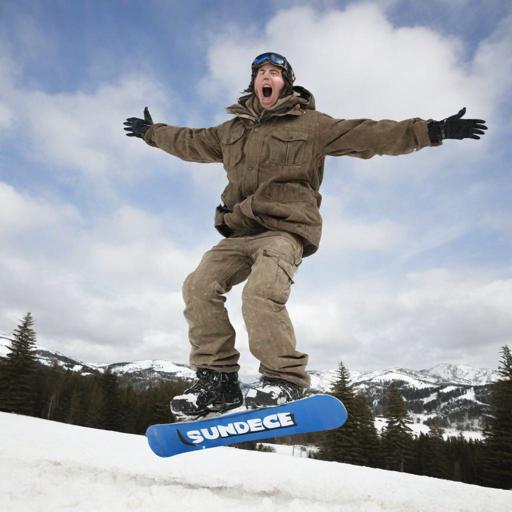}
& \includegraphics[width=.160\linewidth]{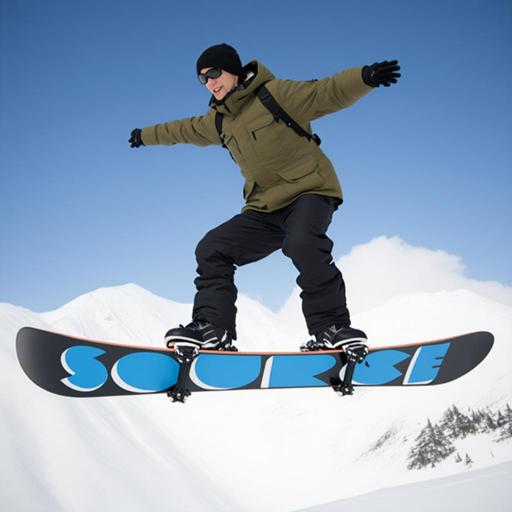} \\ [-3pt]
\multicolumn{6}{c}{hold \textrightarrow jump snowboard}\\
\includegraphics[width=.160\linewidth]{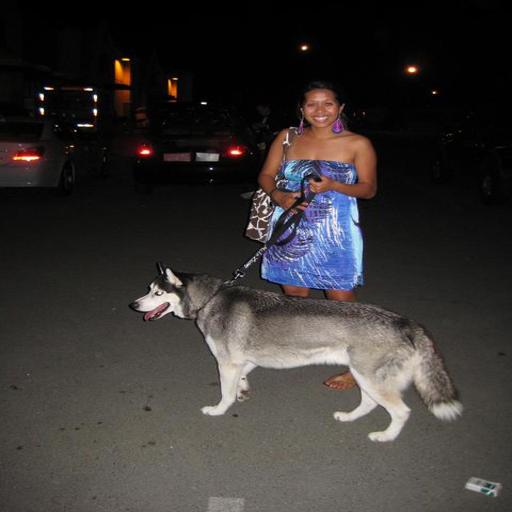}
& \includegraphics[width=.160\linewidth]{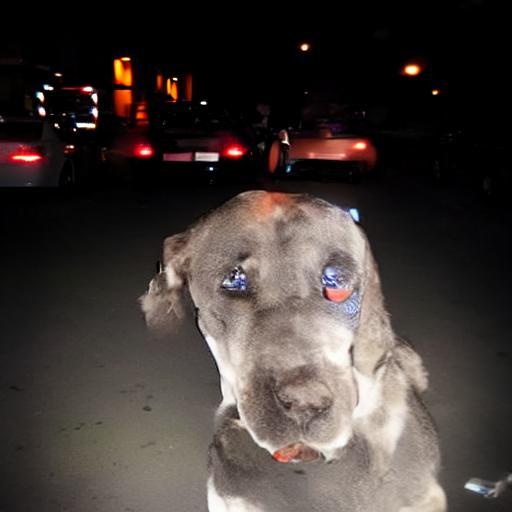}
& \includegraphics[width=.160\linewidth]{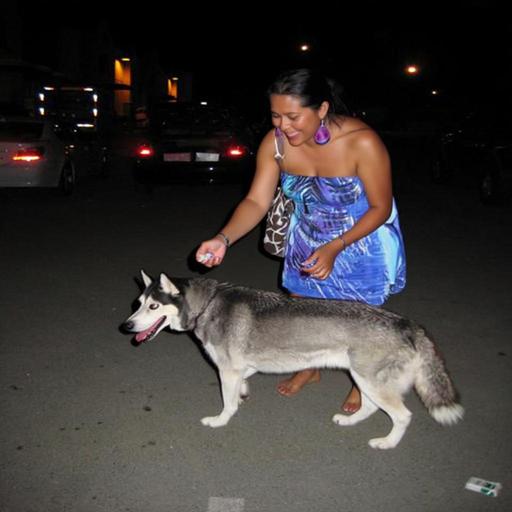}
& \includegraphics[width=.160\linewidth]{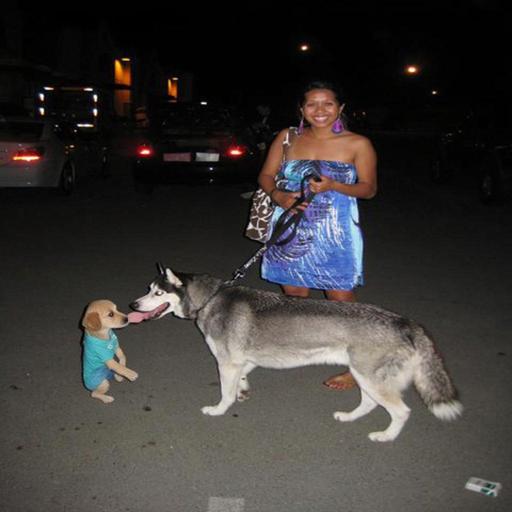}
& \includegraphics[width=.160\linewidth]{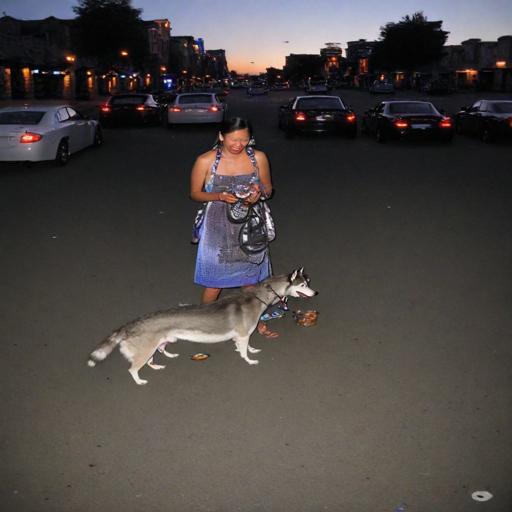}
& \includegraphics[width=.160\linewidth]{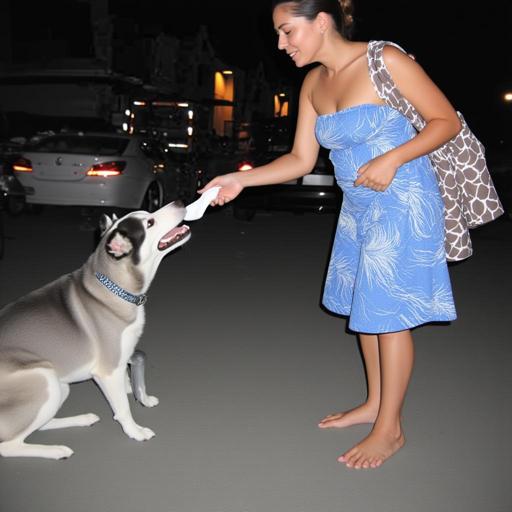} \\ [-3pt]
\multicolumn{6}{c}{walk \textrightarrow feed dog}
\end{tabular}
\vspace{-10pt}
\caption{Additional qualitative comparisons for layout-free HOI edits. \textbf{Row 1 (jump → sit on skateboard)}: Baselines show incorrect poses (Qwen, Flux.1), unnatural actions (InteractEdit), or severe artifacts (HOIEdit), while ours renders ``sit on'' interaction. \textbf{Row 2 (kick → hold ball)}: Most baselines fail to alter the pose, while ours renders the ``hold'' action. \textbf{Row 3 (hold → jump snowboard)}: Most methods failed to render ``jump''. Although InteractEdit renders jump, it fails to preserve the snowboard's identity. Ours renders the jump while maintaining the identity of both the person and the snowboard. \textbf{Row 4 (walk → feed dog)}: Only ours renders a coherent ``feeding'' interaction while preserving the identities of both subjects, demonstrating its superior capability in handling complex relational changes.}
\label{fig:editing_supp}
\vspace{-8pt}
\end{figure*}
\begin{figure*}
\centering
\setlength{\tabcolsep}{1pt} % General space between cols (6pt standard)
\renewcommand{\arraystretch}{1} % General space between rows (1 standard) 3.6 / 3.3 
\begin{tabular}{cccccc}
Source & Target & Source & Target & Source & Target\\
\includegraphics[width=.160\linewidth]{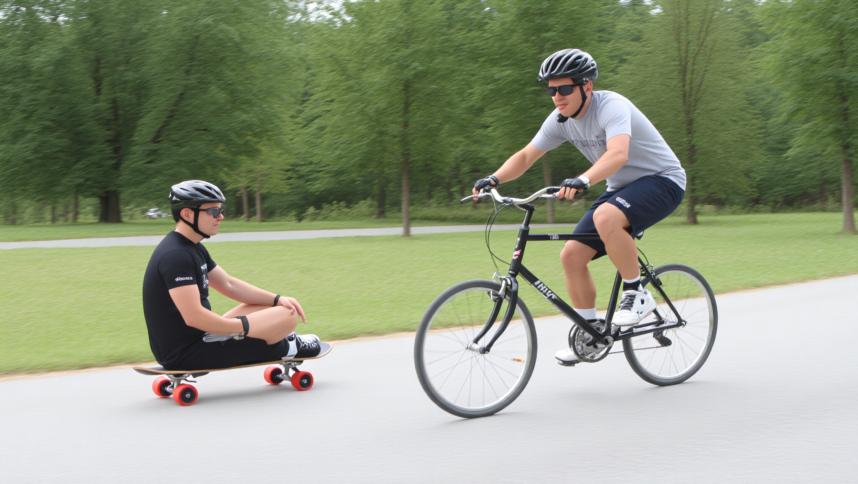}
& \frame{\includegraphics[width=.160\linewidth]{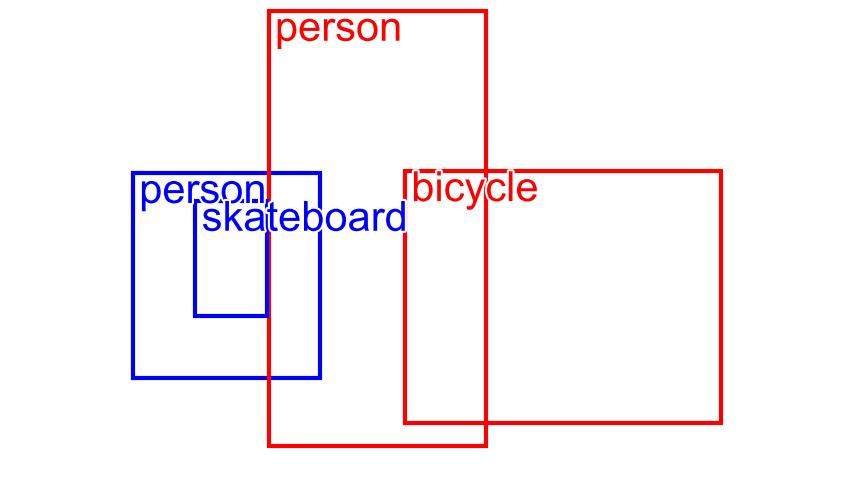}}
& \includegraphics[width=.160\linewidth]{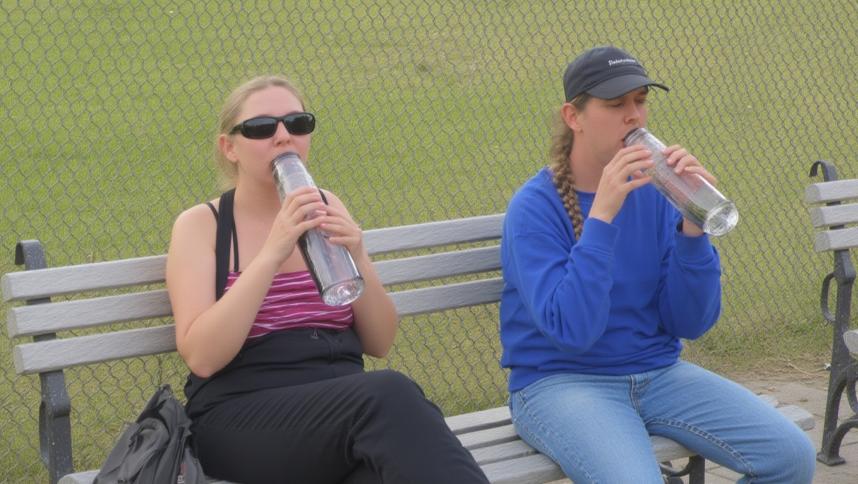}
& \frame{\includegraphics[width=.160\linewidth]{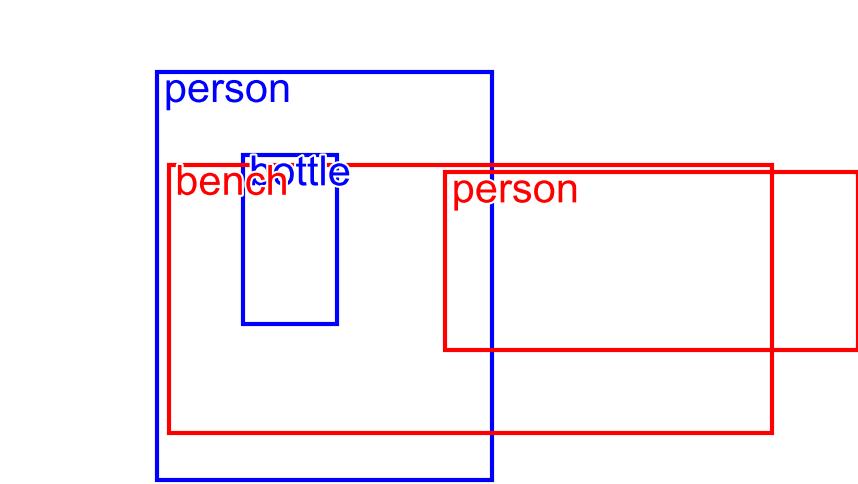}}
& \includegraphics[width=.160\linewidth]{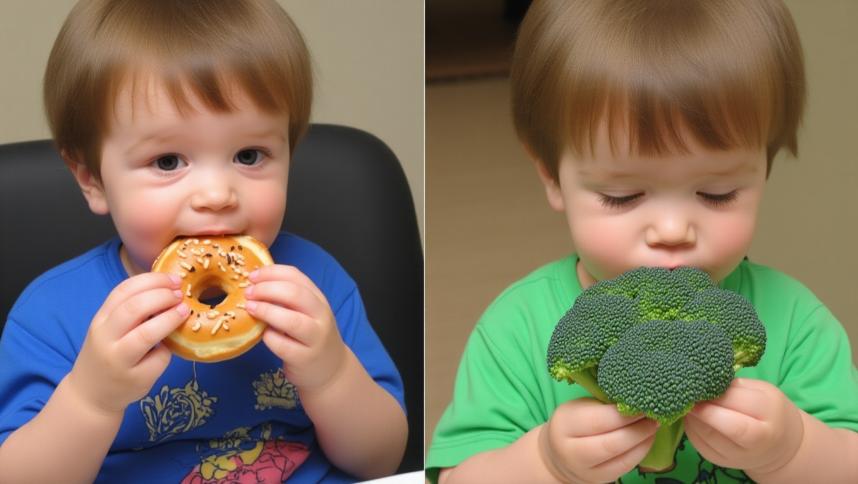}
& \frame{\includegraphics[width=.160\linewidth]{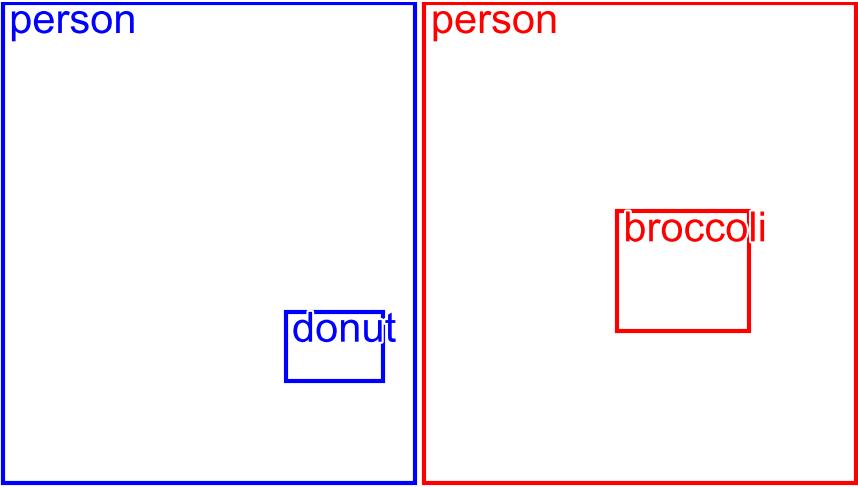}}\\[-4pt]

\multicolumn{2}{c}{\small \textcolor{blue}{sit on \textrightarrow pick up skateboard}} 
& \multicolumn{2}{c}{\small \textcolor{blue}{drink with \textrightarrow carry bottle}} 
& \multicolumn{2}{c}{\small \textcolor{blue}{eat \textrightarrow make donut}}\\[-4pt]
\multicolumn{2}{c}{\small \textcolor{red}{ride \textrightarrow wash bicycle}} 
& \multicolumn{2}{c}{\small \textcolor{red}{sit on \textrightarrow lie on bench}} 
& \multicolumn{2}{c}{\small \textcolor{red}{eat \textrightarrow hold brocolli}}
\\
\includegraphics[width=.160\linewidth]{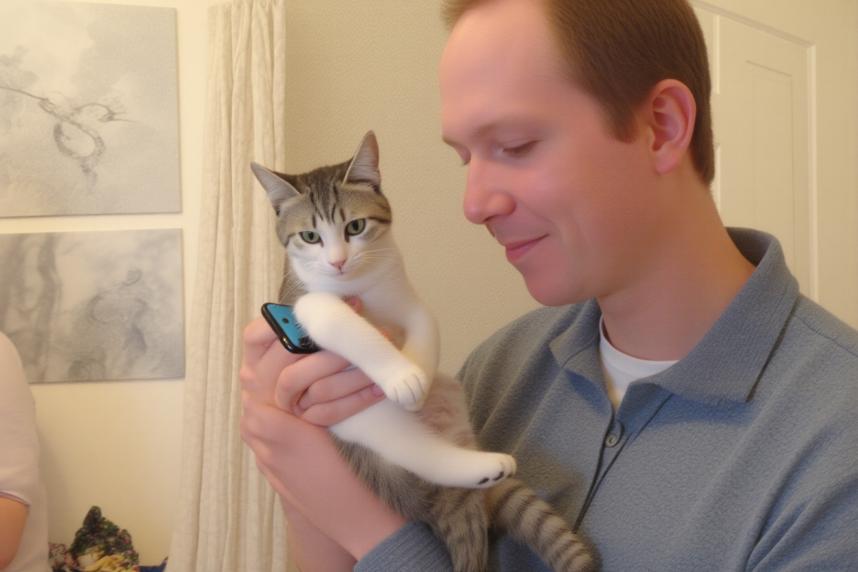}
& \frame{\includegraphics[width=.160\linewidth]{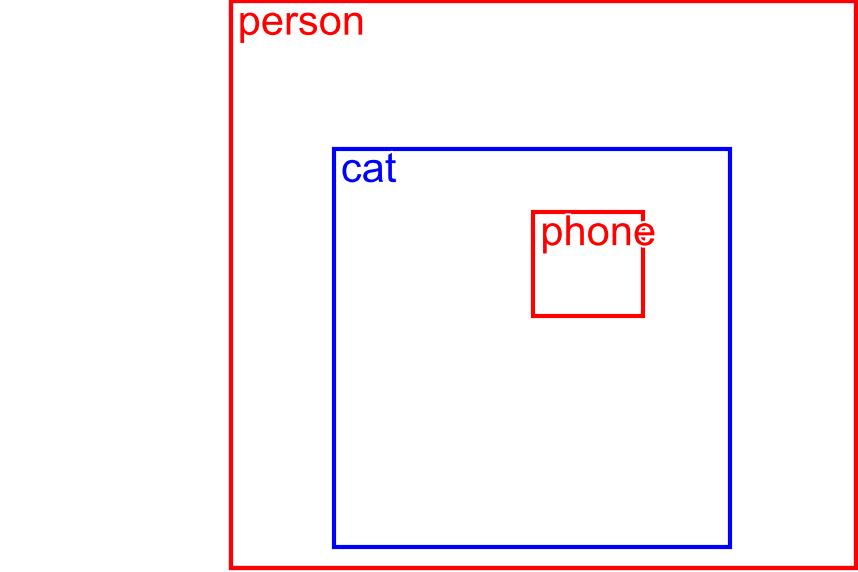}}
& \includegraphics[width=.118\linewidth]{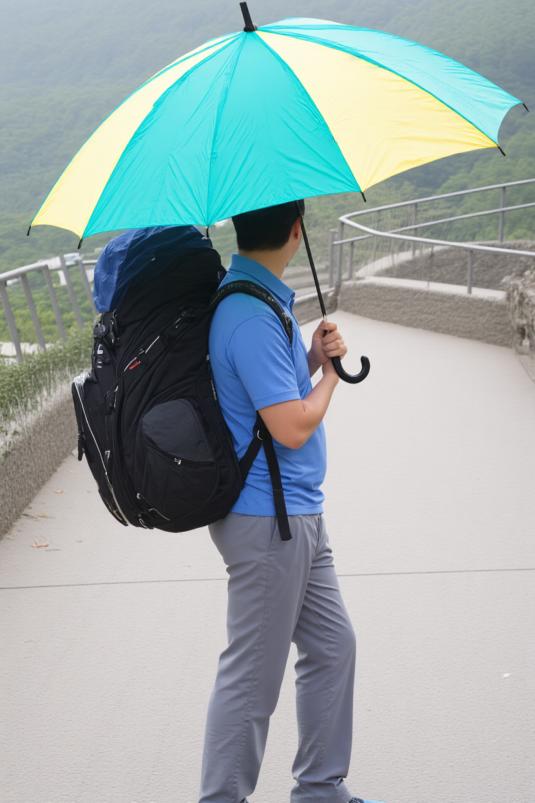}
& \frame{\includegraphics[width=.118\linewidth]{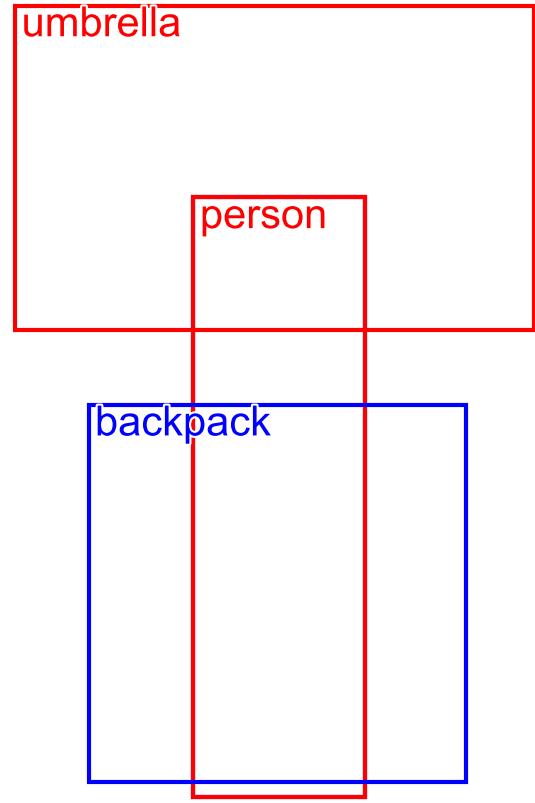}}
& \includegraphics[width=.135\linewidth]{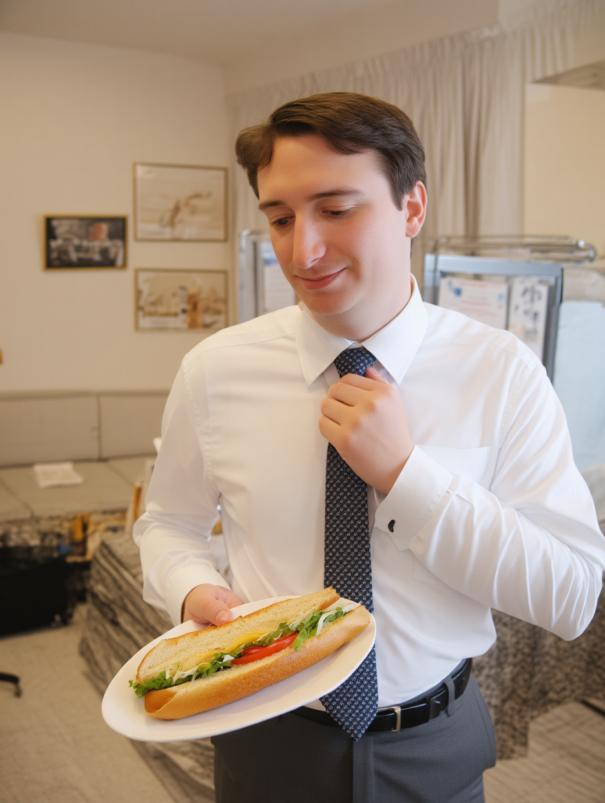}
& \frame{\includegraphics[width=.135\linewidth]{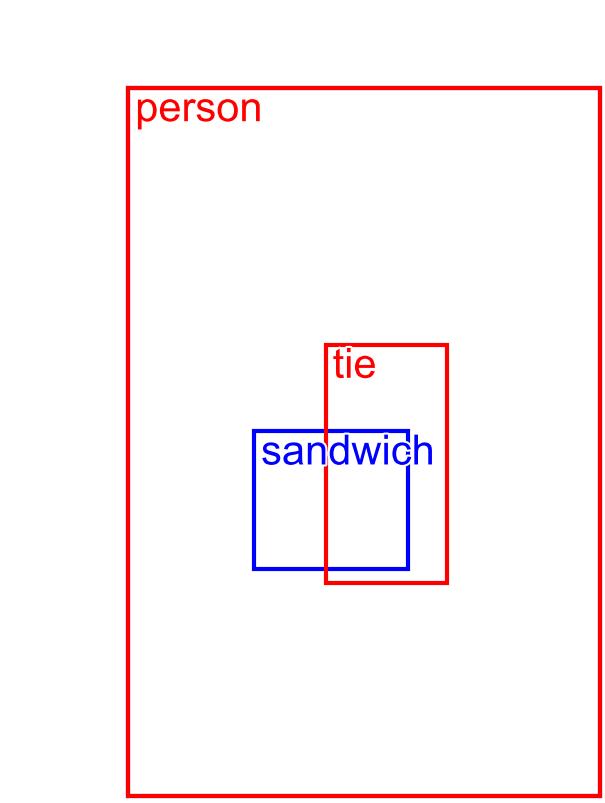}} \\ [-4pt]
\multicolumn{2}{c}{\small \textcolor{blue}{hold \textrightarrow hug cat}} 
& \multicolumn{2}{c}{\small \textcolor{blue}{carry \textrightarrow open backpack}} 
& \multicolumn{2}{c}{\small \textcolor{blue}{carry \textrightarrow eat sandwich}}\\[-4pt]
\multicolumn{2}{c}{\small \textcolor{red}{hold \textrightarrow text on cell phone}} 
& \multicolumn{2}{c}{\small \textcolor{red}{carry \textrightarrow hold umbrella}} 
& \multicolumn{2}{c}{\small \textcolor{red}{adjust \textrightarrow wear tie}}
\end{tabular}
\vspace{-12pt}
\caption{Examples from the MultiHOIEdit benchmark.}
\label{fig:example_multi_supp}
\end{figure*}

\end{document}